\documentclass[reqno]{amsart}

\usepackage[margin=3cm]{geometry}

\usepackage[T1]{fontenc}
\usepackage[numbers,sort&compress]{natbib}
\let\citealp\citep
\usepackage{arXiv_style}
\usepackage{fontawesome}

\makeatletter
\def\blfootnote{\gdef\@thefnmark{}\@footnotetext}
\makeatother

\usepackage{booktabs}
\usepackage{makecell}
\usepackage{amsmath,amssymb}
\usepackage{xfrac}
\usepackage{graphicx}
\usepackage{xcolor}
\usepackage{xspace}
\usepackage{microtype}
\usepackage{tikz}
\usetikzlibrary{arrows.meta,positioning,calc,shapes.geometric,fit,backgrounds,decorations.pathreplacing}
\definecolor{blueM}{HTML}{0072BD}
\definecolor{orangeM}{HTML}{D95319}
\definecolor{yellowM}{HTML}{EDB120}
\definecolor{purpleM}{HTML}{7E2F8E}
\definecolor{greenM}{HTML}{77AC30}
\definecolor{cyanM}{HTML}{4DBEEE}
\definecolor{redM}{HTML}{A2142F}
\definecolor{greyYed}{HTML}{8C8C8C}
\definecolor{cream}{HTML}{FBF6EC}
\definecolor{owlA}{HTML}{884093}
\definecolor{owlB}{HTML}{3F968A}
\definecolor{owlC}{HTML}{4366B5}

\colorlet{cStacked}{purpleM}
\colorlet{cUnweighted}{greyYed}
\colorlet{cBest}{redM}
\colorlet{cAvg}{blueM}
\colorlet{cFrontier}{orangeM}
\colorlet{cTransfer}{purpleM!45}

\colorlet{vClean}{greenM}
\colorlet{vWin}{greenM!60}
\colorlet{vTie}{greyYed}
\colorlet{vChance}{black!22}

\tikzset{
  axisLine/.style={-{Latex[length=1.8mm]}, black!70, line width=0.5pt},
  tick/.style={black!70, line width=0.4pt},
  gridLine/.style={black!10, line width=0.3pt},
  zeroLine/.style={black!45, dashed, line width=0.5pt},
  axlabel/.style={font=\footnotesize, text=black!80},
  ticklabel/.style={font=\scriptsize, text=black!70},
  dslabel/.style={font=\footnotesize},
  vallabel/.style={font=\scriptsize, text=black!60},
  ciBar/.style={line width=1.0pt, black!45},
  leader/.style={black!35, line width=0.3pt},
  mClean/.style={star, star points=5, star point ratio=2.3, fill=vClean,
                 draw=black!55, line width=0.3pt, inner sep=1.7pt},
  mWin/.style={circle, fill=vWin, draw=black!55, line width=0.3pt, inner sep=2.2pt},
  mTie/.style={rectangle, fill=vTie, draw=black!55, line width=0.3pt, inner sep=2.2pt},
  mChance/.style={diamond, fill=vChance, draw=black!45, line width=0.3pt, inner sep=2.0pt},
  pStacked/.style={circle, fill=cStacked, draw=black!55, line width=0.3pt, inner sep=2.2pt},
  pTransfer/.style={circle, fill=cTransfer, draw=cStacked, line width=0.5pt, inner sep=2.2pt},
  pBest/.style={rectangle, fill=cBest, draw=black!55, line width=0.3pt, inner sep=2.0pt},
  pUnw/.style={diamond, fill=cUnweighted, draw=black!55, line width=0.3pt, inner sep=2.0pt},
  collapseArrow/.style={-{Latex[length=1.6mm]}, cStacked, line width=0.9pt},
}

\newcommand{\claude}[1]{}
\DeclareFontFamily{U}{mathb}{\hyphenchar\font45}
\DeclareFontShape{U}{mathb}{m}{n}{<5> <6> <7> <8> <9> <10> gen * mathb <10.95> mathb10 <12> <14.4> <17.28> <20.74> <24.88> mathb12}{}
\DeclareSymbolFont{mathb}{U}{mathb}{m}{n}
\DeclareMathSymbol{\blackdiamond}{\mathbin}{mathb}{"0C}

\newif\iflogosready
\logosreadytrue
\newcommand{\logo}[1]{%
  \iflogosready\raisebox{-.5\height}{\includegraphics[height=0.55cm,width=0.95cm,keepaspectratio]{figures/logos/#1}}%
  \else{\color{black!22}\rule{0.8cm}{0.55cm}}\fi}

\newif\ifowlready
\owlreadytrue
\newcommand{\owl}[3]{%
  \ifowlready\includegraphics[height=#2]{figures/logos/#1}%
  \else{\color{#3}\rule{#2}{#2}}\fi}

\renewcommand{\eg}{e.g.\xspace}
\renewcommand{\ie}{i.e.\xspace}
\newcommand{\wrt}{w.r.t.\xspace}

\newcommand{\kp}{\ensuremath{\kappa}}
\newcommand{\bacc}{\textsc{bacc}}
\newcommand{\defon}{\texttt{def\_on}}

\newcommand{\yes}{\ensuremath{\bullet}}
\newcommand{\no}{\ensuremath{\circ}}
\newcommand{\partialy}{\ensuremath{\odot}}

\newcommand{\dsname}[1]{\textsc{#1}\xspace}
\newcommand{\mdname}[1]{\texttt{#1}\xspace}

\newcommand{\medhallu}{\dsname{MedHallu}}
\newcommand{\ragtruth}{\dsname{RAGTruth}}
\newcommand{\xsum}{\dsname{XSum}}
\newcommand{\cnn}{\dsname{CNN}}
\newcommand{\wice}{\dsname{WiCE}}
\newcommand{\expertqa}{\dsname{ExpertQA}}
\newcommand{\factscore}{\dsname{FActScore}}
\newcommand{\truthfulqa}{\dsname{TruthfulQA}}
\newcommand{\aggrefact}{\dsname{LLM-AggreFact}}

\newcommand{\llama}{\mdname{Llama}}
\newcommand{\qwen}{\mdname{Qwen}}
\newcommand{\mistral}{\mdname{Mistral}}
\newcommand{\gemma}{\mdname{Gemma}}
\newcommand{\phimini}{\mdname{Phi}}
\newcommand{\yi}{\mdname{Yi}}
\newcommand{\commandr}{\mdname{Command-R}}
\newcommand{\glm}{\mdname{GLM}}
\newcommand{\granite}{\mdname{Granite}}
\newcommand{\falcon}{\mdname{Falcon}}

\newcommand{\sonnetfull}{\dsname{Claude Sonnet}}
\newcommand{\sonnet}{\dsname{Sonnet}}

\newcommand{\qwenbig}[1]{\mdname{Qwen2.5-#1}}

\definecolor{unrev}{rgb}{0.85,0.45,0.0}

\providecommand{\tablelegend}[1]{\par\vspace{3pt}\parbox{\linewidth}{\scriptsize #1}}
\usepackage{etoolbox}
\AtBeginEnvironment{table}{\setlength{\abovecaptionskip}{0pt}\setlength{\belowcaptionskip}{4pt}}
\AtBeginEnvironment{table*}{\setlength{\abovecaptionskip}{0pt}\setlength{\belowcaptionskip}{4pt}}

\makeatletter
\renewcommand\paragraph{\@startsection{paragraph}{4}{\z@}{.5\linespacing\@plus.7\linespacing}{-.5em}{\normalfont\itshape}}
\renewcommand{\tocsection}[3]{\indentlabel{\@ifnotempty{#2}{\ignorespaces#2.\quad}}#3}
\makeatother
\usepackage[htt]{hyphenat}

\newcommand{\regimeforestscale}{1}
\newcommand{\regimeforestgap}{2em}

\newcommand{\raimverdictsurl}{\url{https://github.com/eOnofri04/raim-verdicts}}
\newcommand{\raimanalysisurl}{\url{https://github.com/eOnofri04/raim-analysis}}

\hypersetup{
  pdftitle={RAIM: Robust Aggregation of Inexpensive Models for Hallucination Detection},
  pdfauthor={Elia Onofri, Roberto Di Pietro},
  pdfsubject={Computation and Language},
  pdfkeywords={hallucination detection, faithfulness evaluation, LLM-as-a-judge, judge panels, stacked aggregation, correlated errors, frontier substitution}
}

\title[RAIM: Robust Aggregation of Inexpensive Models for Hallucination Detection]{RAIM: Robust Aggregation of Inexpensive Models for Hallucination Detection}

\author[E.\ Onofri and R.\ Di Pietro]{}

\begin{document}

\blfootnote{$^{\star}$ Corresponding Author, (\href{mailto:elia.onofri@kaust.edu.sa}{\faEnvelopeO}) \texttt{elia.onofri@kaust.edu.sa}, (\href{https://www.eliaonofri.it}{\faGlobe}) \texttt{www.eliaonofri.it}}

\maketitle

\vspace{-1em}

\begin{center}
    \begin{minipage}{.89\linewidth}\centering
        \textsc{Elia Onofri}$^{\, \textsc{a},\, \star,\, \orcidlink{0000-0001-8391-2563}}$,
        \textsc{Roberto Di Pietro}$^{\, \textsc{a},\, \orcidlink{0000-0003-1909-0336}}$.
        \\
        \bigskip
        \begin{minipage}{.9\linewidth}\centering
            \footnotesize
            $^\textsc{a}$Computer, Electrical and Mathematical Sciences and Engineering (CEMSE) Division,\\
            King Abdullah University of Science and Technology (KAUST)\\
            Thuwal 23955, Saudi Arabia
        \end{minipage}
    \end{minipage}
\end{center}

\medskip
\thispagestyle{empty}

\begin{abstract}

Automatic evaluation of faithfulness increasingly relies on a large language model acting as a judge, yet the most reliable judges are proprietary frontier models.
Their per-call cost, rate limits, and off-premises inference, however, make them ill-suited to the continuous, high-throughput monitoring that benchmarking and production use demand.

\noindent In this work we investigate three fundamental questions: whether a panel of cheap open-weight judges ($4$--$9$B parameters) can be aggregated to stand in for a frontier one, what the substitution sacrifices in agreement (Cohen's \kp) and accuracy, and when it is worth making.
To address them, we propose RAIM, an aggregation scheme robust to the members' correlated errors, which couples:
($a$) a cross-fitted stacked logistic regression that fits the members' weights jointly, therefore penalising weaker judges echoing a stronger one;
and ($b$) an admissibility test that, read from the members' own outputs, identifies a regime in which aggregating them improves on their best member and stays within reach of the frontier judge.

\noindent We instantiate RAIM under ten judges drawn from disjoint families and we measure the substitution across eight faithfulness benchmarks.
Read as paired differences against \sonnetfull, the substitution clearly improves on one benchmark and clearly worsens on three (only two by a non-negligible margin), leaving the remaining four unresolved.
The panel retains a median $93\%$ of the frontier judge's \kp\ agreement and gives up only $2.9$ points of balanced accuracy on average.
Inference runs at a sixty-fourth of the frontier's price at conventional cloud rates, or at the electricity cost on premises, so the operative expense is a one-time in-domain calibration of some fifty to a hundred labelled records, amortised over a few tens of thousands of evaluated items.

\noindent Against the literature, the panel is competitive with purpose-trained detectors on their home benchmarks, losing $1.3$ points of accuracy to \mdname{GPT-4o} and $1.9$ to the best \aggrefact leaderboard model;
yet the strongest detector we reran falls $6$ points behind on our grounded sets, and cannot score the ungrounded ones.

\noindent In conclusion, whether pursuing the aggregation is worthwhile depends on the members themselves, making the gain auditable from how widely competence is spread across them and how far their errors decorrelate --- both read at no further cost off the aggregator's calibration set.
If several capable members err on different items, the panel improves on its best judge and approaches the frontier;
conversely, where one dominates, the stacker recovers the leader, yet only there does the frontier remain materially ahead.
A panel of cheap judges can therefore stand in for a frontier one at a small fraction of the cost wherever this audit admits it.

\medskip


\end{abstract}

\vspace{-.5em}

\begin{multicols}{2}\footnotesize
    \tableofcontents
\end{multicols}

\newpage

\section{Introduction}
\label{sec:intro}

Automatic evaluation of faithfulness --whether a given claim, a retrieved passage, or a model's output is supported by its source or constitutes a hallucination-- increasingly relies on a large language model acting as a judge (LLM-as-a-judge;~\citealp{Zheng_Chiang_Sheng_etal_23,Li_Dong_Chen_etal_24}).
Single-model judges, however, carry systematic biases~\citep{Panickssery_Bowman_Feng_24,Chen_Goldfarb-Tarrant_25} and are reported to reach human-level agreement only at scales well above the small models (${<}10$B parameters) a deployer can serve locally~\citep{Han_Titericz-Junior_Balough_etal_25}.
As a consequence, the judges one can most rely on are large proprietary models served from the cloud, so each evaluated item is a metered, rate-limited call, whose cost is small individually yet adds up at scale, and which sends the text off-premises to an endpoint the deployer does not control.

Three main strategies have been proposed to reduce this dependence: \emph{escalate}, keeping a frontier judge but calling it only where a cheap model is unsure~\citep{Chen_Zaharia_Zou_23,Jung_Brahman_Choi_25}; \emph{specialise}, training one small model expressly for the task~\citep{Tang_Laban_Durrett_24,Kim_Suk_Longpre_etal_24}; or \emph{aggregate}, running several cheap judges and combining their verdicts~\citep{Verga_Hofstatter_Althammer_etal_24}.
Escalation, however, retains a frontier judge as a fallback, and we read it as a complementary solution (Appendix~\ref{app:related}).
Specialists are typically trained for reference-conditioned verification and, whilst holding an advantage on home ground, they cannot score a task that supplies no reference at all.
Moreover, they are reported to generalise poorly off the distribution they were trained on~\citep{Huang_Bu_Zhou_etal_25}, as we confirm by running them ourselves across our suite (Appendix~\ref{app:ptj}).
Specialisation and aggregation both remove the frontier model from the loop, as the constraints beyond price actually require (\ie on-premises inference, no rate limit, no deprecation).
Of the two, we pursue aggregation, the only one that needs no training of model weights and keeps the generality just noted.
The practical question is therefore not whether a frontier judge is good, but whether one has to be paid for: can several cheap, open-weight judges, run locally and combined, stand in for it?
We refer to Appendix~\ref{app:related} for a deeper account of these routes, of the noisy-voter literature the aggregator inherits, and of how recent panel studies compare with ours (Appendix~\ref{app:roadmap} maps every appendix to the claim it supports).

The intuition that the \emph{aggregate} solution might work is the classical one behind ensembling and the Condorcet jury theorem~\citep{Condorcet_85}, and a panel of small judges has indeed been reported to track human ratings better than a single frontier model at a fraction of its price~\citep{Verga_Hofstatter_Althammer_etal_24}.
That intuition rests, however, on an assumption which is fragile for exactly these estimators.
Weak voters rival a strong one when their errors are \emph{independent}, whereas judges drawn from a common pretraining prior err together~\citep{Zhao_Shin_Huang_etal_26,Hossain_Yousefi_Lim_26}.
Indeed, a panel of nine frontier judges is reported to carry only about two effective independent votes, leaving the best single judge unbeaten~\citep{Kohli_26}.
What this line of work leaves open is hence not if a cheap panel pays on average, but under which conditions it pays at all, and what the substitution costs where it does not.

\begin{figure*}[t]
\centering
\resizebox{\linewidth}{!}{\begin{tikzpicture}[
    lnode/.style={inner sep=0pt, minimum width=0.95cm, minimum height=0.60cm},
    owln/.style={inner sep=0pt},
    panelbox/.style={draw=cStacked!70, dashed, thick, rounded corners, inner sep=7pt},
    innerbox/.style={draw=black!35, rounded corners=2pt, inner sep=4pt},
    aggbox/.style={draw=cStacked, rounded corners, fill=cStacked!15,
                   minimum width=1.7cm, minimum height=1.25cm},
    outn/.style={draw=black!55, rounded corners, fill=white, minimum width=1.95cm,
                 minimum height=0.85cm, align=center, font=\small},
    cmpn/.style={draw, rounded corners, minimum width=2.25cm, minimum height=0.8cm,
                 align=center, font=\scriptsize, fill=white},
    badge/.style={rounded corners=2pt, inner sep=2.5pt, font=\scriptsize, draw},
    ar/.style={-{Latex[length=1.8mm]}, line width=1pt, black!78},
    arg/.style={-{Latex[length=1.6mm]}, line width=0.9pt},
]
  \def\lpx{0.98}
  \def\lpy{0.62}
  \def\owlh{0.60cm}

  \begin{scope}[shift={(3.31,0)}]
    \foreach \fam/\c/\r in {%
        llama/0/0, qwen/1/0, mistral/2/0, gemma/3/0, phi/4/0,
        yi/0/1, command/1/1, glm/2/1, granite/3/1, falcon/4/1}{
      \node[lnode] (j-\fam) at ({\c*\lpx-1.28}, {0.31-\r*\lpy}) {\logo{\fam}};
    }
  \end{scope}
  \node[innerbox, fit=(j-llama)(j-phi)(j-yi)(j-falcon)] (parl) {};
  \node[font=\small\itshape, align=center, above=1pt of parl] (lbl)
        {a parliament of ten cheap judges (4--9B)};

  \node[aggbox, right=2.4cm of parl] (agg) {};
  \begin{scope}[shift={(agg.center)}]
    \draw[black!40, line width=0.5pt, -{Latex[length=1mm]}] (-0.5,-0.3) -- (0.55,-0.3);
    \draw[black!40, line width=0.5pt, -{Latex[length=1mm]}] (-0.5,-0.32) -- (-0.5,0.34);
    \draw[cStacked, line width=1.2pt, smooth, samples=40, domain=-0.45:0.45]
        plot ({\x},{0.55/(1+exp(-13*\x))-0.3});
  \end{scope}
  \node[below=1pt of agg, font=\scriptsize, align=center, text=cStacked!85!black]
        (agglbl) {stacked logistic\\regression};

  \node[owln] (owlA) at (1.80,-1.50) {\owl{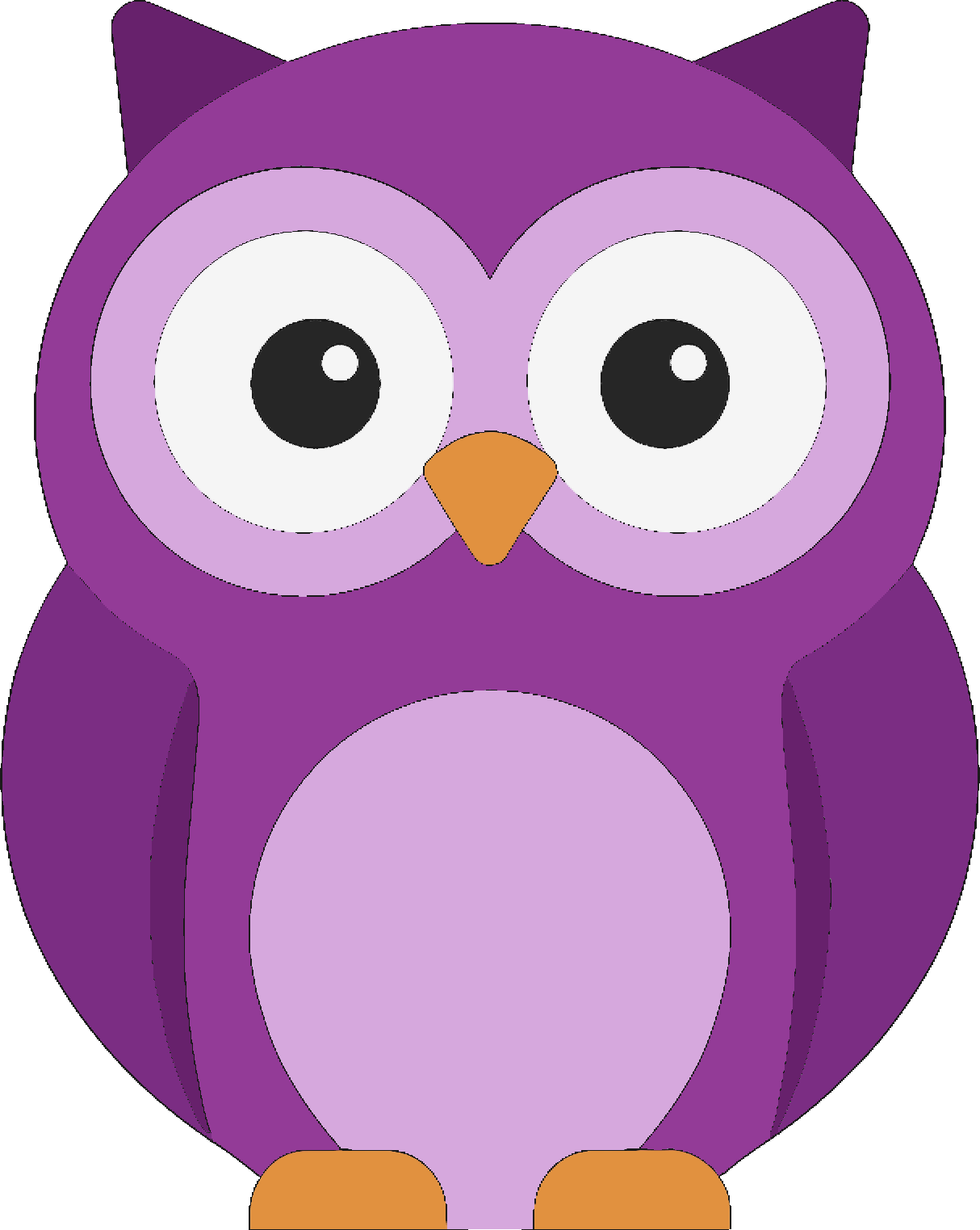}{\owlh}{owlA}};
  \node[owln] (owlB) at (2.33,-1.34) {\owl{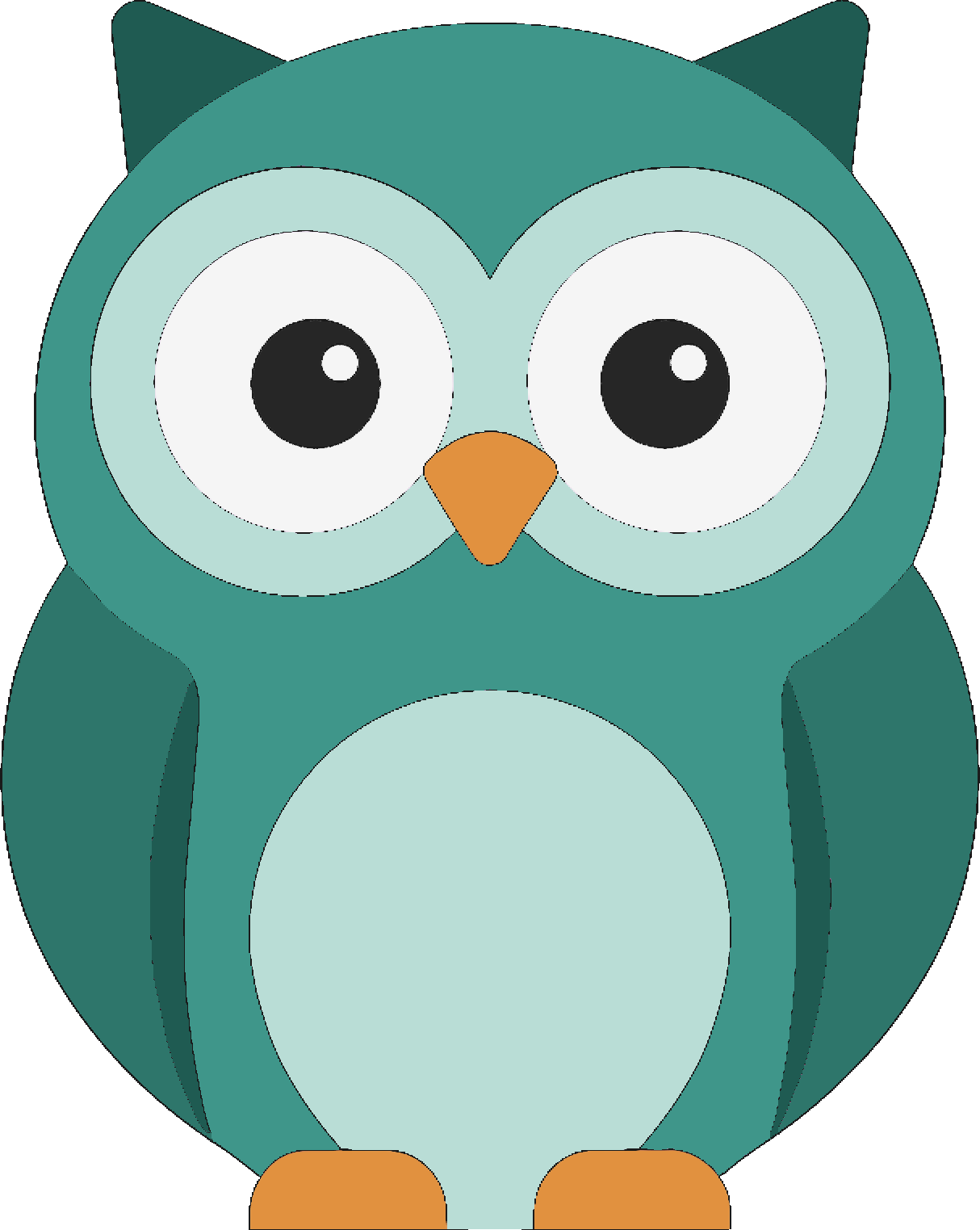}{\owlh}{owlB}};
  \node[owln] (owlC) at (2.86,-1.50) {\owl{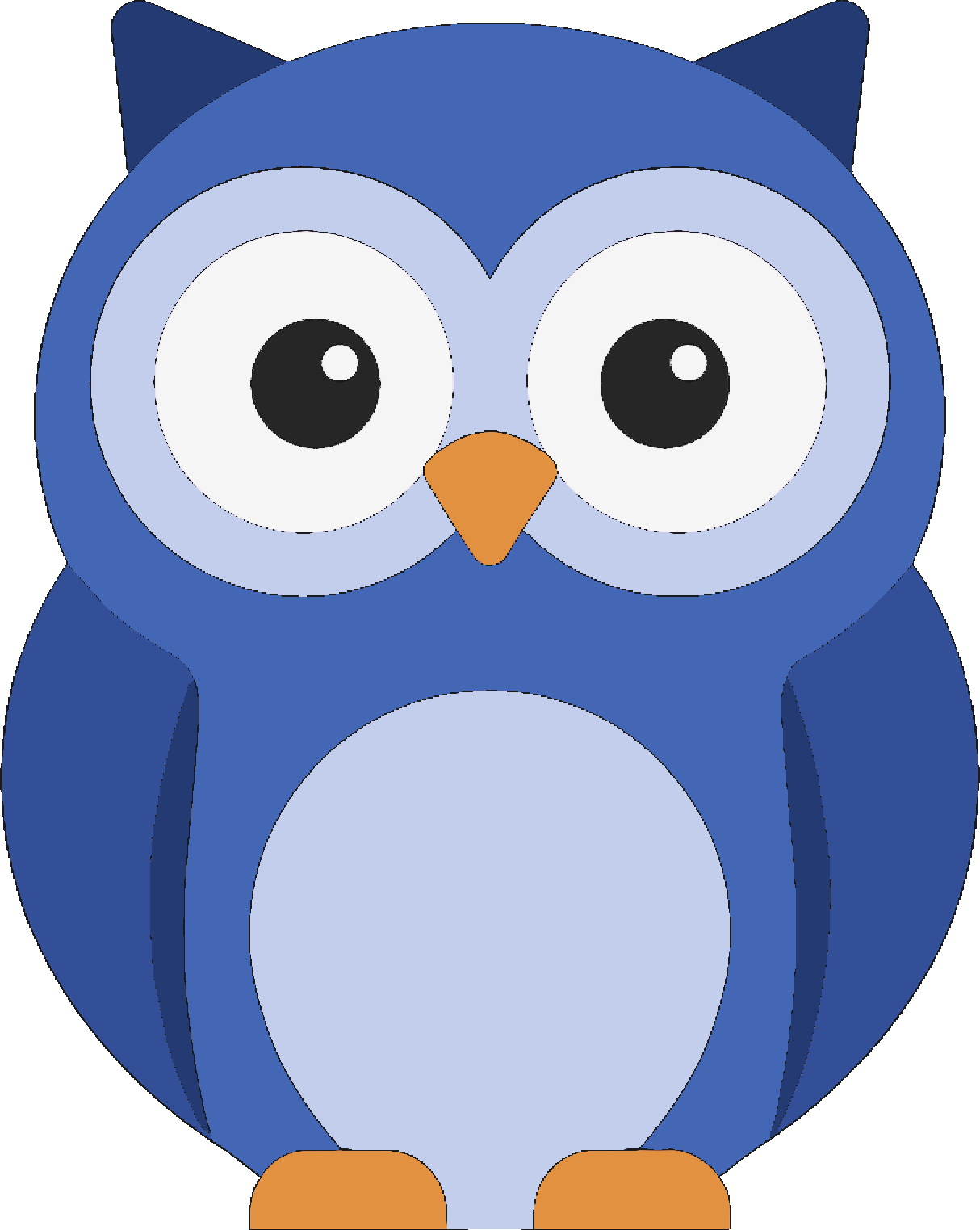}{\owlh}{owlC}};
  \node[font=\scriptsize, text=black!55] (raimlbl) at (2.33,-1.98) {\textsc{raim}};

  \draw[decorate, decoration={brace, amplitude=4pt, mirror, raise=1pt},
        black!45, line width=0.6pt] (parl.south west) -- (parl.south east);
  \node[innerbox, minimum width=2.7cm, minimum height=0.70cm, align=center,
        font=\scriptsize, text=black!70] (test) at (5.00,-1.55)
        {member competence\\$\times$ error correlation};
  \draw[-{Latex[length=1.4mm]}, line width=0.8pt, black!55]
        ($(parl.south)+(0,-0.20)$) -- (parl.south |- test.north);

  \draw[ar] (parl) -- (agg);
  \coordinate (gate) at ($(parl.east)!0.55!(agg.west)$);
  \draw[black!75, line width=0.8pt, fill=white]
        ($(gate)+(-0.18,0.20)$) -- ($(gate)+(-0.18,-0.20)$) -- (gate) -- cycle;
  \draw[black!75, line width=0.8pt, fill=white]
        ($(gate)+(0.18,0.20)$) -- ($(gate)+(0.18,-0.20)$) -- (gate) -- cycle;
  \draw[dotted, black!55, line width=0.9pt] (test.east)
        -- node[above=0pt, font=\scriptsize, align=center, text=black!70]
           {admissibility\\test} (test.east -| gate) -- (gate);

  \node[panelbox, fit=(owlA)(owlB)(owlC)(raimlbl)(lbl)(parl)(test)(agg)(agglbl), inner sep=5pt]
       (raimbox) {};

  \node[draw=black!55, fill=white, rounded corners=1pt, minimum width=1.05cm,
        minimum height=1.2cm, left=0.95cm of raimbox] (src) {};
  \foreach \yy in {0.34,0.17,0.0,-0.17,-0.34}
    \draw[black!30, line width=0.5pt] ([xshift=3pt,yshift=\yy cm]src.west)
        -- ([xshift=-3pt,yshift=\yy cm]src.east);
  \node[font=\scriptsize, align=center, above=0pt of src] (srclbl)
        {source\\{\scriptsize(context)}};
  \node[draw=black!55, fill=cStacked!6, rounded corners, minimum width=1.5cm,
        minimum height=0.62cm, align=center, font=\scriptsize, below=-0.1cm of src]
        (cand) {candidate\\claim};
  \node[font=\scriptsize, left=1pt of src] (x) {$x\!=\!$};
  \draw[ar] ($(src.east)+(0.15,0)$) -- (raimbox.west);

  \node[outn, right=1.0cm of raimbox] (out) {verdict $\hat{y}$\\[1pt]%
        {\scriptsize\textcolor{vClean}{\textsc{faithful}} \textcolor{black!45}{/}
         \textcolor{cBest}{\textsc{hallucinated}}}};
  \draw[ar, rounded corners=3pt] (agg.east) -- ++(0.55,0) |- (out.west);

  \node[cmpn, above right=0.32cm and 1.0cm of out, fill=cBest!8, draw=cBest!65] (best)
        {best single judge\\{\scriptsize(CV-selected)}};
  \node[cmpn, below right=0.32cm and 1.0cm of out, fill=cFrontier!15, draw=cFrontier] (son)
        {frontier judge\\\raisebox{+.3em}{\scriptsize\scalebox{0.6}{\logo{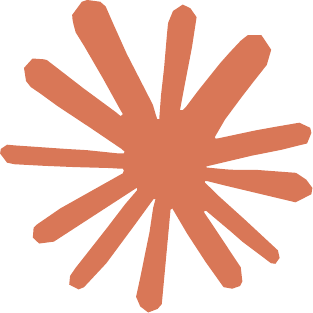}}}~Claude Sonnet};
  \draw[arg, cBest!75] (out.east) to[out=32, in=180] (best.west);
  \draw[arg, cFrontier] (out.east) to[out=-32, in=180] (son.west);
  \node[badge, draw=vClean!70, fill=vClean!12, text=vClean!50!black]
        at ($(out.east)!0.5!(best.west)+(-.70,0.40)$) {$\Delta\kappa\!\geq\!0$ on $8/8$};
  \node[badge, draw=cFrontier!70, fill=cFrontier!10, text=cFrontier!60!black]
        at ($(out.east)!0.5!(son.west)+(-.70,-0.40)$) {$\Delta\kappa\!\geq\!0$ on $5/8$};

  \begin{scope}[on background layer]
    \node[fill=cream, rounded corners=7pt, draw=black!12,
          fit=(x)(srclbl)(cand)(raimbox)(out)(best)(son), inner sep=7pt] {};
  \end{scope}
\end{tikzpicture}}
\caption{The RAIM protocol at a glance.
Ten cheap open-weight judges ($4$--$9$B) each score a candidate against its source, and a cross-fitted stacked logistic regression combines their votes into one verdict.
The same calibration set yields an admissibility test, read off the members, which identifies a regime in which aggregating them pays.
The verdict is read against the CV-best single judge \emph{of} the panel, and against a paid frontier judge the panel is meant to replace.}
\label{fig:overview}
\end{figure*}
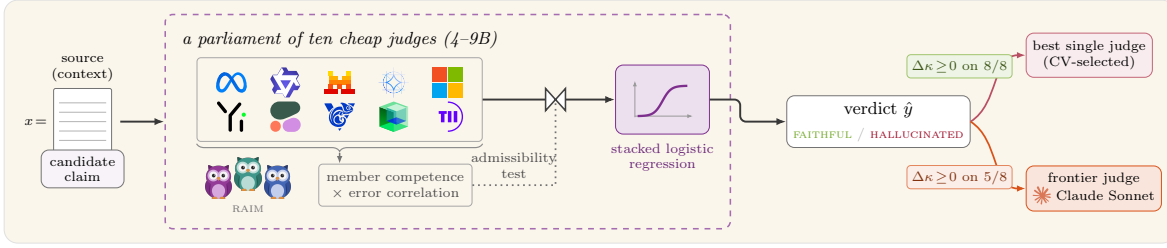

We take up both questions together, and we do so by introducing RAIM (Robust Aggregation of Inexpensive Models; Figure~\ref{fig:overview}), an aggregation scheme that couples two components.
It aggregates a panel of cheap open-weight judges by ($a$) a cross-fitted stacked meta-learner, which fits member weights \emph{jointly} and so discounts a judge echoing a stronger one, instead of weighting each by its own accuracy;
and it pairs that with ($b$) an admissibility test, read off the single members, which identifies the regime in which aggregating them pays.
Together, the two make the aggregation \emph{robust}, where a plain ensemble is only redundant.
We instantiate the scheme with ten small judges from disjoint families, and study its behaviour across eight faithfulness benchmarks and against a ladder of comparators ordered by the labelled information each of them consumes.
Combining noisy voters is itself classical~\citep{Dawid_Skene_79,Wolpert_92}, and we adopt the simplest form that expresses the joint weighting correlated errors require.
A heavier estimator would consume labels the calibration budget cannot afford, and would offer a rival explanation for whatever the panel gains; with ten fitted coefficients, instead, the gain becomes attributable to the members alone.

The result proves conditional.
A frontier judge is not, in general, replaced without cost, yet that cost is small and concentrated;
against its own best member, conversely, the panel improves on all four core grounded sets, though the sample resolves that improvement on only one of them, with no clear worsening anywhere.
The eight benchmarks therefore yield an account of \emph{which} tasks fall on which side, with the members' joint distribution of \emph{competence} and the \emph{correlation} of their errors at the centre: aggregation pays where several judges are individually capable and disagree in their mistakes, and falls back on the leader where one judge dominates or all of them are weak and err together.

\paragraph{Contributions.}
We summarise our contributions as follows.

\noindent\textbf{(i)} We introduce RAIM, an aggregation scheme coupling ($a$) a cross-fitted meta-learner that fits the members' weights jointly with ($b$) a label-cheap admissibility test which, read off the members' own verdicts rather than the panel's, identifies a regime in which aggregating them pays.
Along with it, we adopt a comparator ladder of baselines and a decision rule built on paired differences, which keeps the direction of an effect apart from the power to resolve it.%

\noindent\textbf{(ii)} We instantiate RAIM on ten judges of $4$--$9$B parameters from disjoint families over eight faithfulness benchmarks, releasing the whole chain from generation to analysis.

\noindent\textbf{(iii)} We measure what the panel trades off against a frontier judge, finding that it retains a median $93\%$ of \sonnetfull's \kp\ and falls materially behind on two of the eight benchmarks.
Against the purpose-trained detectors that are its sharpest competitors, it stays competitive on the benchmarks those detectors were trained for, and leads them elsewhere.

\noindent\textbf{(iv)} We price the substitution, including the labels the aggregator consumes, showing that inference runs at a sixty-fourth of the frontier's cost at rented cloud rates, so the operative expense is the one-time in-domain calibration and not the evaluation:
fifty labelled records suffice for the aggregator to beat a free majority vote on seven of eight sets, and a hundred are enough to approach full supervision, which places the break-even against the frontier judge between $40{,}000$ and $81{,}000$ evaluated items.

\noindent\textbf{(v)} We identify what decides the outcome, and it is not the grounded-versus-ungrounded contrast one would expect but a property of the members: how widely competence is spread across them, and how far their errors decorrelate.
We confirm this finding at the item level, over the $1{,}998$ items on which the panel's own leader errs, and we show both coordinates to be readable from the calibration set the aggregator already consumes, so the regime is known by the time the panel is fitted.

\section{The RAIM protocol}
\label{sec:setup}

Let us consider a panel made of $N_v$ members (LLMs) which, given a prompt, return one binary verdict on a candidate statement: \emph{supported} (gold$=0$) or \emph{hallucinated}/unsupported (gold$=1$).
We evaluate the panel on two tasks: \emph{grounded} faithfulness, where the candidate is judged against an explicit reference; and \emph{ungrounded} factuality, where it is judged against world knowledge alone.
Datasets are built class-balanced, and items are grouped into clusters by the source record or grounding document they derive from (Appendix~\ref{app:construction}).
In what follows, we describe the RAIM protocol in two steps: the framework first and the instantiation after.
We defer the description of the admissibility test to~\S\ref{ssec:budget}, where the quantities it reads are defined.
The judge prompt, the decoding settings, and the coverage audit are deferred to Appendix~\ref{app:coverage}, and the per-dataset item construction to Appendix~\ref{app:construction}.

\paragraph{Aggregation and the comparator ladder.}
\label{ssec:agg}
Given $N_v$ votes per query, we adopt a \emph{stacked logistic regression}~\citep{Wolpert_92} to aggregate them: a vote over the votes, taking the member verdicts as its features and the gold label as its target, so that each member earns a weight from how well its verdict predicts the truth \wrt the others.
Writing $v_i \in \{0, 0.5, 1\}$ for member $i$'s verdict on an item ($i \in \{1, \dots, N_v\}$; abstentions imputed as $0.5$, tested in Appendix~\ref{app:coverage}), the meta-learner estimates
\begin{equation}
  \mathbb{P}(y = 1 \mid v_1,\dots,v_{N_v}) = \sigma\!\left(\beta_0 + \sum_{i=1}^{N_v} \beta_i v_i\right)\ ,
  \label{eq:stacker}
  \qquad
  \sigma(z) = \frac{1}{1+e^{-z}}\ ,
\end{equation}
fitting $\beta$ by penalised maximum likelihood under an untuned $\ell_2$ penalty\footnote{Of weight $1/C$ on the squared coefficients, left at the conventional $C=1$; with only ten features the fit is insensitive to the choice, and the smallest sets (\cnn, \wice, \factscore) are where it matters most.}.
The coefficients $\beta_i$ are what a vote weighted by each member's own accuracy cannot express: they are fitted \emph{jointly}, so a member that merely echoes a stronger one is discounted rather than counted twice, and a member that is informative but anti-correlated with gold can take a negative weight.
To avoid leakage, the stacker is fitted by 5-fold cross-fitting, so that every item's stacked prediction is produced out-of-fold.
A verdict is read at the conventional $0.5$ threshold, which also breaks the unweighted majority's ties at $N_v=10$.
Fitting $\beta$, however, requires gold labels, a cost we measure in Appendix~\ref{app:budget} and price in~\S\ref{sec:frontier}.

We compare the stacked panel against a ladder of comparators, ordered by the labelled information each of them consumes.
The unweighted majority panel, the average member, and the unsupervised Dawid--Skene aggregator
need no labels and are fully deployable; the best-on-average single is one global label-based selection deployed everywhere; the cross-validated (CV) best single, selected on a development split, and the stacked panel are both supervised and both evaluated strictly out-of-fold; the oracle best single, selecting the argmax on the very test set it is scored against, is the most generous yet undeployable comparator; and above them all sits the one the panel is meant to replace rather than beat, a proprietary frontier judge, named and read in~\S\ref{sec:frontier}.

\paragraph{Metrics and the decision rule.}
\label{ssec:metrics}
Our primary metric is Cohen's \kp~\citep{Cohen_60}, which rescales the observed agreement $p_o$ with the gold label by the agreement $p_e$ that a random sample would produce, so that only agreement above chance is credited: \(\kp = (p_o - p_e)/(1 - p_e)\).
We accompany it with balanced accuracy (\bacc), \ie the mean of the two per-class recalls (on our class-balanced sets coinciding with plain accuracy), which accounts for an abstention from the opposite side: an abstained item is dropped from \kp{}, yet charged in \bacc\ as an error against its own class; Appendix~\ref{app:coverage} audits the denominator asymmetry this creates between an abstaining baseline and the always-imputing panel.
All quantities carry $95\%$ paired cluster-bootstrap percentile intervals over $B=2000$ resamples of the clustering unit, so that co-clustered instances always resample together, and every central estimate is followed by its $[\text{lower}, \text{upper}]$ interval.
When comparing two methods, every verdict is read from the paired difference of their scores: within each draw, both sides are recomputed on the same resampled clusters, so that scores are always compared on the very same items; this is materially sharper than assessing if the marginal intervals overlap, since the two scores rise and fall together across draws\footnote{On \medhallu the stacked panel spans $[0.695, 0.756]$ and its best single judge $[0.658, 0.727]$, largely overlapping, while their paired difference is $+0.032$ $[+0.004, +0.060]$.}.

On that basis we define three verdicts: a \emph{win} requires the stacked \kp\ to significantly beat the CV-best single, \ie the lower bound of the paired difference to exceed zero; a \emph{clean win} additionally requires it to beat the oracle single; a \emph{tie} is a difference straddling zero, and admits no intermediate grade, though where a tie's interval comes close to excluding zero we report its one-sided $p$.
Every paired difference we report --that one included, and equally those against the average member, the label-free aggregators, or the frontier judge of~\S\ref{sec:frontier}-- is described in one four-way form: an \emph{improvement} or a \emph{worsening}, \emph{clear} where the $95\%$ paired interval, uncorrected for multiplicity, excludes zero and \emph{unresolved} where it does not (Appendix~\ref{app:coverage} gives Holm-adjusted $p$-values for the best-single contrasts).

\paragraph{Benchmarks and judges.}
\label{ssec:data}
We run RAIM on eight faithfulness benchmarks\footnote{
  Question-answering grounded: \medhallu and \ragtruth; claim-verification grounded: \wice, \xsum, \cnn, and \expertqa (from \aggrefact); ungrounded: \factscore and \truthfulqa.
}: six \emph{grounded} and two \emph{ungrounded} (Appendix~\ref{app:perdataset}, Table~\ref{tab:datasets}).
Two of the six grounded sets are held back from the primary tally on measurability grounds: \cnn is too small to power the paired contrast ($n=114$), and \expertqa is a low-headroom task, with no published method on the \aggrefact leaderboard~\citep{Tang_Laban_Durrett_24} exceeding a \bacc\ of ${\approx}\,61\%$ (Appendix~\ref{app:extbase}, Table~\ref{tab:extbase}).
The primary denominator is therefore the \emph{core grounded} four --\medhallu, \ragtruth, \xsum, and \wice-- with all six grounded sets as a secondary denominator and the suite of eight as the widest.

\label{ssec:panel}
Over those benchmarks, we set $N_v = 10$, instantiating the panel with ten open-weight instruction-tuned judges of $4$--$9$B parameters drawn from disjoint model families to diversify failure modes: \llama, \qwen, \mistral, \gemma, \phimini, \yi, \commandr, \glm, \granite, and \falcon being the family tag by which we refer to each throughout (for details see Appendix~\ref{app:coverage}, Table~\ref{tab:roster}).
A member returning no parseable verdict is recorded as abstaining.
The prompt supplies the task definition, and takes one of two frames: the rubric we label \defon\ on the four question-answering sets, scaffolding the tuple (question, evidence, candidate) on \medhallu and \ragtruth and dropping the evidence block on the ungrounded \factscore and \truthfulqa; and a MiniCheck-style~\citep{Tang_Laban_Durrett_24} \emph{claim-support} frame, pairing (document, claim), on the four \aggrefact claim-verification sets (\wice, \xsum, \cnn, \expertqa), whose structure requires it (Appendix~\ref{app:extbase}).
The frame is fixed per dataset and identical across all ten members and the frontier judge.

\section{Core results}
\label{sec:results}

We report the study's core results in three steps: first the panel measured against the other ways of combining the same ten votes, then against the best single judge \emph{inside} it, and finally the member-level account of when either contrast comes out in the panel's favour (\S\ref{ssec:mechanism}); the comparison with the frontier judge is deferred to~\S\ref{sec:frontier}.
Table~\ref{tab:master} carries all of it on one comparator ladder, whose columns are ordered by the labelled information each baseline consumes: from the label-free aggregators a deployer can always run, through the one globally best member, to the per-dataset best single and the frontier judge.
Reading a row left to right is therefore reading what each increment of supervision buys.
The full per-method ladders and the rescue and aggregator contrasts are in Appendix~\ref{app:perdataset} (Table~\ref{tab:aggregators}).

\begin{table*}[tp]
\centering\scriptsize
\setlength{\tabcolsep}{2pt}
\caption{Comparator ladder over the eight benchmarks.
Columns are ordered by the labelled information each comparator consumes. \kp\ is given as a decimal and \bacc\ as a percentage.}
\label{tab:master}
\resizebox{\textwidth}{!}{%
\begin{tabular}{l r rrr rr r rr rr r r r r}
\toprule
 & & \multicolumn{3}{c}{\emph{label-free}} & \multicolumn{3}{c}{\emph{single judge, label-selected}} & \multicolumn{2}{c}{\emph{stacked panel}} & \multicolumn{2}{c}{\emph{frontier} (\sonnet)} & \multicolumn{4}{c}{\emph{paired} $\Delta$ \emph{(panel $-$ comparator)}} \\
\cmidrule(lr){3-5}\cmidrule(lr){6-8}\cmidrule(lr){9-10}\cmidrule(lr){11-12}\cmidrule(lr){13-16}
Dataset & $n$ & avg.\ mem. & unw. & D--S & best (avg) & best (CV) & oracle & \kp & \bacc & \kp & \bacc & avg (\kp) & best (\kp) & \sonnet (\kp) & \sonnet (\bacc) \\
\midrule
\multicolumn{16}{l}{\emph{Core grounded}} \\
\medhallu & 2000 & 0.590 & 0.708 & 0.716 & 0.656 & 0.693$^\dagger$ & 0.693 & 0.725 & 86.3 & 0.761 & 88.1 & $\mathbf{+0.135}$ & $\mathbf{+0.032}$$^\ddagger$ & $\mathbf{-0.036}$ & $\mathbf{-1.79}$ \\
\wice & 222 & 0.385 & 0.449 & 0.486 & 0.547 & 0.484$^\dagger$ & 0.547 & \underline{0.540} & 77.0 & 0.548 & 77.4 & $\mathbf{+0.155}$ & $+0.056$ & $-0.008$ & $-0.39$ \\
\ragtruth & 1362 & 0.181 & 0.179 & 0.325 & 0.399 & 0.399 & 0.399 & 0.409 & 70.4 & 0.601 & 80.0 & $\mathbf{+0.228}$ & $+0.010$ & $\mathbf{-0.192}$ & $\mathbf{-9.62}$ \\
\xsum & 546 & 0.312 & 0.395 & 0.420 & 0.377 & 0.400$^\dagger$ & 0.437 & \underline{0.437} & 71.9 & 0.418 & 70.9 & $\mathbf{+0.125}$ & $+0.037$ & $+0.019$ & $+0.96$ \\
\multicolumn{16}{l}{\emph{Grounded (secondary)}} \\
\cnn & 114 & 0.088 & 0.070 & 0.088 & 0.298 & 0.298 & 0.298 & \underline{0.349} & 67.5 & 0.469 & 73.5 & $\mathbf{+0.261}$ & $+0.051$ & $-0.120$ & $-6.05$ \\
\expertqa & 1462 & 0.154 & 0.146 & 0.205 & 0.217 & 0.240 & 0.240 & \underline{0.235} & 61.8 & 0.165 & 58.2 & $\mathbf{+0.082}$ & $-0.005$ & $\mathbf{+0.070}$ & $\mathbf{+3.50}$ \\
\multicolumn{16}{l}{\emph{Ungrounded (contrast)}} \\
\factscore & 330 & 0.259 & 0.485 & 0.515 & 0.444 & 0.444$^\dagger$ & 0.444 & \underline{0.503} & 75.2 & 0.552 & 77.6 & $\mathbf{+0.244}$ & $+0.059$ & $-0.049$ & $-2.45$ \\
\truthfulqa & 1634 & 0.274 & 0.374 & 0.409 & 0.471 & 0.471$^\dagger$ & 0.471 & 0.472 & 73.6 & 0.623 & 81.1 & $\mathbf{+0.198}$ & $+0.000$ & $\mathbf{-0.151}$ & $\mathbf{-7.56}$ \\
\midrule
\emph{count} $\uparrow$/$\downarrow$ &  &  &  &  &  &  &  &  &  &  &  & \textbf{8}\,(--)\,/\,--\,(--) & \textbf{1}\,(6)\,/\,--\,(1) & \textbf{1}\,(1)\,/\,\textbf{3}\,(3) & \textbf{1}\,(1)\,/\,\textbf{3}\,(3) \\
\bottomrule
\end{tabular}}
\tablelegend{
\emph{avg.\ mem.}: average member; \emph{unw.}: unweighted majority; \emph{D--S}: Dawid--Skene.
The four rightmost columns are paired cluster-bootstrap differences on the same resampled clusters ($B=2000$), \textbf{bold} where the $95\%$ interval excludes zero (cf.\ Figure~\ref{fig:regimeforest}(b)); the \emph{count} line tallies them as improvement ($\uparrow$) or worsening ($\downarrow$), as `\textbf{clear}\,(unresolved)', read before rounding.
$^\ddagger$: beats the oracle too; \underline{underline}: the frontier is not clearly ahead; $^\dagger$: an under-covering CV-best baseline (Appendix~\ref{ssec:confound}).}
\end{table*}

\begin{figure}[tp]
\centering
\includegraphics[width=.9\linewidth]{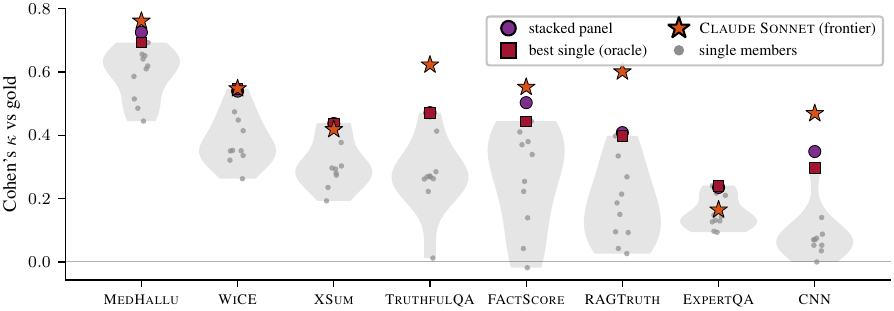}
\caption{
  Single-member competence per dataset (Cohen's \kp\ vs gold).
  Violins and strips carry the ten judges' \kp, with the stacked panel ($\bullet$), the oracle best single ($\blacksquare$), and the frontier judge \sonnet ($\star$; \S\ref{sec:frontier}) overlaid.
  Datasets are ordered by descending mean single-member competence.
  }
\label{fig:competence}
\end{figure}

\subsection{The panel against its own naive vote}
\label{ssec:rescue}
Which panel is on offer must be established first, and the answer is not the obvious one: combining the ten votes naively gives a panel that has to be rescued from itself.
Wherever competence is concentrated in one leader or spread unevenly across a field with actively harmful members, the unweighted majority panel sits \emph{significantly below} the best single judge --by $0.094$ to $0.228$ \kp\ on \ragtruth, \truthfulqa, \cnn, and \expertqa-- because the weaker, mutually redundant members outvote the few capable ones.
Stacking reverses this: reading the \emph{unw.} and \emph{stacked} columns of Table~\ref{tab:master} against \emph{best (CV)}, the panel is no longer clearly below the best single judge on any of the four, and the paired delta \mbox{(stacked $-$ unweighted)} is significantly positive on each (up to $+0.279$ on \cnn) and borderline on \medhallu (intervals in Appendix~\ref{app:perdataset}, Table~\ref{tab:aggregators}).
The effect is, by contrast, small or absent where the naive vote is not clearly below the best single judge, \ie where competence is spread (\medhallu, \wice, \factscore) or redundant (\xsum); the size of the rescue thus depends on how heterogeneous the members are.
Weighting each vote by that member's own agreement with gold (each in isolation) gives the graded form of the same point: both accuracy- and log-odds-weighted voting sit below the stacker on every dataset (Appendix~\ref{app:perdataset}).

\paragraph{Comparison with an unsupervised aggregator.}
\label{ssec:otheragg}
A sharper test than the naive vote is another \emph{aggregator} over the same ten votes, and the natural choice is the model of \citet{Dawid_Skene_79} (\emph{D--S}), which estimates each member's reliability without gold and so isolates what the labels buy (Appendix~\ref{app:related}).
The stacked panel is never significantly beaten by it (Appendix~\ref{app:perdataset}), and the delta \mbox{(stacked $-$ DS)} offers a clear improvement on \ragtruth, \truthfulqa, and \cnn, and straddles zero elsewhere;
D--S itself falls significantly below the CV-best single wherever the best member must be picked out of a weak or correlated field.
This is the competence/correlation account (\S\ref{ssec:mechanism} below) seen from the aggregator side: where errors decorrelate and competence is spread, an unsupervised aggregator suffices; if one judge must be recovered, only supervised re-weighting does.

\subsection{The panel against the best single judge}
\label{ssec:beating}
Against the fair (CV-best) baseline, the paired difference is an improvement on all four core grounded datasets, and on one of them it is clear:
\medhallu offers a delta \mbox{(stacked $-$ best single) $= +0.032$} clear against both the CV-best and the oracle (\granite) --the suite's one \emph{clean win} in the sense of~\S\ref{ssec:metrics}-- and still clear on the items the baseline commits on ($+0.031$ $[+0.003, +0.058]$); on the other three, the improvement stands but the sample does not resolve it.
A Holm correction~\citep{Holm_79} over the four core deltas (cf.~\S\ref{ssec:data}) leaves that win significant (adjusted one-sided $p=0.048$), whereas over all eight it would not ($p=0.096$, see also Appendix~\ref{app:coverage}).
With nine competent members under the claim-support frame ($\kp\geq0.30$;
cf.\ Figure~\ref{fig:regimeforest}(a)) and the best single being already strong, \wice offers an unresolved improvement ($+0.056$) at $n=222$.
On \ragtruth, conversely, one member (\gemma) dominates a weak field and is reliably found by cross-validation, leaving almost no headroom ($+0.010$).
Finally, on \xsum the members carry moderate, redundant competence (four at $\kp\geq0.30$) and the edge ($+0.037$, one-sided $p=0.045$) is the suite's closest unresolved improvement.
The two grounded sets held back are unresolved as well (\cnn as an improvement and \expertqa as a worsening), and are reported in Appendix~\ref{app:perdataset}.
This fixes the scope of the internal contest: the difference favours the panel on seven out of eight datasets, yet is clear on one core set of four.%

\paragraph{Comparing baselines needing no per-dataset labels.}
\label{ssec:deployment}
Identifying the best single, however, requires labels a deployer might not have; against the baselines needing no such per-dataset selection (Table~\ref{tab:master}, leftmost block) the panel is strong wherever there is signal: it improves on the label-free average member by $+0.082$ to $+0.261$ \kp, clearly on every dataset in the suite, and is nowhere clearly below the single best-on-average judge (\gemma).
The one-of-four verdict above is thus the same panel read against a baseline that has already spent those labels, and its position on that ladder is what~\S\ref{sec:frontier} prices.

\subsection{The competence/correlation account}
\label{ssec:mechanism}

The pattern of paired differences above is not arbitrary:
it is ordered by two member-level quantities recovered from the calibration set the aggregator already requires, namely the \emph{distribution of competence} --how many members are individually capable, rather than how good the best one is-- and the \emph{correlation of errors} --- whether the members, when wrong, are wrong on the same items, which we summarise by their mean pairwise error-correlation $\bar\phi$.
The ordering, however, does not rest on eight dataset points alone: restricting to the items on which the best single judge errs isolates the recovery the \emph{other} members carry.
Under this framing, the panel recovers the item more often precisely when a larger fraction of members is individually correct on it (logistic regression with dataset fixed effects, coefficient $+4.67$; point-biserial $r=+0.41$, $p<10^{-20}$, positive within all eight datasets; Appendix~\ref{app:predictor}).
Distributed, decorrelated competence is thus what the panel harnesses, and what the stacker's own weights reward in place of average accuracy (Appendix~\ref{app:perdataset}, Figure~\ref{fig:weights}).

\begin{figure}[tp]
\centering
\includegraphics[scale=\regimeforestscale]{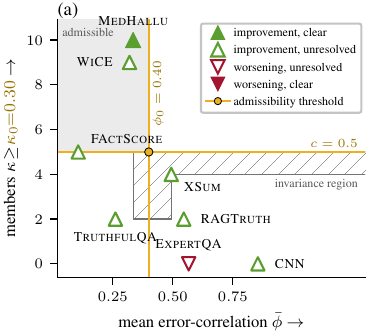}%
\hspace{\regimeforestgap}%
\includegraphics[scale=\regimeforestscale]{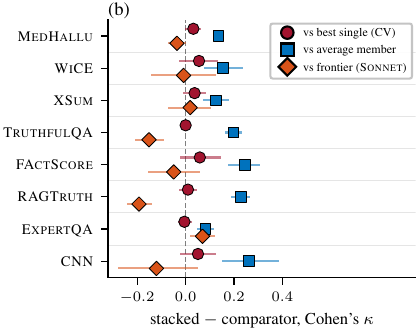}
\caption{The regime ordering.
\textbf{(a)} The eight datasets (markers) scattered by the members' mean pairwise error-correlation $\bar\phi$ ($x$-axis) against the number of members at $\kp\geq\kappa_0=0.30$ ($y$-axis).
The gold lines are the thresholds $\phi_0 = 0.40$ and $c = \sfrac12$ (five members), and bound the admissible region (shaded quadrant);
the hatched area is the invariance region, the positions of the gold point at which no dataset changes side, unbounded to the right.
Varying $\kappa_0$ moves the points rather than an edge (Appendix~\ref{app:budget}).
\textbf{(b)} The panel's paired delta against the CV-best single ($\bullet$), the average member ($\blacksquare$), and the frontier judge ($\blackdiamond$), reported per dataset with $95\%$ paired cluster-bootstrap intervals.}
\label{fig:regimeforest}
\end{figure}

The two coordinates order the suite, and Figure~\ref{fig:regimeforest}(a) places the eight datasets in the plane they span (qualitatively, eight points being too few to resolve either coordinate statistically, cf.\ Appendix~\ref{app:predictor}, Table~\ref{tab:predictor}).
The panel's improvement over the best single is largest where competence is distributed over several capable members whose errors decorrelate, shrinks where that competence is moderate and mutually redundant, and vanishes where one model dominates a weak field.
The same ordering governs the frontier comparison.
Read against all three comparators at once (Figure~\ref{fig:regimeforest}(b)), the difference over the \emph{best} single narrows to zero as one model comes to dominate, and the difference against the \emph{frontier} judge worsens along with it (\S\ref{sec:frontier} below), while the difference over the \emph{average} member does neither, staying uniformly large wherever there is signal.
The regime in which a panel improves on its own leader is thus also one in which the frontier is not materially ahead.
The frontier has to be kept only in the dominant-leader regime the test rejects, which is why we answer the two questions together.
Interestingly, it is not the regime the literature would predict:
grounded-versus-ungrounded is numerically present but mechanistically empty, the two ungrounded sets falling in \emph{opposite} regimes, and what moves with the task is instead the identity of the best single judge --- which is what a panel is meant to supply, and what fails to transfer (Appendices~\ref{app:predictor} and~\ref{app:transfer}).

\paragraph{The admissibility test.}
\label{ssec:budget}
The account enters the method only if a deployer can tell, on a new task, which side of it they are on:
this is the second component of RAIM (\S\ref{sec:intro}), which, along with the stacker's down-weighting of redundant members, makes the scheme \textit{robust}.
The test reads the two coordinates off the calibration set, and admits a task when enough members are competent and their errors are weakly correlated.
Formally, given a competence bar $\kappa_0$, a share $c$, and a correlation bar $\phi_0$, a task is \emph{admissible} when at least a share $c$ of the members reach $\kp\geq\kappa_0$ and $\bar\phi\leq\phi_0$;
in the plane of Figure~\ref{fig:regimeforest}(a), at fixed $\kappa_0$, this is a quadrant, and, in what follows, we adopt $(\kappa_0, c, \phi_0) = (0.30, \tfrac12, 0.40)$ as an operating point, resulting in \medhallu, \wice, and \factscore being admitted.
Tightening any of the three admits fewer tasks, and where to stop depends on what a deployer loses by dropping a frontier judge that was needed, against paying for one that was not;
moreover, with just eight datasets, the calls change only where a threshold crosses a dataset's own coordinate (hatched region of Figure~\ref{fig:regimeforest}(a); Appendix~\ref{app:budget}).
What holds at every operating point is the direction of the test:
over a grid of shares and correlation bars, and on $1{,}888$ (sub-)panels drawn from the same ten members, the admitted panels are more often within reach of the frontier judge, and more often above their own best member, than the rejected ones (Appendix~\ref{app:budget}).
Therefore, stackers on admissible tasks are expected to improve on the CV-best member;
no task failing the test shows a clear gain over that member, and on dominant-leader ones, the frontier stays materially ahead (\S\ref{sec:frontier}).

The two coordinates are not, however, equally costly.
The correlation one is free:
two binary verdicts against a common gold label agree exactly when both are equally right or wrong, so the members' mean pairwise \emph{agreement} equals that of their error indicators and needs no gold (only its normalisation into $\bar\phi$ does), and it orders the suite like $\bar\phi$ (Spearman $\rho = +0.93$).
The competence one has no assumption-free unsupervised counterpart: a Dawid--Skene fit to the same votes (\S\ref{ssec:otheragg}) rates all ten members competent where, \eg, on \cnn and \expertqa none clears $\kp = 0.30$.
On calibration samples, the regime call is reproduced $85.5\%$ of the time from fifty labelled records and $91.9\%$ from a hundred, the budgets at which the aggregator itself becomes worth fitting, so the diagnostic and the calibration are one purchase, whereas naming the best member from a hundred succeeds only $69\%$ of the time (Appendix~\ref{app:budget}).
Notably, the error-correlation is largely inherited from pretraining rather than induced by alignment (Appendix~\ref{app:base}), and on four of the eight datasets our cheap panel still carries more independent votes than \citeauthor{Kohli_26}'s~\citep{Kohli_26} nine frontier judges reach on any of theirs (Appendix~\ref{app:neff}).

\section{Standing in for a frontier judge}
\label{sec:frontier}

We now ask what is lost and what is gained by running the panel instead of paying for a frontier judge.
We adopt \sonnetfull~\citep{Anthropic_25} (\mdname{claude-sonnet-4-6}; temperature $0$) as frontier judge, and difference it against the panel item by item under the same paired cluster bootstrap (\S\ref{ssec:metrics}).

\paragraph{The quality cost.}
Read as eight paired differences (the two rightmost columns of Table~\ref{tab:master}; intervals in Appendix~\ref{app:scaling}, Table~\ref{tab:frontier}), the substitution offers an improvement on two datasets (one of which is \textit{clear}) and worsens on the remaining six (three \textit{unresolved} and three clear, one of which is negligible).
The clear improvement is \expertqa ($\Delta\kp = +0.070$ $[+0.024, +0.117]$), the set on which \sonnet is weakest by far ($\kp = 0.165$, the next-weakest being $0.418$), below even the panel's best single member (Figure~\ref{fig:competence}), a shortfall not ascribable to the prompt frame, the better of the two we ran for \sonnet (Appendix~\ref{app:extbase}).
\xsum's improvement and \wice's, \factscore's, and \cnn's worsenings are all unresolved, leaving those four statistically too close to the frontier to claim a result in either direction.
Of the three clear worsenings, \medhallu's is too small to matter in deployment ($\Delta\kp = -0.036$ $[-0.069, -0.003]$, \ie $1.8$ points of balanced accuracy on a task the panel already scores at $86.3$), leaving only \ragtruth ($-0.192$) and \truthfulqa ($-0.151$) materially adverse: precisely the two on which one strong judge already captures the task, so the panel has nothing worth aggregating and a frontier model is simply a better single judge.
The results are comparable when read on \bacc\ (which charges every abstention on the full denominator): the panel gives up $2.9$ points on average ($73.0$ against $75.9$) and at most $9.6$ (\ragtruth), while gaining $3.5$ on \expertqa and $1.0$ on \xsum.
A frontier judge therefore has to be kept in one regime only: the panel holds a median $93\%$ of the frontier's \kp\ and is clearly and materially behind on two of eight, and the account of~\S\ref{ssec:mechanism} names those two as the dominant-leader sets --- a reading available from the members themselves, at the same labelling cost the aggregator already requires (Appendix~\ref{app:budget}).

\paragraph{Cost and deployment profile.}
Table~\ref{tab:profile} prices that deficit against what a deployer would weigh.
The ten members together run at \$$0.039$ per thousand items against \$$2.52$ for the \sonnet \textsc{api}, about $1.55\%$ of the price (Appendix~\ref{app:cost} derives both; Appendix~\ref{app:limits} bounds them).
Moreover, the panel runs on premises, so no evaluated text leaves the deployer's infrastructure and no rate limit or model deprecation enters the evaluation loop, a solution the escalation route cannot offer (Appendix~\ref{app:related}).
Inference is not, however, where the panel's money goes.
The stacker needs in-domain labels, so the break-even is a volume rather than a rate, and how large a volume turns on how many labels the aggregator requires, which we measure at fifty to a hundred labelled records (Appendix~\ref{app:budget}, Figure~\ref{fig:budget}(b)), the same operating range as the diagnostic of~\S\ref{ssec:budget}.
Priced at \$$1$ per label, and two labels to a record on the paired sets, that is an outlay of \$$100$ to \$$200$, which the panel's cheaper inference repays between some $40{,}000$ and $81{,}000$ evaluated items --- a large yet sustainable volume.

\begin{table*}[tp]
\centering\footnotesize
\caption{Quality, cost and applicability of each route. \bacc\ is evaluated over three nested groups: the four \aggrefact\ sets, the six grounded sets and the full suite.
Published cells (top) are from the literature (Tables~\ref{tab:extbase} and~\ref{tab:extf1}; \kp\ in Table~\ref{tab:ptjudge}).}
\label{tab:profile}
\setlength{\tabcolsep}{4pt}
\resizebox{\textwidth}{!}{%
\begin{tabular}{l l ccc rl cc}
\toprule
 & & \multicolumn{3}{c}{\emph{quality} (\bacc) \emph{by scope}} & \multicolumn{2}{c}{\emph{cost} (\textsc{usd})} & \multicolumn{2}{c}{\emph{scope}} \\
\cmidrule(lr){3-5}\cmidrule(lr){6-7}\cmidrule(lr){8-9}
Route & size & \aggrefact-4 & grounded 6 & suite 8 & per 1k & one-time & open & ungr. \\
\midrule
\multicolumn{9}{l}{\emph{purpose-trained and prompted comparators}} \\
\mdname{Bespoke-MiniCheck-7B}~{\scriptsize (leaderboard)} & 7B & $71.4$ & --- & \emph{n/a} & $0.004^{\P}$ & fine-tune & \yes & \no \\
\mdname{MiniCheck-Flan-T5-L}~{\scriptsize\citep{Tang_Laban_Durrett_24}} & 0.8B & $68.8$ $(68.4^{\dagger})$ & $66.6^{\dagger}$ & \emph{n/a} & $0.004^{\P}$ & fine-tune & \yes & \no \\
\mdname{Granite Guardian 3.3}~{\scriptsize (leaderboard)} & 8B & $69.5$ & --- & --- & $0.004^{\P}$ & fine-tune & \yes & \partialy \\
\mdname{Prometheus-2}~{\scriptsize\citep{Kim_Suk_Longpre_etal_24}} & 7B & $59.3^{\dagger}$ & $60.0^{\dagger}$ & $60.4^{\dagger}$ & $0.004^{\P}$ & fine-tune & \yes & \yes \\
\mdname{GPT-4o}, prompted~{\scriptsize (leaderboard)} & --- & $70.8$ & --- & --- & $1.93$ & \emph{none} & \no & \yes \\
\midrule
\multicolumn{9}{l}{\emph{measured here, one protocol across all eight sets}} \\
average cheap open judge & $4$--$9$B & $59.6$ & $62.4$ & $62.5$ & $0.004$ & \emph{none} & \yes & \yes \\
best single member (CV-selected) & $4$--$9$B & $67.7$ & $68.5$ & $69.3$ & $0.004$ & \$100--\$200 & \yes & \yes \\
larger open judge (\qwenbig{32B}) & 32B & $69.4$ & $72.1$ & $72.6$ & $0.008$ & \emph{none} & \yes & \yes \\
\textbf{stacked cheap panel (this work)} & $10\times4$--$9$B & $\mathbf{69.5}$ & $\mathbf{72.5}$ & $\mathbf{73.0}$ & $0.039$ & \$100--\$200 & \yes & \yes \\
frontier \textsc{api} judge (\sonnet) & --- & $70.0$ & $74.7$ & $75.9$ & $2.52$ & \emph{none} & \no & \yes \\
\bottomrule
\end{tabular}}
\tablelegend{
$^{\dagger}$:~our rerun under the system's released weights;
\emph{---}:~no data;
\emph{n/a}:~scope the judge cannot be run on;
\emph{ungr.}:~applicability to ungrounded factuality (\yes~yes, \partialy~partly, \no~no);
\emph{cost}:~inference per $1{,}000$ items (\textsc{api} or cloud \textsc{gpu} rates, Appendix~\ref{app:cost}) and one-time outlay on a new domain (\$$1$ per label);
$^{\P}$:~estimated by comparison.}
\end{table*}

\paragraph{Comparison with the \aggrefact leaderboard.}
Four of our eight sets are subsampled from the \aggrefact test split and read in balanced accuracy, the metric that leaderboard reports;
Appendix~\ref{app:extbase} sets out what licenses reading our numbers beside the published ones, showing an agreement within about one point of balanced accuracy, so gaps under three points are not informative.
On those four, the panel reaches $69.5$, drawing level with \mdname{Granite Guardian 3.3} and sitting inside the agreement band of the leaderboard's top system (\mdname{Bespoke-MiniCheck-7B}, $71.4$) and of \mdname{GPT-4o} ($70.8$); all this from a field whose best member scores $67.7$ (average member $59.6$).
This is, however, the comparison least favourable to the panel, with specialists being scored on the datasets they were built for: run at their documented operating points across our eight datasets, the ordering materially reverses (Table~\ref{tab:profile}; per-dataset in Appendix~\ref{app:ptj}); and
the strongest specialist we reproduce, \mdname{MiniCheck-Flan-T5-L}, scores $66.6$ against the panel's $72.5$ across the six grounded sets, $63.1$ against $86.3$ on \medhallu alone, and cannot be evaluated on the full suite as its interface requires a document the two ungrounded sets cannot supply.
The evaluation-tuned judges, which do run everywhere, are weak throughout (\mdname{Prometheus-2} at $60.4$ over the suite, against our $73.0$).
A cheap panel does not beat a trained detector on the detector's own ground, the two serving different ends: a specialist gives the best quality on the distribution it was trained for, while a generalist panel gives comparable verdicts that keep working if the domain moves and the reference disappears.

\paragraph{Outside \aggrefact.}
Two further benchmarks admit a published comparison once the metric is matched, \medhallu and \ragtruth being reported in hallucinated-class F1 (Appendix~\ref{app:extbase}, Table~\ref{tab:extf1}).
On \medhallu the stacked panel reaches $86.2$, drawing level with the strongest published open judge (\qwenbig{14B}, $85.2$) and with \mdname{GPT-4o-mini} ($84.1$), while resting within two points of the published \mdname{GPT-4o} ($87.7$).
Three of the published detectors are \emph{members of our own panel}, which gives a second harness calibration (Appendix~\ref{app:extbase}).
On \ragtruth, however, the published numbers sit on that corpus's test split, at a hallucinated rate of $0.178$ against the $0.5$ our construction imposes, and hallucinated-class F1 moves with prevalence, so our rows are not comparable in level with them.
The anchors themselves show the pattern of this paper, with the strongest prompt-only system (\mdname{GPT-4-turbo}, $45.6$) falling far below the three detectors fine-tuned on the corpus ($68.2$ to $74.8$).
On \medhallu prevalence is matched; however, we charge abstentions as missed detections and the two sides are not scored on the same items, so results compare levels and license no ranking.

\paragraph{Open-judge scaling.}
Size alone does not close the remaining gap either: a larger \emph{open} judge does not improve monotonically with scale, \qwenbig{32B} being the better of the two on four sets, and \qwenbig{72B} trails \sonnet on four of the eight, drawing level mostly where the frontier is weakest (Appendix~\ref{app:scaling}).
Within the one open family we scaled, parameter count is therefore not what separates an open judge from a frontier model.
A single \qwenbig{32B} is nonetheless a close rival on average ($72.6$ against $73.0$ suite \bacc, Table~\ref{tab:profile}), yet nothing tells a deployer in advance where it will hold, whereas the panel's regime is read from its own members.

\section{Conclusion}
\label{sec:conclusion}
In this work, we studied whether a panel of cheap open-weight judges can stand in for a frontier one, what the substitution sacrifices, and when it is worth making.
We built that panel from ten $4$--$9$B members of uneven and, on any new task, unknown competence:
none of them clears our competence bar on all eight benchmarks, and naming the strongest from a hundred labelled records succeeds only $69\%$ of the time.
Without knowing which to trust, we still read from all ten a verdict at least as good as the best of them, holding a median $93\%$ of \sonnetfull's \kp\ and giving up $2.9$ points of balanced accuracy on average, for a sixty-fourth of the inference cost, and with nothing leaving the deployer's premises.
Out of the eight analysed benchmarks, the panel clearly leads on one and falls materially behind on only two.
The panel's cost is a one-time in-domain calibration of $50$--$100$ records, amortised against \sonnet after some $40$--$81$K items.
Against the label-free ways of reading the same votes the gain is unconditional, the panel being clearly ahead of the average member everywhere and never clearly beaten by an unsupervised aggregator;
against its own best member it is not, the panel leading where competence is spread and errors decorrelate, and otherwise recovering its leader.
Both coordinates of that condition are read from the same calibration set the aggregator already consumes, so a deployer can estimate the regime before committing to the panel.
The main limitation is that the panel is capped by its roster: the gain requires several individually capable members whose errors decorrelate, a condition that might drift from roster to roster and from one model version to the next, and one we characterise for a single fixed roster of frozen generalists.
Adding members pays only where they already decorrelate from one another:
since error-correlation rises with model accuracy~\citep{Kim_Garg_Peng_etal_25,Goel_etal_25}, larger members should not be expected to relax that condition, which makes roster composition the lever worth pulling.

\addtocontents{toc}{\protect\setcounter{tocdepth}{-1}}
\section*{Code and data availability}
\noindent All code, together with the per-dataset run artefacts from which every result is produced, is available as two GitHub repositories, split where a reader could regenerate everything without intensive or costly computation.\footnote{\raimverdictsurl\ and \raimanalysisurl.}
\texttt{raim-verdicts} carries the full inference pipeline, from dataset construction and prompt rendering to the model backends and the verdict parser, and is readily extensible to further models (mainly via Hugging Face) or benchmarks (mainly via the Python \texttt{datasets} library);
\texttt{raim-analysis} carries everything that regenerates from the verdicts, namely the stacked aggregator and cross-fit, the cluster-bootstrap machinery, and the table- and figure-generation pipeline.
The full verdicts, paid \textsc{api} generations included, are released in \texttt{raim-verdicts} as well.

Both repositories are released under CC~BY-NC-SA~4.0.
That licence cannot cover the benchmarks, whose terms a \texttt{NOTICE} file records source by source;
we distribute only our own verdict records, keyed to a content-addressed index against which a third party can verify the join to each publisher's data without our redistributing any source text.
The decision rule and the bootstrap protocol are specified in~\S\ref{ssec:metrics}, the judge prompt, decoding constraints, and coverage accounting in Appendix~\ref{app:coverage}, and the full per-dataset ladders in Appendix~\ref{app:perdataset}.

Every reported number other than the re-run agreements below is generated programmatically from in-repo artefacts: the \texttt{QUICKSTART} of \texttt{raim-analysis} gives the commands that verify the data, rebuild every table and figure from the committed derived tier in minutes, print the headline numbers of the text beside the files they are read from, and regenerate that tier from the plain verdicts within about an hour, with no \textsc{gpu} and no network, under Python~3.12 to~3.14 on macOS and Linux.
The analysis is deterministic at \texttt{seed}~$=0$ on a given machine, whereas across hardware the order in which the numerical libraries accumulate sums perturbs the ninth decimal and beyond;
the release therefore checks a regenerated tier against the committed one at a $10^{-8}$ tolerance on floating-point values and exactly on counts, labels, judging conditions, and panel membership, so that a number which has really moved fails the check whereas a change of hardware does not.
Inference, conversely, is not bit-reproducible across \textsc{gpu} model or serving stack, which is why the verdicts are released as data: a full re-run on independent hardware, with vLLM and \texttt{transformers} a major version ahead, moved individual verdicts (median per-judge agreement $0.98$ for the generative panel, $0.99$ for the constrained scorer) but left every win/tie verdict of~\S\ref{sec:results} unchanged.

\section*{Authors' contribution}

\begin{multicols}{2}\raggedcolumns
\noindent
Conceptualisation: E.O., R.D.P.\\
Data Curation: E.O.\\
Formal analysis: E.O.\\
Funding acquisition: R.D.P.\\
Investigation: E.O.\\
Methodology: E.O., R.D.P.\\
Project administration: R.D.P.\\
Resources: R.D.P.\\
Software: E.O.\\
Supervision: R.D.P.\\
Validation: E.O.\\
Visualisation: E.O.\\
Writing - Original Draft: E.O.\\
Writing - Review \& Editing: E.O., R.D.P.
\end{multicols}

\section*{Ethics statement}
\noindent This work evaluates existing model outputs and collects no new human-generated data: all eight benchmarks are publicly released datasets with published gold labels, used under their respective licences and terms, and no human subjects or annotators were involved.
We redistribute only derived per-item judge verdicts and aggregate results, never the source datasets, which can be fetched from their publishers (all but \factscore through the Python \texttt{datasets} module).
The method's purpose, \ie detecting unsupported content in generated text, is safety-positive and, being an evaluator rather than a generator, offers little scope for misuse.
One benchmark (\medhallu) contains medical question-answering material with synthesised hallucinations; our detector is a research artefact evaluated on that benchmark's terms and is not a clinical tool.

\addtocontents{toc}{\protect\setcounter{tocdepth}{2}}
\bibliographystyle{abbrvnat}
\bibliography{RAIM}

\newpage

\appendix

\section{Roadmap to the appendices}
\label{app:roadmap}

The appendices that follow are organised by topic for navigability, so they do not follow the order in which the main body cites them.
For ease of discoverability, we introduce them briefly here.
Appendices~\ref{app:predictor}--\ref{app:transfer} extend the account the paper is built on, from the item-level test that supports it to the transfer result that bounds it.
Appendices~\ref{app:extbase}--\ref{app:scaling} carry the comparisons against systems we did not build, and the calibrations that license them.
Appendices~\ref{app:perdataset}--\ref{app:limits} hold the per-dataset tables, the measurement protocol, and the reference material a reproduction would need.

Table~\ref{tab:roadmap} maps each appendix to what it carries and to the part of the body it supports, so that any headline in the main text can be traced back to the measurement behind it.
No appendix is required to follow the argument;
each is written to be read on its own, and the body cites it at the point its content is used.

\begin{table}[h]
\centering\footnotesize
\setlength{\tabcolsep}{5pt}
\renewcommand{\arraystretch}{1.15}
\caption{Roadmap to the appendices: what each one carries, and the sections of the main text it supports.}
\label{tab:roadmap}
\begin{tabular}{@{}c p{0.60\linewidth} l@{}}
\toprule
App. & What it carries & Supports \\
\midrule
\multicolumn{3}{@{}l}{\emph{The account, its extensions, and its bounds}} \\
\ref{app:predictor} & The competence/correlation account in full: the item-level recovery test that carries it, the cross-dataset predictor that does not, the regime ordering read dataset by dataset, the member-level views, the panel-size sweep, and the effective-independence comparison against a published frontier panel & \S\ref{ssec:mechanism} \\
\ref{app:budget} & The identity that makes the correlation coordinate free, the obstruction that keeps the competence one from being so, the threshold family and its check on $1{,}888$ (sub-)panels, the labelling budget for both the test and the aggregator, and which half of the rule these eight datasets actually test & \S\ref{ssec:mechanism}, \S\ref{sec:frontier} \\
\ref{app:base} & Matched base-versus-instruct probe: how instruction tuning moves the panel's error-correlation & \S\ref{ssec:mechanism} \\
\ref{app:transfer} & Leave-one-dataset-out transfer, and why the in-domain advantage does not survive it & \S\ref{ssec:mechanism} \\
\addlinespace
\multicolumn{3}{@{}l}{\emph{External comparisons and their calibration}} \\
\ref{app:extbase} & Shared rows, the exact metric conversion, and the harness calibrations behind the comparison with the published leaderboard, and the provenance of the claim-support frame as a natural experiment on member competence & \S\ref{sec:setup}, \S\ref{ssec:mechanism}, \S\ref{sec:frontier} \\
\ref{app:ptj} & The purpose-trained detectors, run through our own harness across the whole suite rather than on their home benchmark alone & \S\ref{sec:intro}, \S\ref{sec:frontier} \\
\ref{app:scaling} & Larger open judges under the same protocol, the panel--frontier paired intervals (Table~\ref{tab:frontier}), and the cost model behind the quoted break-even & \S\ref{sec:frontier} \\
\addlinespace
\multicolumn{3}{@{}l}{\emph{Protocol, tables, and reference material}} \\
\ref{app:perdataset} & Full per-dataset ladders on both metrics, the rescue and aggregator contrasts, and the weights the stacker learns & \S\ref{sec:setup}, \S\ref{sec:results} \\
\ref{app:construction} & How each dataset's class-balanced item set and its clusters were built & \S\ref{sec:setup} \\
\ref{app:coverage} & The judge roster, the prompt and decoding settings, the coverage accounting, and the multiplicity audit & \S\ref{sec:setup}, \S\ref{sec:results} \\
\ref{app:related} & Extended related work: judges, panels of them, cascades, purpose-trained specialists, and the benchmarks & \S\ref{sec:intro}, \S\ref{sec:results}, \S\ref{sec:frontier} \\
\ref{app:limits} & The scope conditions beyond the roster limitation the conclusion names & \S\ref{sec:frontier}, \S\ref{sec:conclusion} \\
\bottomrule
\end{tabular}
\end{table}

\section{The competence/correlation account in full}
\label{app:predictor}
This appendix carries the account of~\S\ref{ssec:mechanism} in full: the two quantitative views of it --an item-level test with real power, and a cross-dataset predictor without it-- then the regime ordering read dataset by dataset, the member-level views, the panel-size sweep, and the effective-independence comparison the body only points at.

\paragraph{Item-level recovery test (confirmatory).}
The mechanism predicts that the panel beats the best single by recovering the leader's errors from decorrelated member competence.
We test this directly: pooling the eight datasets and restricting to the items on which the CV-best single judge is wrong --so that any recovery must come from the other members-- we relate the panel's per-item recovery to the fraction of members individually correct on that item (Table~\ref{tab:recovery}).
The relation is strong and orderly (Table~\ref{tab:recovery}(a)): the panel never recovers an item on which every member shares the leader's error ($0$ of $529$ such items), recovers about one in six once two members dissent, and about six in seven where nine of the ten are right.
A logistic regression with dataset fixed effects gives a coefficient of $+4.67$ (log-odds) on the members-correct fraction, and the correlation between recovery and that fraction --a Pearson correlation between the binary recovery outcome and the continuous fraction, hence a point-biserial one-- is $+0.41$ pooled over the $1{,}998$ items ($p<10^{-20}$) and positive within each of the eight datasets, median $+0.48$ (Table~\ref{tab:recovery}(b)).
This test, and not the dataset-level correlation below, carries the mechanism's quantitative support, for two reasons.
The first is that conditioning on the leader being wrong removes the leader from the test: on these items its verdict is fixed at incorrect, so a relation between recovery and the \emph{other} members' competence cannot be an echo of how strong the leader is, nor of the selection that made it the leader.
The second is that the unit of analysis is the item, so the estimate rests on nearly two thousand points instead of the eight a dataset-level correlation must rest on --- and eight points cannot resolve an effect of this size, as the next paragraph shows.

\begin{table}[t]
\centering\footnotesize
\caption{Item-level mechanism test on the $1{,}998$ items on which the CV-best single judge is wrong.
Items are pooled over the eight datasets, so that any recovery must come from the \emph{other} members.
\textbf{(a)} Recovery probability against the number of members individually correct on the item, counting only those that returned a verdict;
\textbf{(b)} the same relation within each dataset.}
\label{tab:recovery}
\begin{minipage}[t]{0.44\linewidth}
\centering
\setlength{\tabcolsep}{5pt}
\begin{tabular}{c c c}
\toprule
members correct & $P(\text{recovers})$ & $n$ \\
\midrule
0 & 0.000 & 529 \\
1 & 0.058 & 327 \\
2 & 0.162 & 235 \\
3 & 0.101 & 189 \\
4 & 0.157 & 210 \\
5 & 0.225 & 160 \\
6 & 0.310 & 145 \\
7 & 0.374 & 91 \\
8 & 0.468 & 77 \\
9 & 0.857 & 35 \\
\bottomrule
\end{tabular}
\par\smallskip
{\scriptsize (a) by members correct}
\end{minipage}\hfill
\begin{minipage}[t]{0.52\linewidth}
\centering
\setlength{\tabcolsep}{4pt}
\begin{tabular}{l c c c}
\toprule
Dataset & $n$ & $P(\text{recovers})$ & $r$ \\
\midrule
\medhallu & 253 & 0.316 & $+0.75$ \\
\ragtruth & 410 & 0.173 & $+0.16$ \\
\truthfulqa & 431 & 0.074 & $+0.30$ \\
\factscore & 89 & 0.303 & $+0.62$ \\
\wice & 57 & 0.228 & $+0.62$ \\
\xsum & 163 & 0.153 & $+0.53$ \\
\cnn & 40 & 0.100 & $+0.43$ \\
\expertqa & 555 & 0.068 & $+0.37$ \\
\bottomrule
\end{tabular}
\par\smallskip
{\scriptsize (b) by dataset}
\end{minipage}
\tablelegend{$r$:~correlation between recovery and the fraction of members correct on the item.}
\end{table}

\begin{figure}[tp]
\centering
\includegraphics[width=\linewidth]{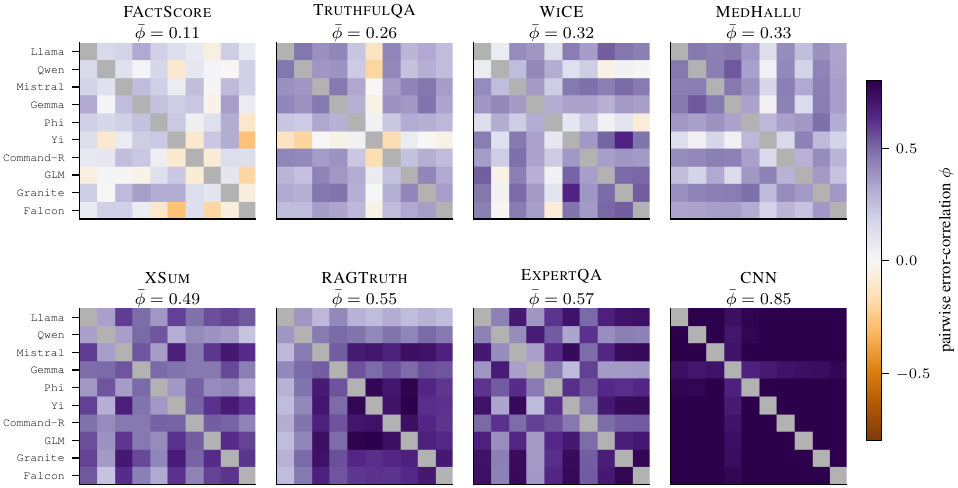}
\caption{Pairwise error-correlation $\phi$ between the ten members, on all eight datasets in ascending $\bar\phi$ (panels share the member order and the colour scale; the diagonal is masked).
The panels run from \factscore (lowest $\bar\phi$) to \cnn (highest).}
\label{fig:paircorr}
\end{figure}

\paragraph{Cross-dataset predictor (illustrative only).}
The dataset-level counterpart of that test is reported here for completeness, and is the one quantity in this appendix the paper does not lean on.
The account also makes a prediction across datasets --the panel's edge should be largest where member errors decorrelate and competence is spread over several members-- and it is fair to ask whether the eight datasets bear that out numerically as well as qualitatively.
We form, per dataset, the product of the two quantities the account names: $1-\bar\phi$, where $\bar\phi$ is the mean pairwise error-correlation of the ten members (the quantity Figure~\ref{fig:paircorr} displays in full), and the number of members reaching $\kp\geq0.30$, our count of how far competence is spread; these are the two coordinates of the plane in Figure~\ref{fig:regimeforest}(a).
Rank-correlating that composite against each dataset's \mbox{(stacked $-$ best-single)} \kp\ gap gives Spearman $\rho=+0.47$ --- the predicted sign, and consistent with the regime ordering read off in~\S\ref{ssec:mechanism}, but not significant ($p=0.24$).
Nor could it easily be, since at $n=8$ only a strong ordering ($\rho \gtrsim 0.74$) would register, and a composite that squeezes two quantities into one number cannot be expected to produce one.

Table~\ref{tab:predictor} reads the two coordinates separately, which is the more informative view and the less flattering one.
The competence coordinate carries the stronger association in every subset ($+0.48$ to $+0.74$), and the correlation coordinate the weaker one ($-0.36$ to $-0.61$).
Both, however, are sensitive to the obvious check of dropping the sets held back in~\S\ref{ssec:data}, yet most of that sensitivity comes from one dataset.
\cnn, in fact, carries the most correlated panel in the suite ($\bar\phi=0.854$, or an effective $n_{\mathrm{eff}}=1.15$ independent votes on the scale defined below) together with the third-largest gap ($+0.051$, behind only \factscore and \wice), which is the opposite of what the account predicts.
It is also the set held back for having too few clusters to resolve any paired difference, so that gap is a central estimate whose interval accommodates either sign.
\wice carried a distortion of another kind, one that re-framing the four \aggrefact sets removed.
Under the misframed prompt none of its members was competent, yet the stacker beat the best single judge by $+0.116$, by correcting a miscalibrated leader rather than by pooling complementary competence (Appendix~\ref{app:extbase});
under the corrected frame, nine of its ten members clear $\kp=0.30$.
The re-framing thus removed a confound, though not the lack of power.

No entry in Table~\ref{tab:predictor} reaches significance at $n\leq8$.
Read instead against the panel's \emph{absolute} stacked \kp, rather than against its gap over the best single, the two coordinates give $\rho=-0.76$ for $\bar\phi$ and $\rho=+0.95$ for the member count; both, however, are confounded by how hard the task is, and neither is a statement about the panel's advantage.
The competence/correlation account does not depend on this reading, its quantitative test being the item-level one of Table~\ref{tab:recovery}, over some two thousand points rather than eight.
We therefore read the dataset-level axis qualitatively, through the regime ordering and the plane of Figure~\ref{fig:regimeforest}(a).

\begin{table}[t]
\centering\small
\caption{Cross-dataset predictor of the panel's gain over its best single judge. Each coordinate is taken separately and as the composite; rows drop, in turn, the two sets held back in~\S\ref{ssec:data}.}
\label{tab:predictor}
\setlength{\tabcolsep}{6pt}
\begin{tabular}{l c c c c}
\toprule
 & & \multicolumn{3}{c}{Spearman $\rho$ (two-sided $p$) against the \mbox{(stacked $-$ best-single)} \kp\ gap} \\
\cmidrule(lr){3-5}
subset & $n$ & $\bar\phi$ & members $\kp\geq0.30$ & $(1-\bar\phi)\times$ members \\
\midrule
all eight & 8 & $-0.36$ (0.39) & $+0.48$ (0.23) & $+0.47$ (0.24) \\
without CNN & 7 & $-0.61$ (0.15) & $+0.74$ (0.06) & $+0.71$ (0.07) \\
without CNN and ExpertQA & 6 & $-0.37$ (0.47) & $+0.58$ (0.23) & $+0.54$ (0.27) \\
\bottomrule
\end{tabular}
\tablelegend{$n$: number of datasets left in the subset.}
\end{table}

\paragraph{The regime ordering by dataset.}
The regime ordering that~\S\ref{ssec:mechanism} reads off Figure~\ref{fig:regimeforest}(a) places the eight datasets in four groups.
The improvement over the best single is largest where competence is distributed over several capable members whose errors decorrelate (\factscore, \medhallu, \wice);
real but unresolved where the members carry moderate, mutually redundant competence (\xsum);
negligible where one model dominates a weak field and cross-validation reliably finds it (\ragtruth and \truthfulqa, in each of which stacking simply recovers \gemma);
and unresolvable either way where competence is scant and the errors are the most correlated in the suite (\cnn, \expertqa).
That ordering carries two qualifications.
The first concerns power rather than effect.
The point gain over the best single is largest on the three least-powered sets, \factscore ($+0.059$ at $n=330$), \wice ($+0.056$ at $n=222$), and \cnn ($+0.051$ at $n=114$), all above \medhallu's clear $+0.032$;
their paired intervals, however, run two to three times wider and cannot exclude zero, so the one resolved win bounds the panel's advantage from below, and the favourable end of the ordering is under-resolved rather than inactive.
The second concerns the contrast the literature would predict.
Grounded-versus-ungrounded is numerically present in the suite yet mechanistically empty: \factscore is the most panel-favourable dataset we measure, whereas \truthfulqa is a textbook dominant-leader set whose difference against the best single is indistinguishable from the grounded \ragtruth's.
The two ungrounded sets thus occupy opposite regimes, and ``ungrounded'' is not a coherent behavioural category for this question.

\paragraph{Member-level views.}
Two member-level figures complete the account, and both are to be read by regime, not member by member.

Figure~\ref{fig:paircorr} gives the pairwise error-correlation structure of the panel on all eight datasets, ordered by $\bar\phi$, and the progression across the eight is the axis the account rests on.
On \factscore ($\bar\phi=0.11$) the members are close to independent and several pairs are mildly \emph{anti}-correlated, so nearly every member carries information the others lack; by \ragtruth ($0.55$) whole blocks of the matrix have gone dark and only a few pairs remain informative; and on \cnn ($0.85$) the panel is close to one judge repeated ten times, which is why aggregation has so little to exploit there whatever the members' individual competence.
The figure also shows the two axes of~\S\ref{ssec:mechanism} to be distinct: \truthfulqa sits at the decorrelated end ($0.26$) and is nonetheless a dominant-leader set whose difference against the best single does not resolve, because only two of its members are individually competent, so low correlation buys nothing without competence to decorrelate.

\begin{figure}[tp]
\centering
\includegraphics[width=\linewidth]{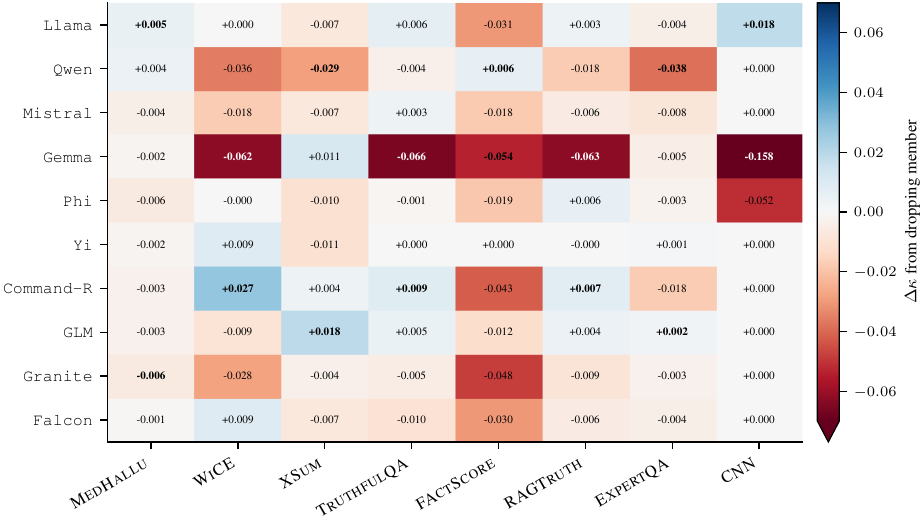}
\caption{Leave-one-out change in stacked \kp\ from dropping each member ($\Delta=\text{stacked without member}-\text{full stacked}$): red marks a load-bearing member, whose removal costs the panel, blue one the panel can spare.
Every cell carries its value, and each column's extremes are set in bold.
The colour scale is clipped at $\pm0.07$ for legibility, \gemma on \cnn at $-0.158$ being off-scale.}
\label{fig:loo}
\end{figure}

Figure~\ref{fig:loo} reads the same regimes member-wise, by refitting the stacker without each member in turn.
Where the panel's improvement is largest, the profile is flat: on \medhallu no member is worth more than $0.006$ \kp\ to the panel, and on \factscore five members each cost $0.03$ or more, so the gain is jointly held rather than owed to any one judge.
Where a single judge dominates, one cell carries the column: dropping \gemma costs $0.066$ on \truthfulqa and $0.063$ on \ragtruth, against at most $0.018$ for any other member; on \wice it costs $0.062$ against $0.036$ for the next member, a leader standing out of a competent field rather than dominating a weak one.
The extreme case is \cnn, where dropping \gemma costs $0.158$ --- by some margin the largest effect in the figure, and a reminder that a $114$-item set with the suite's highest error-correlation is carried by whichever member happens to be competent on it.
No removal helps by more than $+0.03$, so a blue cell marks a member the panel can spare rather than one it would be better without.
Read as a rule, a flat profile signals distributed competence and predicts the larger improvement over the best single, whereas a single deep notch signals a dominant leader and predicts an unresolved one;
in neither case is the gain the contribution of one strong member alone.

\paragraph{Panel size.}
Sweeping the panel from one to ten members corroborates the same ordering directly (Figure~\ref{fig:budget}(a), members added in a greedy decorrelating order and the stacker refitted at each $K$).
Competence keeps accruing with each added member on the low-correlation sets (\medhallu, \factscore);
the leader-dominated (\ragtruth, \truthfulqa) and the redundant (\xsum) sets stay flat;
and \wice peaks early before diluting.
Aggregation therefore rewards decorrelated competence and not member count, so we read the sweep as a diagnostic and not as a recipe for choosing $K$.

\paragraph{Effective independence against a frontier panel.}
\label{app:neff}
The correlation coordinate has a standard summary, which lets us place our panel against one we did not run (\S\ref{ssec:mechanism}).
Writing $\bar\phi$ for the mean pairwise error-correlation over the $\binom{k}{2}$ member pairs (the correlation coefficient amongst the pairwise diversity measures of \citet{Kuncheva_Whitaker_03}, computed on correct/incorrect outputs and hence identical on error indicators), Kish's effective sample size $n_{\mathrm{eff}} = k/(1 + (k-1)\,\bar\phi)$ gives the number of independent judges a panel of $k$ correlated ones is worth.
Our ten members are worth $n_{\mathrm{eff}} = 5.11$ independent votes on \factscore, $2.98$ on \truthfulqa, $2.58$ on \wice, $2.49$ on \medhallu, $1.84$ on \xsum, $1.69$ on \ragtruth, $1.64$ on \expertqa, and $1.15$ on \cnn, which places the correlation axis of the account on a single scale.
\citet{Kohli_26} evaluates the same $\bar\phi$ statistic for a panel of \emph{nine frontier judges} on three natural-language-inference (\textsc{nli}) corpora, reporting $n_{\mathrm{eff}}$ between $2.18$ and $2.48$, with the best single judge unbeaten and an asymptote at $1/\bar\phi \approx 2.6$ read there as a bound on what panels of current models can reach.
No row is shared between the two studies and the tasks differ, so this compares a derived, panel-internal quantity rather than a score;
yet on that quantity our cheap panel exceeds their frontier panel's best on four of eight datasets and passes the asymptote on two, reaching nearly twice that value on \factscore.

In concurrent work on preference data, \citet{Hossain_Yousefi_Lim_26} claim the same ordering on a shrinkage estimate of $\bar\phi$, ten open judges being worth $3.5$ independent ones and three frontier judges $1.4$.
We interpret this as the prediction of \citet{Kim_Garg_Peng_etal_25} and \citet{Goel_etal_25} realised --error-correlation rises with model accuracy, so a panel assembled from weaker and more heterogeneous judges has more independence to aggregate-- and recover Kohli's negative result:
a panel of nine strong, mutually redundant judges sits just above our four least favourable sets on that scale, and on those four we too find aggregation recovers the leader and no more.
It also shows the asymptote to be a property of that roster rather than of panels, since it is $\bar\phi$ that fixes the bound and $\bar\phi$ is a choice of members.

Finally, \citet{Li_Hai_26} argue the same dependence analytically for committees of language models: under a one-factor model, positively correlated errors leave every rule that reads only the votes a positive error floor however large the committee, and reduce a committee of fixed size to single-member behaviour as the correlation approaches one.
The two axes remain distinct here as elsewhere:
\truthfulqa carries $n_{\mathrm{eff}} = 2.98$ and still yields nothing, only two of its members being individually competent.

\section{The admissibility test and its labelling budget}
\label{app:budget}
This appendix carries the measurement behind~\S\ref{ssec:budget}: the rule itself, the identity that makes one regime coordinate free, the obstruction that keeps the other from being so, and the budget at which the call stabilises.
Figure~\ref{fig:budget} collects the two budgets a deployer sets by dataset, the panel size of Appendix~\ref{app:predictor} and the calibration size measured here.

\paragraph{The rule.}
The test reads two statistics off a labelled calibration sample, both computed from the members' own verdicts and neither from the stacker's weights.
The first is the competence spread $s = |\{i \mid \kp_i \geq \kappa_0\}|$, the number of members whose own \kp\ against gold clears a competence bar $\kappa_0$;
the second is $\bar\phi$, the mean over the $\binom{N_v}{2}$ member pairs of the Pearson correlation between their error indicators $e_i = \mathbf{1}[v_i \neq y]$, each pair read on the items both members committed on.
For a competence bar $\kappa_0$, a share $c$, and a correlation bar $\phi_0$, a task is \emph{admissible} when $s \geq c\,N_v$ and $\bar\phi \leq \phi_0$.
The numbers we report are those of one instance, $(\kappa_0, c, \phi_0) = (0.30, \tfrac12, 0.40)$, declared once and applied unchanged to every dataset and every calibration sample, the competence bar being the one of~\S\ref{ssec:mechanism} and Figure~\ref{fig:regimeforest}.
These values were fixed with the eight results in view, yet not tuned: eight datasets are too few for any tuning to be stable, as the next paragraph shows, and the paper therefore relies on what holds across the family rather than on the instance itself.

In use, admissibility decides whether to fit the stacker rather than select a single member, and on this suite every admitted task is also one on which the frontier judge is not materially ahead;
whether the frontier judge is worth keeping on a task that fails it is read, qualitatively, from the same two coordinates, as the dominant-leader corner of the plane (\S\ref{sec:frontier}).
On the full samples the test admits \medhallu ($s=10$, $\bar\phi=0.335$), \wice ($9$, $0.320$), and \factscore ($5$, $0.106$), and rejects the other five (Table~\ref{tab:regimebudget});
the suite's one clear improvement over the CV-best single, on \medhallu, falls inside the admissible set, and no rejected task carries one.

It is however worth stating that the test itself is a decision heuristic rather than a hypothesis test, and carries no interval of its own.
We measure its reliability below instead, as the probability that a calibration sample of a given size reproduces the full-sample call, and name there its two failure cases, the boundary sets \factscore and \truthfulqa.

\paragraph{The rule as a family of thresholds.}
Two properties of the rule, and not of our data, make its thresholds a choice rather than an estimate.
First, the admissible set is monotone in each threshold: raising $\kappa_0$ or $c$, or lowering $\phi_0$, can only remove tasks from it.
The family therefore runs from a screen that admits every task to one that admits none.
As such, choosing a member of it means choosing an operating point between two errors, \ie dropping the frontier judge where it was needed or paying for it where it was not.
Their relative cost, consequently, belongs to the deployer, and the data cannot estimate it.

Second, on a finite suite the calls are piecewise constant in the thresholds, changing only where a threshold crosses some dataset's own coordinate.
Any criterion scored on the suite is therefore optimised by a cell of that partition rather than by a point, and which cell wins is decided by the datasets on its edges.
At $\kappa_0 = 0.30$, our instance lies in a cell of the plane spanned by the required count $\times$ $\phi_0$ that is L-shaped rather than rectangular (hatched area in Figure~\ref{fig:regimeforest}(a)).
The count may run from three to five members and $\phi_0$ from $0.335$ to $0.494$, the upper limit becoming unbounded at a count of five, since no rejected dataset has five competent members.
Its edges are set by the datasets they touch: \medhallu's $\bar\phi$ on the left, \truthfulqa's two competent members below, \factscore's five above, and \xsum at the inner corner.
The competence bar moves the points of that plane rather than an edge, and with the other two thresholds fixed, the result is unchanged for $0.269 \leq \kappa_0 \leq 0.339$, between the fifth-ranked members of \truthfulqa and \factscore;
no call in the paper changes within either range.

The sub-panels of our ten members show both properties at work.
Each of the $1{,}888$ (sub-)panels we draw on the same items (per dataset, every subset of eight or nine members, sixty drawn of five, six, and seven, and the full ten) has its own two coordinates, which we score with the paper's stacker against its own CV-best member and against the frontier judge.
We score a threshold pair by Youden's index~\citep{Youden_50}, $J = \mathrm{TPR} + \mathrm{TNR} - 1$, the rate at which the test admits the panels that meet a target plus the rate at which it rejects those that do not, less one (equivalently, $2\,\bacc - 1$ of the call against the target), so that $J = 0$ for a call unrelated to the target and $J = 1$ for a perfect one.
Selecting $c$ and $\phi_0$ on the panels by $J$, with the target that the panel stays within $0.10$~\kp\ of the frontier judge, gives $c = 0.4$ with any $\phi_0 \geq 0.53$ rather than our instance.
Selected instead on seven datasets and applied to the eighth, the optimum moves from fold to fold ($c$ from $0.3$ to $0.5$, $\phi_0$ from $0.43$ upwards), as the partition predicts, yet it returns our call on all eight held-out datasets, and on seven of eight when the target is to beat the CV-best member.

What does not move, however and most importantly, is the overall direction.
We scan a grid of $c \in \{0.3, 0.4, \dots, 0.8\}$ and $\phi_0 \in \{0.25, 0.26, \dots, 0.90\}$, $6 \times 66 = 396$ pairs, and keep those that split the panels, admitting some and rejecting others.
At every such pair $J > 0$ holds for both targets: the admitted panels are more often within reach of the frontier judge than the rejected ones, and more often above their own best member.
This holds at all $396$ pairs on the drawn panels, every one of which splits them, and at all $375$ that split the eight full ones, the other $21$ ($c \geq 0.6$ with $\phi_0 < 0.32$) admitting none of the eight;
likewise, with the frontier margin tightened to $0.05$~\kp, it still holds at $92\%$ of the pairs on the drawn panels, and at all of them on the full ones.
We therefore offer the test as a family whose every member points the same way, and our instance as one operating point of it.
The sub-panels share members and items, so this is evidence across panel configurations rather than across tasks, and a new task remains the test the suite cannot supply.

\begin{figure}[tp]
\centering
\includegraphics[width=.9\linewidth]{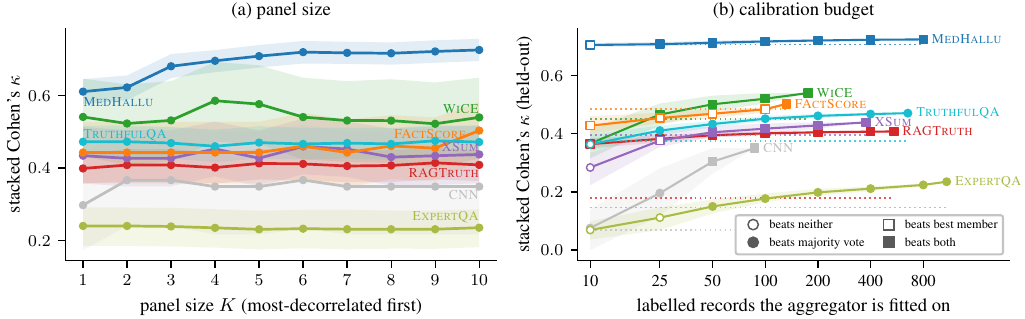}
\caption{The two budgets a deployer sets by dataset.
\textbf{(a)} Stacked \kp\ against panel size $K$, members added in a greedy decorrelating order and the stacker refitted at each $K$ (Appendix~\ref{app:predictor}).
\textbf{(b)} Stacked \kp\ against the labelled \emph{records} the aggregator is fitted on, drawn within the reported stacker's own folds and pooled across them, interquartile range over $R=100$ draws shaded.
The dotted line reports the dataset's label-free majority vote on the same folds; markers encode what a budget beats.}
\label{fig:budget}
\end{figure}

\paragraph{The agreement identity.}
We write $v_i \in \{0,1\}$ for member $i$'s verdict on an item carrying gold $y$, and $e_i = \mathbf{1}[v_i \neq y]$ for its error indicator.
Since $y$ is common to the two members being compared, $v_i = v_j$ holds exactly when $e_i = e_j$, so the pairwise agreement \emph{rate} is the same statistic read through either lens, and averaging over the $45$ member pairs preserves the equality.
Member disagreement is therefore an observable proxy for error-decorrelation.
The identity is not new: it is the pairwise case of the agreement equations from which \citet{Platanios_Blum_Mitchell_14} estimate classifiers' error rates without labels, each agreement rate being written through the two members' error rates and their joint one, whereas we use it only to make the correlation coordinate observable.
The Pearson $\bar\phi$ of Figure~\ref{fig:paircorr} is the same agreement centred and scaled by the members' error rates, and it is that normalisation, not the association, which consumes gold; the unnormalised rate ranks the eight datasets as $\bar\phi$ does at Spearman $\rho = +0.93$ ($p = 0.001$).

\paragraph{The competence obstruction.}
The natural candidate is the Dawid--Skene fit already used as an aggregator baseline (\S\ref{ssec:otheragg}), which estimates each member's reliability by expectation-maximisation without gold.
It fails here for a structural reason.
Its estimate is a member's agreement with a latent consensus, and where the panel shares an error the consensus is the error: on \cnn, the suite's most correlated panel, it rates all ten members competent at $\kp \geq 0.30$ against the consensus, whereas not one of them clears that bar against gold, and the same happens on \expertqa.
Across the eight datasets its count of competent members rank-correlates with the gold count at $\rho = -0.01$, and it recovers the true best member on three datasets of eight --- failing, in particular, on \ragtruth, the suite's clearest dominant-leader set.
\citet{Balasubramanian_Podkopaev_Kasiviswanathan_26} claim a complementary failure on the aggregation side:
on correlated judges, an aggregator that assumes conditional independence can be confidently wrong even when handed each judge's true accuracy and, under some dependence structures, it stays strictly suboptimal however large the panel grows.

Nor does a richer label-free model escape the obstruction without an assumption.
The agreement rates of all subsets of the $N_v$ members supply $2^{N_v}-N_v-1$ equations for the $2^{N_v}-1$ individual and joint error rates, short by exactly the $N_v$ individual ones~\citep{Platanios_Blum_Mitchell_14}, so every such estimator closes the system by an assumption on how the errors depend on one another --conditional independence for Dawid--Skene, the least dependence the agreements allow for \citeauthor{Platanios_Blum_Mitchell_14}, independence between groups of dependent members for \citet{Jaffe_Fetaya_Nadler_etal_16}-- and by members better than chance.
A panel whose errors load on one shared source and whose average member sits near chance, as on \cnn ($\kp = 0.088$), is hence precisely where those assumptions fail.
An unlabelled sample can say that a panel agrees with itself; it cannot say whether the panel is right.

\paragraph{The budget sweep.}
For each dataset and each budget $n \in \{10, 25, 50, 100, 200\}$ not exceeding the dataset, we draw $R = 200$ calibration samples and recompute both coordinates and the panel's apparent leader on each.
Sampling is by cluster and truncated at the cluster boundary, matching the resampling unit of~\S\ref{ssec:metrics}: a deployer buys labelled records, and taking half a matched pair would give the sample a class balance the labelled set would not have.
Table~\ref{tab:regimebudget} reports how often the full-sample regime call is reproduced: on average $85.5\%$ of the time from fifty labelled records and $91.9\%$ from a hundred, with $\bar\phi$ itself recovered to within a median $0.022$ at the latter.
Two datasets sit at the decision boundary and are correspondingly unstable, \factscore carrying the five competent members the rule requires and no more, and \truthfulqa's members clustering at $\kp = 0.30$.

One property of the rule at our instance should be stated plainly.
There, the correlation bar is not what rejects any dataset.
Every dataset with at least five competent members also satisfies $\bar\phi\leq0.40$, and every dataset above the bar has already failed the count, so at the instance the call coincides with the competence threshold alone on all eight.
This is a property of eight points rather than of the rule.
The bar binds from below, since a value under \medhallu's $0.335$ would reject it, and at a share of three or four members it is the bar alone that rejects \xsum;
what the suite lacks at our share is a dataset with five competent members and strongly shared errors, which is why the hatched region of Figure~\ref{fig:regimeforest}(a) runs unbounded at a count of five.
We therefore report the conjunction as the account requires it, noting that the eight full panels do not test its correlation half at our instance.
The sub-panels do populate that cell: $106$ of the $680$ with enough competent members exceed the correlation bar, on \medhallu, \wice, and \xsum, and on the two datasets where both sides of the bar occur the more correlated ones gain less over their best member ($+0.030$ against $+0.038$ on \medhallu, $-0.035$ against $+0.023$ on \wice), the direction the correlation half predicts, although on two datasets only.
The optimum above is flat for every $\phi_0 \geq 0.53$ for the same reason: at $c = 0.4$ no drawn panel with enough competent members lies above that value.

The instability of the two boundary sets belongs to the threshold and not to the estimate: replacing the count by the continuous competence the panel carries above the bar, $\sum_i \max(0, \kp_i - 0.30)$, and normalising each statistic by its own spread across the eight datasets, the continuous form resolves about twice as finely at a fifty-record budget ($0.055$ against $0.100$).
We keep the count in the regime plane of Figure~\ref{fig:regimeforest}(a), where it is read qualitatively and its interpretability is worth more than its efficiency, and report the continuous companion in the released artefact.

\begin{table}[tp]
\centering\footnotesize
\caption{Recovery of the regime call from a limited labelling budget. Per dataset and budget, $R=200$ calibration samples of that many labelled records are drawn as whole clusters (\S\ref{ssec:metrics}), both coordinates of~\S\ref{ssec:mechanism} being re-estimated on each.}
\label{tab:regimebudget}
\setlength{\tabcolsep}{4pt}
\begin{tabular}{l r rr c rrrrr}
\toprule
 & & \multicolumn{3}{c}{\emph{full sample}} & \multicolumn{5}{c}{$\Pr$(call matches) \emph{at a budget of}} \\
\cmidrule(lr){3-5}\cmidrule(lr){6-10}
Dataset & $n$ & $\bar\phi$ & spread & regime & 10 & 25 & 50 & 100 & 200 \\
\midrule
\medhallu & 2000 & $0.335$ & $10$ & admissible & $0.70$ & $0.78$ & $0.88$ & $0.96$ & $0.99$ \\
\wice & 222 & $0.320$ & $9$ & admissible & $0.54$ & $0.67$ & $0.81$ & $0.97$ & $1.00$ \\
\ragtruth & 1362 & $0.546$ & $2$ & not admissible & $0.98$ & $1.00$ & $1.00$ & $1.00$ & $1.00$ \\
\xsum & 546 & $0.494$ & $4$ & not admissible & $0.79$ & $0.87$ & $0.94$ & $0.99$ & $1.00$ \\
\cnn & 114 & $0.854$ & $0$ & not admissible & $0.99$ & $1.00$ & $1.00$ & $1.00$ & --- \\
\expertqa & 1462 & $0.566$ & $0$ & not admissible & $0.96$ & $0.99$ & $1.00$ & $1.00$ & $1.00$ \\
\factscore & 330 & $0.106$ & $5$ & admissible & $0.69$ & $0.54$ & $0.67$ & $0.81$ & --- \\
\truthfulqa & 1634 & $0.261$ & $2$ & not admissible & $0.36$ & $0.51$ & $0.55$ & $0.62$ & $0.76$ \\
\midrule
\emph{suite}: mean $\Pr$(call) & & & & & $\mathbf{0.75}$ & $\mathbf{0.79}$ & $\mathbf{0.85}$ & $\mathbf{0.92}$ & $\mathbf{0.96}$ \\
\emph{suite}: $\Pr$(same leader) & & & & & $0.32$ & $0.46$ & $0.56$ & $0.69$ & $0.76$ \\
\emph{suite}: median $|\Delta\bar\phi|$ & & & & & $0.081$ & $0.046$ & $0.033$ & $0.022$ & $0.015$ \\
\bottomrule
\end{tabular}
\tablelegend{\emph{regime}: the full-sample call, a task being admissible at a competence spread of at least five of the ten members at $\kp\geq0.30$ and $\bar\phi\leq0.40$. Budget columns: fraction of samples reproducing that call; ---: budget exceeding the dataset.}
\end{table}

\paragraph{The aggregator's own budget.}
The sweep above prices the \emph{diagnostic}; the aggregator has a budget of its own, and the two are measured the same way.
Table~\ref{tab:stackercurve} fits the stacker on $n$ labelled records and scores it out-of-fold, against the best single member selected on those same records and the label-free majority vote.
Three design points matter for reading it.
Budgets are in labelled \emph{records}, the clustering unit of~\S\ref{ssec:metrics}, which on the paired sets is two items; a deployer buys records, not items.
The folds are the reported stacker's own, and within each fold the budget is drawn from that fold's training clusters, the fit predicts that fold's test items, and the predictions are pooled before \kp\ is taken --- so the estimator is the paper's, and at the full training pool the curve reproduces the reported value to within $0.003$, which is the check that the protocol has not drifted.
Averaging per-fold \kp, instead, would be a different and biased estimator, \kp\ not being linear in the confusion counts, and disagrees with the pooled one at the full pool by up to $0.010$, on \cnn, whose training pool is the smallest.
The label-free majority vote is drawn on the same folds as a dotted floor per dataset, because what a deployer needs to know is whether the labels they can afford beat spending none.

Two readings the body compresses are clearer per dataset.
The budget at which the fitted stacker overtakes that vote runs from ten labelled records (\ragtruth, \cnn) through twenty-five (\medhallu, \wice, \truthfulqa) to fifty (\xsum, \expertqa), and on \factscore not until its full pool of $132$ --- the vote there being within $0.018$ \kp\ of full supervision, so there is little for labels to buy.
\expertqa is the one set that keeps paying throughout, rising from $\kp = 0.069$ at ten records to $0.235$ at its full pool of $1{,}088$.
That ordering is not the regime ordering of~\S\ref{ssec:mechanism}: the sets on which a handful of labels already pays are the rejected \ragtruth and \cnn, and the admissible \factscore is the slowest, its free vote being already within reach of full supervision.
The two alternatives are not nested, and neither dominates the other.
The free vote is the harder bar on \medhallu and \factscore at every budget, on \wice below fifty records and on \xsum below a hundred;
everywhere else, including on \wice and \xsum once their thresholds are passed, the harder bar is the best member selected on the same labels.
This is why Figure~\ref{fig:budget}(b) encodes the two on separate channels rather than on one ordinal scale.

The two headline budgets quoted in the body summarise that table, and each has its own exceptions.
Fifty labelled records suffice to clear the label-free vote on seven of the eight datasets, \factscore excepted for the reason above;
and a hundred bring the stacker within $5\%$ of its fully supervised \kp\ on six of the eight, the exceptions being \expertqa, which is still $25\%$ short at that budget and keeps paying to its full pool, and \cnn, whose training pool is too small for the hundred-record point to exist.
Below the crossing point a calibration set is worse than none: at ten labelled records the fitted stacker sits \emph{beneath} the free vote on six of the eight sets, all but \ragtruth and \cnn, so a deployer who cannot reach the crossing is better served by not fitting an aggregator at all.

\begin{table}[tp]
\centering\footnotesize
\caption{Gold-label learning curve of the aggregator: mean Cohen's \kp\ over $R=100$ draws per budget, budgets being in labelled \emph{records}, the clustering unit of~\S\ref{ssec:metrics}. Folds are the five of the reported stacker, the budget drawn from each fold's training clusters and the out-of-fold predictions pooled before \kp\ is taken.}
\label{tab:stackercurve}
\setlength{\tabcolsep}{4pt}
\begin{tabular}{l rr rrrrrrrr}
\toprule
Dataset & train & unw. & \multicolumn{8}{c}{\emph{labelled records the stacker is fitted on}} \\
\cmidrule(lr){4-11}
 & & & 10 & 25 & 50 & 100 & 200 & 400 & 800 & full \\
\midrule
\medhallu & 800 & $0.708$ & $0.706$ & $0.709$ & $0.713$ & $0.718$ & $0.721$ & $0.724$ & $0.725$ & $0.725$ \\
\wice & 176 & $0.450$ & $0.367$ & $0.466$ & $0.502$ & $0.521$ & --- & --- & --- & $0.541$ \\
\ragtruth & 544 & $0.179$ & $0.365$ & $0.383$ & $0.395$ & $0.402$ & $0.405$ & $0.408$ & --- & $0.408$ \\
\xsum & 377 & $0.396$ & $0.284$ & $0.377$ & $0.405$ & $0.418$ & $0.429$ & --- & --- & $0.440$ \\
\cnn & 87 & $0.070$ & $0.076$ & $0.196$ & $0.304$ & --- & --- & --- & --- & $0.351$ \\
\expertqa & 1088 & $0.146$ & $0.069$ & $0.112$ & $0.150$ & $0.177$ & $0.199$ & $0.211$ & $0.224$ & $0.235$ \\
\factscore & 132 & $0.485$ & $0.428$ & $0.454$ & $0.469$ & $0.485$ & --- & --- & --- & $0.503$ \\
\truthfulqa & 653 & $0.375$ & $0.364$ & $0.411$ & $0.434$ & $0.452$ & $0.462$ & $0.468$ & --- & $0.471$ \\
\bottomrule
\end{tabular}
\tablelegend{\emph{train}: smallest per-fold training pool, capping the grid; \emph{unw.}: label-free majority vote on the same folds; \emph{full}: the whole pool, reproducing the reported cross-fitted value to within 0.003; ---: budget beyond that pool.}
\end{table}

\section{Base-versus-instruct probe}
\label{app:base}
\begin{figure}[tp]
\centering
\includegraphics[width=\linewidth]{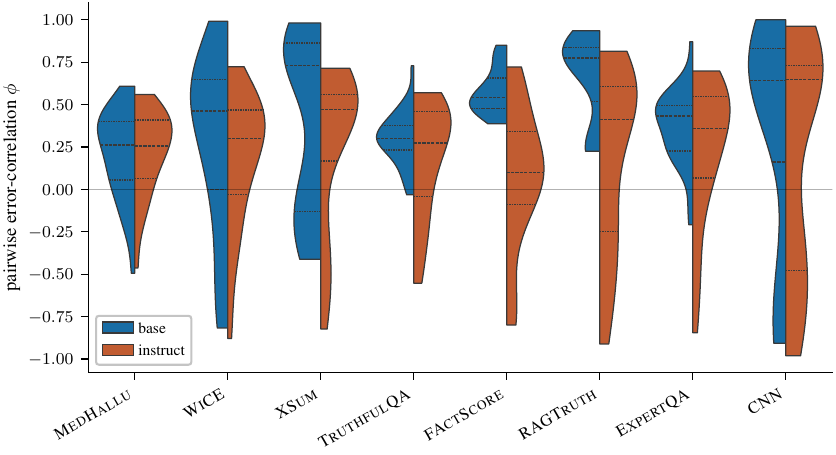}
\caption{Within-panel pairwise error-correlation $\phi$ (the 28 member pairs) for the matched base and instruct panels, per dataset (each dataset under its own scoring frame, \S\ref{ssec:data}; constrained decoding).
The shift $\Delta\phi$ (instruct$-$base) is summarised in Appendix~\ref{app:base}.}
\label{fig:baseprobe}
\end{figure}

This appendix gives the procedure behind the finding of~\S\ref{ssec:mechanism} on where the panel's error-correlation comes from.
To separate a shared answer-bias from shared task-difficulty as the source of the panel's error-correlation, we re-run the eight panel families that release a public base (pre-instruction-tuning) checkpoint, \ie all members except \phimini and \commandr, which ship only in aligned form.
Each is run in both its base and its instruction-tuned variant under a matched constrained-decoding scorer, holding the underlying models fixed.
For each variant we compute the within-panel mean pairwise error-correlation ($\phi$, Pearson on per-item error indicators) across all eight datasets; the two distributions are plotted in Figure~\ref{fig:baseprobe}.
The shift $\Delta\phi$ (instruct$-$base) is negative on seven of the eight datasets, ranging from $-0.085$ on \wice to $-0.495$ on \factscore, and null on \medhallu ($+0.018$, interval including zero); the base panel's $\phi$ reaches $0.667$ on \ragtruth.
Excluding from both panels the one instruct member that collapses to a single class under the constrained scorer (\falcon), the shift remains clearly negative on six of the eight datasets, \xsum's becoming unresolved ($+0.027$ $[-0.021, +0.073]$) and \medhallu's small positive shift resolving ($+0.046$ $[+0.023, +0.067]$).
Instruction tuning thus lowers the panel's error-correlation rather than raising it: the correlation the aligned panel carries is already present, and larger, in the base checkpoints, which is the sense in which~\S\ref{ssec:mechanism} reads it as inherited from pretraining rather than induced by alignment.
Being computed on eight members under constrained decoding, the probe's $\phi$ is not the ten-member $\bar\phi$ of Table~\ref{tab:regimebudget}, and only its shift is read here.

\section{Transfer to an unseen dataset}
\label{app:transfer}
The in-domain gain raises the deployment question directly: can a panel calibrated on some datasets be deployed on another without fresh labels?
It cannot, unchanged, and this appendix gives the measurement, the mechanism, and the boundary of the negative.
The conclusion we draw from it is a scope condition on~\S\ref{sec:frontier}'s costing --the calibration budget is per domain-- rather than a result we develop further.

\begin{figure}[tp]
\centering
\includegraphics[width=.8\linewidth]{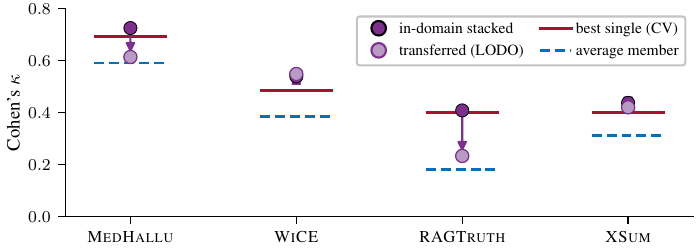}
\caption{Leave-one-dataset-out transfer on the four core grounded sets: in-domain stacked and transferred (\textsc{lodo}), against the CV-best single judge and the average member per dataset.}
\label{fig:transfer}
\end{figure}

\paragraph{Measurement and mechanism.}
We test with a pooled stacker under a leave-one-dataset-out (\textsc{lodo}) protocol on the four core grounded sets, fitting on the other three and evaluating on the held-out set (Figure~\ref{fig:transfer}).
Off-domain, the panel returns to roughly the level of the label-free baselines and wins nothing: zero of four transfer wins, with a significant transfer cost --the paired \mbox{(in-domain $-$ transferred)} gap-- on the two question-answering sets, \medhallu ($+0.111$) and \ragtruth ($+0.175$).
\medhallu's clean win is lost --the transferred stacker drops below even the best single ($-0.079$)-- and \ragtruth's unresolved in-domain improvement becomes an outright transfer loss ($-0.165$).
On \wice and \xsum the transferred stacker shows no clear transfer cost and does not fall below the best single, though the sample cannot rule out a cost there either.

The mechanism is the one that explains the in-domain pattern: the pooled stacker learns to trust the members that are reliable \emph{on average} over the training pool, but the most reliable judge changes by dataset, so the fixed off-domain weight vector systematically under-weights the held-out set's specific leader.
On \medhallu the in-domain leader is \granite, yet the off-domain fit gives \granite little weight against \gemma, so the transferred stacker lands even below \medhallu's own best-on-average single.
The apparent \medhallu balanced-accuracy gain on transfer is a coverage artefact: it beats only the abstaining CV-best baseline, and loses to the non-abstaining label-free baselines.
The negative therefore reinforces the mechanistic account of~\S\ref{ssec:mechanism}:
the panel helps by dataset-specific competence re-weighting, and a fixed weight vector is what that mechanism predicts should fail.

\paragraph{Scope of the negative result.}
The measurement above is taken in one setting, and the scope matters for what may be concluded from it.
That setting is the generative panel, read through hard verdicts, with a stacker pooled over the other three datasets.
It says that a weight vector fitted on other domains does not carry the in-domain gain, and it identifies why --- the leader moves, and a fixed vector cannot follow it.
It does not say that no panel transfers.
In particular, the members are read here through the interface a closed judge would also expose, a verdict rather than a graded confidence, so whether a panel whose members are put on a common scale across domains transfers better is a question this setting cannot reach and we do not take up.
We accordingly take the conservative reading, that the calibration budget of~\S\ref{sec:frontier} is per domain, and leave a calibrated-panel transfer study to future work.

\section{Comparability with published results}
\label{app:extbase}
The comparison of~\S\ref{sec:frontier} against the published results needs three things to be legitimate: a shared set of rows, a shared metric, and evidence that our harness and the leaderboard's measure the same thing on them.
This appendix supplies all three, states what the comparison cannot reach, and reports the two sets whose published anchors exist but sit on a different denominator.

\paragraph{Shared rows and metric conversion.}
Four of our grounded sets (\cnn, \xsum, \wice, \expertqa) are drawn from the test split of the \aggrefact benchmark~\citep{Tang_Laban_Durrett_24}, whose public leaderboard\footnote{\url{https://llm-aggrefact.github.io/}, retrieved 2026-07-08.} reports balanced accuracy per dataset, so published methods and ours are scored on the same underlying rows.
On a class-balanced set, the chance-agreement term of Cohen's \kp\ is exactly $0.5$ whatever the judge's marginal, so for a full-coverage judge $\kp = 2\,\bacc - 1$ holds as an identity, not an approximation; abstention perturbs it only through the committed-set treatment of Appendix~\ref{app:coverage}.
Our class balancing subsamples the majority class, which leaves the expected per-class recall --hence balanced accuracy-- unchanged, so a published \bacc\ is comparable to ours in expectation; F1 and precision are prevalence-bound and are, conversely, not.

\paragraph{Harness calibration.}
A shared metric on shared rows still leaves the two harnesses free to disagree, so Table~\ref{tab:extbase} carries two systems run on both sides.
Re-running \mdname{MiniCheck-Flan-T5-L} through the official MiniCheck package on our instances gives $68.4$ mean balanced accuracy against its published $68.8$, and re-running \qwenbig{72B} through our own prompting harness gives $70.4$ against a published $69.2$ for the unquantised model.
Both agree to about a point, in opposite directions, which is the resolution at which we read the table: differences under three points between blocks are not informative.
A second and independent calibration comes from \medhallu, where three of the published detectors are members of our own panel (\S\ref{sec:frontier}): our runs of them agree with their published hallucinated-class F1 to within $2.6$ points, on a different benchmark and in a different metric from the two rows above.
The exception is \mdname{Bespoke-MiniCheck-7B}, which we do not reproduce ($58.3$ against a published $71.4$) and whose published row we therefore read instead of ours: its package scores that checkpoint through a generated-token probability rather than through the classifier head its \mdname{Flan-T5} sibling uses, and the loss shows as a drop in area under the receiver operating characteristic curve (\textsc{auc-roc}) on the very same items, so we attribute it to that scoring path rather than to our protocol and do not pursue it here.

\begin{table*}[tp]
\centering\footnotesize
\caption{The four \aggrefact\ sets in leaderboard shape (balanced accuracy), the only sets on which we score the same rows as the published results.
Upper block published; lower block measured here under the one protocol and per-dataset frame of~\S\ref{ssec:panel}.}
\label{tab:extbase}
\setlength{\tabcolsep}{4pt}
\resizebox{\textwidth}{!}{%
\begin{tabular}{l l cccc c l}
\toprule
System & size & \cnn & \xsum & \wice & \expertqa & mean & source \\
\midrule
\multicolumn{8}{l}{\emph{published, from the leaderboard and the cited papers}} \\
\mdname{Bespoke-MiniCheck-7B} & 7B & $65.5$ & $77.8$ & $83.0$ & $59.2$ & $71.4$ & {\scriptsize leaderboard} \\
\mdname{MiniCheck-Flan-T5-L} & 0.8B & $69.9$ & $74.3$ & $72.2$ & $59.0$ & $68.8$ & {\scriptsize \citealp{Tang_Laban_Durrett_24}} \\
\mdname{FactCG-DeBERTa-L} & 0.4B & $70.1$ & $73.9$ & $74.2$ & $59.1$ & $69.3$ & {\scriptsize \citealp{Lei_Li_Li_etal_25}, leaderboard} \\
\mdname{HalluGuard-4B}$^{\dagger}$ & 4B & $70.7$ & $75.5$ & $80.5$ & $59.4$ & $71.5$ & {\scriptsize \citealp{Bergeron_Buhnila_Francois_etal_26}} \\
\mdname{Granite Guardian 3.3} & 8B & $67.0$ & $74.9$ & $76.6$ & $59.6$ & $69.5$ & {\scriptsize leaderboard} \\
\mdname{GPT-4o} & --- & $68.1$ & $76.8$ & $78.5$ & $59.6$ & $70.8$ & {\scriptsize \citealp{Lei_Li_Li_etal_25}, leaderboard} \\
\mdname{Claude-3.5 Sonnet} & --- & $67.6$ & $75.1$ & $77.7$ & $60.9$ & $70.3$ & {\scriptsize leaderboard} \\
\mdname{Qwen2.5-72B-Instruct} & 72B & $63.6$ & $73.0$ & $80.2$ & $60.1$ & $69.2$ & {\scriptsize leaderboard} \\
\mdname{SummaC-ZS} & 0.4B & $51.1$ & $61.5$ & $62.8$ & $55.2$ & $57.6$ & {\scriptsize \citealp{Laban_Schnabel_Bennett_etal_22}, scored by \citealp{Tang_Laban_Durrett_24}} \\
\midrule
\multicolumn{8}{l}{\emph{this work, one protocol and one frame per dataset across all eight sets}} \\
average panel member & 4--9B & $54.2$ & $65.0$ & $63.9$ & $55.2$ & $59.6$ &  \\
best single member (CV-selected) & 4--9B & $64.9$ & $69.9$ & $73.9$ & $62.0$ & $67.7$ &  \\
\mdname{MiniCheck-Flan-T5-L} \emph{(our run)} & 0.8B & $67.5$ & $74.2$ & $73.0$ & $58.7$ & $68.4$ &  \\
\mdname{Bespoke-MiniCheck-7B} \emph{(our run)}$^{\S}$ & 7B & $55.2$ & $63.4$ & $60.8$ & $53.9$ & $58.3$ &  \\
\qwenbig{72B} \textsc{awq} \emph{(our run)} & 72B & $67.5$ & $72.3$ & $81.5$ & $60.1$ & $70.4$ &  \\
\textbf{stacked cheap panel (ours)} & $10\times$4--9B & $67.5$ & $71.9$ & $77.0$ & $61.8$ & $69.5$ &  \\
\sonnet \emph{(our run)} & --- & $73.5$ & $70.9$ & $77.4$ & $58.2$ & $70.0$ &  \\
\bottomrule
\end{tabular}}
\tablelegend{
\emph{source}:~where each published value is found, \emph{leaderboard} being the benchmark's public leaderboard (\url{https://llm-aggrefact.github.io/}, retrieved 2026-07-08), which carries systems post-dating the papers, and a citation being given only where the cited work itself reports the four values;
$^{\dagger}$:~run by its own authors and reported in their paper, not on the leaderboard;
$^{\S}$:~a checkpoint we do not reproduce ($58.3$ against $71.4$ published), whose published row we read instead (Appendix~\ref{app:extbase}).}
\end{table*}

\begin{table}[t]
\centering\footnotesize
\caption{Hallucinated-class F1 on \medhallu\ and \ragtruth, published and measured here.
Published rows are transcribed from the cited papers (\ragtruth: response-level F1 on its QA subtask, which we adopt); ours are computed on our balanced construction (Appendix~\ref{app:construction}), an abstention counting as a missed detection.
Slices differ: \medhallu's published table pools the $10{,}000$-pair superset against our $1{,}000$-row \mdname{pqa\_labeled} slice, and \ragtruth's sits on its QA test split, at a hallucinated rate of $0.178$ against our $0.5$, so its two blocks are not comparable in level, F1 moving with prevalence.
Three published \medhallu\ detectors are members of our panel (\S\ref{sec:frontier}).}
\label{tab:extf1}
\setlength{\tabcolsep}{4pt}
\begin{tabular}{l l c}
\toprule
System & & F1 \\
\midrule
\multicolumn{3}{l}{\medhallu \emph{--- published}} \\
\quad \mdname{GPT-4o}~{\scriptsize\citep{Pandit_Xu_Hong_etal_25}} & {\scriptsize frontier, prompted} & $87.7$ \\
\quad \mdname{Qwen2.5-14B}~{\scriptsize\citep{Pandit_Xu_Hong_etal_25}} & {\scriptsize single open judge} & $85.2$ \\
\quad \mdname{GPT-4o-mini}~{\scriptsize\citep{Pandit_Xu_Hong_etal_25}} & {\scriptsize budget \textsc{api}} & $84.1$ \\
\quad \mdname{Qwen2.5-7B}~{\scriptsize\citep{Pandit_Xu_Hong_etal_25}} & {\scriptsize \emph{our panel member}} & $83.9$ \\
\quad \mdname{Gemma-2-9B}~{\scriptsize\citep{Pandit_Xu_Hong_etal_25}} & {\scriptsize \emph{our panel member}} & $83.8$ \\
\quad \mdname{Llama-3.1-8B}~{\scriptsize\citep{Pandit_Xu_Hong_etal_25}} & {\scriptsize \emph{our panel member}} & $79.7$ \\
\multicolumn{3}{l}{\emph{measured here, no member trained on the corpus}} \\
\quad best member by F1 (\gemma) & {\scriptsize ours} & $84.3$ \\
\quad \textbf{stacked cheap panel} & {\scriptsize ours} & $\mathbf{86.2}$ \\
\quad \sonnet & {\scriptsize ours} & $89.0$ \\
\midrule
\multicolumn{3}{l}{\ragtruth \emph{--- published}} \\
\quad \mdname{RAG-HAT} (\mdname{Llama-3-8B})~{\scriptsize\citep{Song_Wang_Zhu_etal_24}} & {\scriptsize trained on the \ragtruth\ train split} & $74.8$ \\
\quad \mdname{LettuceDetect-large}~{\scriptsize\citep{Kovacs_Recski_25}} & {\scriptsize trained on the \ragtruth\ train split} & $70.2$ \\
\quad fine-tuned \mdname{Llama-2-13B}~{\scriptsize\citep{Niu_Wu_Zhu_etal_24}} & {\scriptsize trained on the \ragtruth\ train split} & $68.2$ \\
\quad \mdname{GPT-4-turbo}, prompted~{\scriptsize\citep{Niu_Wu_Zhu_etal_24}} & {\scriptsize prompt-only frontier} & $45.6$ \\
\quad SelfCheckGPT (\mdname{GPT-3.5})~{\scriptsize\citep{Niu_Wu_Zhu_etal_24}} & {\scriptsize prompt-only, resampling} & $43.7$ \\
\multicolumn{3}{l}{\emph{measured here, no member trained on the corpus}} \\
\quad best member by F1 (\llama) & {\scriptsize ours} & $62.5$ \\
\quad \textbf{stacked cheap panel} & {\scriptsize ours} & $\mathbf{66.3}$ \\
\quad \sonnet & {\scriptsize ours} & $76.2$ \\
\bottomrule
\end{tabular}
\end{table}

\paragraph{The other four benchmarks.}
Published work reports on the remaining four benchmarks too, and we record where the field stands on them: on \medhallu Table~\ref{tab:extf1} matches the metric and compares levels, and on \ragtruth it matches the metric but not the prevalence, whereas on \truthfulqa and \factscore no published number is a baseline we could be scored against.
Tables~\ref{tab:extf1} and~\ref{tab:extcontext} give the strongest anchor we found per set, the metric its authors report, and the structural difference that separates it from ours.
However, the difference changes from set to set, and is never a matter of prompting alone.

On \medhallu the published detection table pools the full $10{,}000$-pair superset and reports hallucinated-class F1, whereas we use the $1{,}000$-row \mdname{pqa\_labeled} slice~\citep{Pandit_Xu_Hong_etal_25};
the levels nonetheless line up, \mdname{GPT-4o}'s $87.7$ with knowledge against our \sonnet's $89.0$ and our panel's $86.2$ in the same metric, and three of the detectors reported individually there (\mdname{Qwen2.5-7B} at $83.9$, \mdname{Gemma-2-9B} at $83.8$, \mdname{Llama-3.1-8B} at $79.7$) are members of our own panel, so a reader can compare the two tables member by member.
On \ragtruth every strong published detector is trained on that corpus's own training split and scored on its original test split at its own prevalence, whereas ours is a paired, balanced construction from the processed corpus's training split~\citep{Niu_Wu_Zhu_etal_24,Song_Wang_Zhu_etal_24,Kovacs_Recski_25}.
That test split runs at a hallucinated rate of $0.178$ against our $0.5$, and F1 moves with prevalence, so Table~\ref{tab:extf1} sets the two side by side without ranking them.
Read on the question-answering (QA) subtask we take, the prompt-only anchors there are far weaker (\mdname{GPT-4-turbo} at F1 $45.6$) than the trained ones (\mdname{RAG-HAT} at $74.8$), which is the in-domain-supervision axis of this paper appearing in someone else's numbers.
On \truthfulqa the canonical automated judge is fine-tuned on in-domain human ratings of \emph{model generations} and reaches $90$--$96\%$ validation accuracy on that task, whereas we score the dataset's own reference answers zero-shot~\citep{Lin_Hilton_Evans_22}.
Finally, on \factscore the published estimators are retrieval-augmented by construction, approximating the human score to within $2\%$ at corpus level, whereas we supply no evidence by design~\citep{Min_Krishna_Lyu_etal_23}.

Read together, the four make one point, and it is the paper's own.
Wherever a published system clearly leads us, it has either been trained on the target corpus or been given evidence we withhold, and the prompted frontier anchor on \medhallu lands where our frontier judge lands.

A last caveat covers all eight sets: every published number arrives under its own prompting and inference protocol, and factuality-metric comparisons are known to be fragile to exactly such differences~\citep{Godbole_Jia_25}.
Our absolute levels embed one frame per dataset (\defon, or the claim-support frame on the \aggrefact sets), applied uniformly to all ten members and the frontier judge (\S\ref{ssec:panel}), which the within-suite contrasts control for but cross-paper comparisons do not.

\begin{table*}[tp]
\centering\scriptsize
\caption{Published anchors on the two benchmarks without a comparable published baseline: per set, the strongest anchor we found, the metric its authors report, and the structural difference that blocks the comparison.
\medhallu\ and \ragtruth, reported in hallucinated-class F1, are in Table~\ref{tab:extf1}.}
\label{tab:extcontext}
\setlength{\tabcolsep}{3pt}
\begin{tabular}{l p{0.19\textwidth} c l p{0.30\textwidth} c c}
\toprule
 & \multicolumn{4}{c}{\emph{published anchor}} & \multicolumn{2}{c}{\emph{ours} (\bacc)} \\
\cmidrule(lr){2-5}\cmidrule(lr){6-7}
Dataset & strongest reported system & value & metric & what blocks a direct comparison & panel & \sonnet \\
\midrule
\truthfulqa & \mdname{GPT-judge} (fine-tuned \mdname{GPT-3-6.7B})~{\scriptsize \citealp{Lin_Hilton_Evans_22}} & $90$--$96$ & accuracy & judges model \emph{generations} after fine-tuning on in-domain human ratings; we score the dataset's own reference answers zero-shot & $73.6$ & $81.1$ \\
\factscore & retrieval-augmented estimator~{\scriptsize \citealp{Min_Krishna_Lyu_etal_23}} & --- & system-level error & retrieves Wikipedia evidence by construction, approximating the human score to within $2\%$ at corpus level; we supply no evidence and score per claim & $75.2$ & $77.6$ \\
\bottomrule
\end{tabular}
\tablelegend{\emph{ours}:~balanced accuracy on our construction (Appendix~\ref{app:construction}), \emph{not} commensurable with the anchor beside it.}
\end{table*}

\paragraph{Summary of the comparison.}
Two readings survive the caveats, beyond the placement recorded in~\S\ref{sec:frontier}.
\expertqa's low-headroom status is corroborated externally: the leaderboard's best sits at a \bacc\ of $60.9$ across everything up to \mdname{Claude-3.5} and \mdname{Llama-405B}, and our panel reaches that ceiling ($61.8$) without beating its own best single, so the low-headroom verdict of~\S\ref{ssec:data} is a property of the dataset rather than of our panel.
An independent audit is consistent with this, finding \expertqa the most ambiguous of our four \aggrefact sets ($14$ of $100$ sampled items;~\citealp{Seo_etal_25}).
The second reading, that the sharpest competitor to a cheap panel is a small trained verifier rather than another panel, is taken up in the body (\S\ref{sec:frontier}) and not repeated here.

\paragraph{Provenance of the claim-support frame.}
The four \aggrefact sets set the item's question equal to its candidate (the unit judged is one claim against a document), so an initial question--answer rubric (\defon) presented that claim twice and misframed a claim-verification task; we discovered this by comparing our absolute agreement to the leaderboard, where the unweighted \defon\ panel sat some twenty balanced-accuracy points below the best published result on those four sets.
We therefore score these four under a MiniCheck-style claim-support frame (\S\ref{ssec:panel}), which lifts the unweighted panel by $10.7$ of those points on average (Table~\ref{tab:frame}): on \xsum the frontier judge moves from $\kp = 0.269$ to $0.418$ (\bacc\ $63.5 \to 70.9$), and the panel and all ten members rise in step, bringing the stacked panel to within about a point of the published range on every claim-verification set (Table~\ref{tab:extbase}).
The frontier judge is never worse off for the change either, its \kp\ rising on all four sets (Table~\ref{tab:frame}, third block), \expertqa included ($0.137 \to 0.165$), so the frame each set is scored under is the better of the two for the judge the panel is compared with, and not only for the panel.
The two frames sit on opposite sides of the operating point.
Under the lenient question--answer frame the frontier judge catches only $39\%$ of the hallucinated \xsum items, and under the claim-support frame $95\%$.
The shift is therefore one of task scaffolding, not of the judge's ability to read the evidence.

The two frames double as a controlled manipulation of member competence, and the competence/correlation account of~\S\ref{ssec:mechanism} predicts what each should do.
Under the misframed \defon\ prompt, member competence collapsed to the floor on all four sets (no member reached $\kp=0.30$ on any of them; Table~\ref{tab:frame}).
However, the account predicts that a stacker facing crippled, heterogeneously miscalibrated members has the \emph{most} to gain over any single one.
Indeed, the (stacked\,$-$\,best-single) edge was larger under that frame, on three of the four sets and on average ($+0.057$ against $+0.035$).
On \wice it reached a significant win ($+0.116$), which on inspection was the stacker correcting the leader's frame-induced miscalibration rather than harnessing complementary competence.
However, restoring the frame restores member competence (nine of ten members above $\kp=0.30$ on \wice), and the edge reverts to the unresolved improvement of~\S\ref{ssec:beating} ($+0.056$), a competent field also making its best single strong.
The misframing is thus an unplanned natural experiment.
Under a degraded prompt the panel's advantage inflates as the members are crippled and made heterogeneous, and shrinks as restored competence lifts the best single with them, as the mechanism says it should.
This is evidence for the account independent of the within-frame cross-dataset pattern and of the item-level test.

\begin{table}[t]\centering\footnotesize
\caption{Frame contrast on the four \aggrefact\ sets: the question--answer rubric (\defon) against the claim-support frame. \sonnet's \kp\ is read from the judge records the main results use.}
\label{tab:frame}
\setlength{\tabcolsep}{4pt}
\begin{tabular}{l rr l cc cc}
\toprule
 & \multicolumn{3}{c}{unweighted panel \bacc} & \multicolumn{2}{c}{members at $\kp \ge 0.30$} & \multicolumn{2}{c}{\sonnet\ \kp} \\
\cmidrule(lr){2-4}\cmidrule(lr){5-6}\cmidrule(lr){7-8}
Dataset & \defon & claim & $\Delta$ & \defon & claim & \defon & claim \\
\midrule
\xsum & $51.6$ & $69.8$ & $+18.1$\,{\scriptsize[+14.5,\,+21.7]} & $0/10$ & $4/10$ & $0.269$ & $0.418$ \\
\wice & $56.8$ & $72.5$ & $+15.8$\,{\scriptsize[+10.0,\,+22.0]} & $0/10$ & $9/10$ & $0.539$ & $0.548$ \\
\cnn & $50.0$ & $53.5$ & $+3.5$\,{\scriptsize[+0.8,\,+7.0]} & $0/10$ & $0/10$ & $0.261$ & $0.469$ \\
\expertqa & $51.8$ & $57.3$ & $+5.5$\,{\scriptsize[+3.2,\,+7.7]} & $0/10$ & $0/10$ & $0.137$ & $0.165$ \\
\bottomrule
\end{tabular}
\tablelegend{\emph{claim}: claim-support frame. $\Delta$: paired difference (claim $-$ \defon) with its 95\% cluster-bootstrap interval, the same resample applied to both arms.}
\end{table}

\section{Purpose-trained cheap judges off their home benchmark}
\label{app:ptj}
The competing route to a cheap judge fine-tunes a single small model expressly for evaluation, instead of aggregating a panel of off-the-shelf generalists after the fact (Appendix~\ref{app:related}).
It is the natural single-judge foil to the panel, and the published results report it almost exclusively on \aggrefact, which is the corpus family the strongest of these systems were trained for.
We therefore run them ourselves across the whole suite, which is what licenses the scope reading of~\S\ref{sec:frontier} and Table~\ref{tab:profile}.

Two families enter.
The grounded-factuality specialists \mdname{MiniCheck-Flan-T5-L} and \mdname{Bespoke-MiniCheck-7B}~\citep{Tang_Laban_Durrett_24} are run through the official MiniCheck inference package rather than our judge-prompt harness, so their published operating point is preserved; they verify a claim against a document and so apply to the six grounded sets alone.
The evaluation-tuned judges \mdname{Prometheus-2}~\citep{Kim_Suk_Longpre_etal_24}, \mdname{JudgeLM}~\citep{Zhu_Wang_Wang_25}, and \mdname{Auto-J}~\citep{Li_Sun_Yuan_etal_24} are run over the same balanced binary instances through a per-judge adapter that renders each instance in the judge's \emph{native} graded interface (\mdname{Prometheus-2}'s $1$--$5$ absolute-grading rubric, \mdname{JudgeLM}'s and \mdname{Auto-J}'s $1$--$10$ single-answer scores) and parses the native output back to our label space, thresholding at the scale midpoint.
The reference slot of the graded template carries a rubric describing the ideal faithful answer, never the gold label, so no answer leaks; the source evidence is placed in the instruction exactly as the panel sees it.
Scoring is identical to every other judge in the study: the same instances, non-answers charged through coverage, and the same $B=2000$ \texttt{row\_id} cluster bootstrap.

Three caveats bound the reading, and each, if anything, understates the comparators.
The evaluation-tuned three were trained predominantly for general response quality or pairwise preference rather than reference-grounded faithfulness, so we read them as off-the-shelf single-judge baselines and not as faithfulness specialists; \mdname{JudgeLM} in particular is natively pairwise and is run through its single-answer grading path.
The midpoint threshold is a documented default and not a tuned operating point, so these numbers are a floor a dev-split calibration could only raise.
Finally, their context windows differ by architecture: only \mdname{Prometheus-2} (built on a $32$k-context \mdname{Mistral}) reaches the $8192$-token window the panel and the frontier judge use, whereas \mdname{Auto-J} (\mdname{Llama-2}, $4096$) and \mdname{JudgeLM} (\mdname{Vicuna}/\mdname{Llama-1}, $2048$) are architecturally capped below it, so on the long-document sets they see middle-out-truncated evidence --- a property of the checkpoint and not a design choice, which can only depress their scores.
None of the three bears on the specialist comparison that carries the scope argument, since \mdname{MiniCheck-Flan-T5-L} is run at its own published operating point through its own package and reproduces its published \aggrefact value to within a point (Appendix~\ref{app:extbase}).
Table~\ref{tab:ptjudge} gives all five judges' Cohen's \kp\ per dataset, beside the stacked panel, its best single member, and the frontier judge; coverage runs from $0.892$ to $1.000$ across every purpose-trained judge/dataset cell, so none of these rows is decided by a parsing failure.

\paragraph{On the applicability of \mdname{Granite Guardian}.}
\mdname{Granite Guardian 3.3} is the one comparator in Table~\ref{tab:profile} marked only \emph{partly} applicable to ungrounded factuality, and the qualification is worth spelling out, since it is what separates that row from the two \mdname{MiniCheck} rows above it.
It is a multi-risk guardrail model and not a faithfulness verifier, and much of what it detects needs no reference at all: jailbreak attempts, profanity, toxicity, hate speech, and the harm dimensions of the IBM AI Risk Atlas are scored from the prompt and the response alone.
It can therefore be pointed at \factscore and \truthfulqa, which supply no reference at scoring time (\S\ref{ssec:data}), where the \mdname{MiniCheck} interface cannot be run at all.
It cannot, however, deliver there the faithfulness judgement we measure.
Of the risks it defines, only context relevance, groundedness, and answer relevance concern hallucination, and each is scored against a retrieved context, so on an ungrounded item the detector that would carry the faithfulness judgement has nothing to condition on.
The \partialy\ mark records this split, the model running on ungrounded data whilst its faithfulness judgement does not carry over.
We give the row no citation: the family's technical report documents version~3.0, reports no \aggrefact result, and has not been revised since December~2024, so both the results and the version we quote come from the leaderboard.
Nor would its successor change the row: the model card of \mdname{Granite Guardian 4.1}\footnote{\url{https://huggingface.co/ibm-granite/granite-guardian-4.1-8b}, retrieved 2026-09-25.}, which is not on the leaderboard, gives per-dataset results whose mean over our four sets is $68.7$ without and $68.6$ with its reasoning mode, against $69.5$ for version~3.3 with reasoning on the same card.
\begin{table*}[t]
\centering\small
\caption{Purpose-trained cheap judges off their home benchmark, against the panel and the frontier judge: Cohen's \kp\ on identical balanced instances, each dataset under its own scoring frame (claim-support on the \aggrefact\ sets).
Each purpose-trained judge runs off the shelf through its native interface (Appendix~\ref{app:ptj}), thresholded at its documented operating point.}
\label{tab:ptjudge}
\setlength{\tabcolsep}{2.5pt}
\resizebox{\textwidth}{!}{%
\begin{tabular}{l ccccc cc c}
\toprule
 & \multicolumn{5}{c}{purpose-trained judge $\kappa$} & \multicolumn{2}{c}{panel $\kappa$} & \\
\cmidrule(lr){2-6}\cmidrule(lr){7-8}
Dataset & {\mdname{Prometheus-2}} & {\mdname{JudgeLM}} & {\mdname{Auto-J}} & \shortstack{{\mdname{MiniCheck-}}\\{\mdname{Flan-T5-L}}$^{\diamond}$} & \shortstack{{\mdname{Bespoke-}}\\{\mdname{MiniCheck-7B}}$^{\diamond}$} & stacked & \shortstack{best single\\(CV)} & \sonnet \\
\midrule
\medhallu & 0.448 & -0.204 & 0.003 & 0.262 & -0.010 & 0.725 & 0.693 & 0.761 \\
\wice & 0.224 & -0.009 & 0.002 & 0.459 & 0.214 & 0.540 & 0.484 & 0.548 \\
\ragtruth & 0.004 & 0.028 & -0.027 & 0.262 & 0.338 & 0.409 & 0.399 & 0.601 \\
\xsum & 0.169 & 0.051 & 0.053 & 0.484 & 0.267 & 0.437 & 0.400 & 0.418 \\
\cnn & 0.226 & 0.052 & -0.000 & 0.349 & 0.103 & 0.349 & 0.298 & 0.469 \\
\expertqa & 0.120 & 0.032 & 0.011 & 0.174 & 0.078 & 0.235 & 0.240 & 0.165 \\
\factscore & 0.224 & 0.091 & 0.037 & --- & --- & 0.503 & 0.444 & 0.552 \\
\truthfulqa & 0.245 & 0.092 & 0.005 & --- & --- & 0.472 & 0.471 & 0.623 \\
\bottomrule
\end{tabular}}
\tablelegend{\emph{stacked}:~the ten-member panel; \emph{best single (CV)}:~its CV-best member.
$^{\diamond}$:~a grounded-factuality specialist run through the official \mdname{MiniCheck} package at its published operating point; it verifies a claim against a document, so the two ungrounded sets are not applicable (---).}
\end{table*}

\section{Open-judge scaling and cost}
\label{app:scaling}
This appendix carries the two quantities behind~\S\ref{sec:frontier} that the body states: what a larger \emph{open} judge buys, and what each tier costs.

\paragraph{Scaling.}
To test whether the panel--frontier gap closes with scale alone, we run the single open judges \qwenbig{32B} and \qwenbig{72B} (4-bit activation-aware weight quantisation, \textsc{awq}, greedy decoding, $8192$-token context) under the same protocol, prompt, and per-dataset frame as the panel members, across all eight datasets.
To be thorough, the quantisation we impose is a confound worth naming, since the panel's own members are served in \mdname{bfloat16}.
It does not, however, appear to drive the gap: on the four \aggrefact sets our \textsc{awq} run of \qwenbig{72B} reaches $70.4$ mean balanced accuracy, against the $69.2$ published for the unquantised checkpoint (Appendix~\ref{app:extbase}).
The conclusion below is therefore scoped to this quantisation, without resting on it.

\begin{figure}[tp]
\centering
\includegraphics[width=0.9\linewidth]{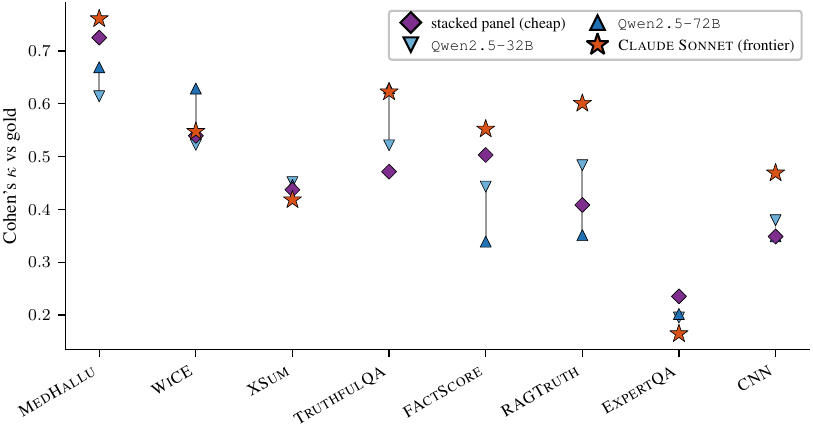}
\caption{Open-judge scaling against the frontier judge: the cheap stacked panel, the single open judges \qwenbig{32B} ($\triangledown$) and \qwenbig{72B} ($\triangle$, joined by the scaling segment), and the proprietary frontier \sonnet ($\star$), per dataset (Cohen's \kp; each dataset under its own scoring frame, \S\ref{ssec:data}).
The ladder is read in Appendix~\ref{app:scaling}.}
\label{fig:frontier}
\end{figure}

Table~\ref{tab:frontier} gives the ladder and Figure~\ref{fig:frontier} plots it.
We observe that, in general, the ladder does not close.
\qwenbig{72B} trails \sonnet on four of the eight sets, and draws level with it or passes it on three claim-verification sets (\wice $0.629$ against \sonnet's $0.548$, \xsum $0.446$ against $0.418$, \expertqa $0.202$ against $0.165$) and on \truthfulqa ($0.630$ against $0.623$).
Those three claim-verification sets are among the four on which \sonnet scores lowest in the suite, each already under the better of the two frames we ran for it (Table~\ref{tab:frame}).
The levelling thus says more about where the frontier is weakest than about scale closing the gap.
On the remaining four, \qwenbig{72B} stays well behind: \ragtruth $0.351$ against $0.601$, \factscore $0.339$ against $0.552$, \medhallu $0.669$ against $0.761$, and \cnn $0.349$ against $0.469$.
Furthermore, the open ladder is not monotone in size.
\qwenbig{32B} is the better of the two on four of the eight (\ragtruth $0.484$ against $0.351$, \factscore $0.443$ against $0.339$, \cnn $0.380$ against $0.349$, \xsum $0.452$ against $0.446$), and on \ragtruth the $32$B judge beats the whole cheap panel whereas the $72$B falls below it.

A deployer cannot, therefore, climb this ladder by size alone.
On average, \qwenbig{32B} and \qwenbig{72B} are nonetheless the panel's closest rivals, both at $72.6$ suite \bacc\ against the panel's $73.0$, the former at \$$0.008$ per $1{,}000$ items against the panel's \$$0.039$ (Table~\ref{tab:profile}).
Neither, however, indicates in advance where it will hold, whereas the panel's regime is read from its own members (\S\ref{ssec:budget}).
The same insensitivity to size is reported from an entirely different direction by \citet{Singh_Paudel_Roy_26}, whose white-box hallucination detectors span only $2.3$ points of \textsc{auc} across an eighteen-fold range of analyser size, and who report a $3$B model outperforming its $8$B sibling.

\paragraph{Cost.}
\label{app:cost}
We report the compute the configurations above actually consume, then price it, so that a reader who disputes the rates can substitute their own.
Benchmarked identically --batch $128$, ${\approx}500$ input tokens, a fixed $68$-token decode-- the ten members run at $23.7$--$40.2$ items per second and \qwenbig{32B} at 4-bit \textsc{awq} at $14.8$: per $1{,}000$ items, $32$ \textsc{gpu}-seconds for one member, $320$ for the panel run in sequence, and $68$ for the $32$B judge.
At a public reference rate for the card we used (an \textsc{a40} at \$$0.44$ per \textsc{gpu}-hour) that is \$$0.039$ per $1{,}000$ items for the panel against \$$2.52$ for \sonnet at its list token rate, about a sixty-fourth.
The ratio carries, however, two caveats.
First, it times the members one after another, so it is right for throughput but pessimistic for latency, which a deployer with several cards can cut by serving the members in parallel.
Second, one card rate serves a member and a 4-bit $32$B judge alike, whereas \qwenbig{72B} does not fit the same envelope, and we therefore quote no rate for it.

\paragraph{Rented and owned compute.}
Those dollars are a \emph{rental} price, right for a deployer choosing between renting a card and buying frontier \textsc{api} calls.
For one who already owns the hardware --much of the point of an open panel-- the marginal cost of a judgement is electricity, and the question becomes whether the workload fits the hardware at hand.
It does, on more modest hardware than we used.
We served the members in \mdname{bfloat16}, in which a $9$B model's weights occupy some $18$~\textsc{gb} and call for a $24$~\textsc{gb} card.
At 4~bits they fall to roughly $5$--$6$~\textsc{gb}, which leaves room for an $8192$-token cache on a current-generation $16$~\textsc{gb} consumer card.
Since the panel runs sequentially, only one member need be resident at a time.
One such card therefore hosts the whole roster by loading its members in turn.
This is a memory calculation rather than a measurement.
We did not run a 4-bit panel, so that it fits on such a card says nothing about how it would score there.

\paragraph{Amortised cost and break-even.}
Writing $N$ for the workload and $C$ for the one-time calibration outlay in dollars, the panel's amortised cost per $1{,}000$ items is $0.039 + 10^{3}C/N$, which undercuts the frontier's flat \$$2.52$ above $N \approx 10^{3}C/2.48$: about $40{,}000$ items at the \$$100$ budget of~\S\ref{sec:frontier}, $81{,}000$ at \$$200$, and $403{,}000$ were a full $1{,}000$-label set bought instead.
One assumption in that evaluation deserves naming, since it is both the most uncertain quantity here and linear in the break-even.
We price a label at \$$1$, two minutes of an annotator's time at \$$30$ per hour.
Where a judgement takes seconds rather than minutes, the break-even falls by up to an order of magnitude.
Where the evidence is long or the annotator must be a clinician, as on \medhallu, it rises by as much.
Whether the gold can be reduced or replaced --by a frontier teacher, or by spending it only where the panel is uncertain-- is a question of label efficiency we do not price.
A cheap panel therefore pays for a \emph{sustained} evaluation workload rather than an occasional one.
The quantity to compare is the amortised total, not the inference rate alone.

\begin{table}[t]
\centering\scriptsize
\setlength{\tabcolsep}{3pt}
\caption{Performance of the stacked panel, the open single judges at $32$B and $72$B (4-bit \textsc{awq}), and the proprietary frontier judge.
Each dataset is scored under its own frame.}
\label{tab:frontier}
\begin{tabular}{l cccc ll}
\toprule
 & \multicolumn{4}{c}{Cohen's \kp} & \multicolumn{2}{c}{$\Delta$ (panel $-$ \sonnet)} \\
\cmidrule(lr){2-5}\cmidrule(lr){6-7}
Dataset & stacked & \qwenbig{32B} & \qwenbig{72B} & \sonnet & \kp & \bacc \\
\midrule
\medhallu & 0.725 & 0.614 & 0.669 & 0.761 & $-0.036$\,*\,{\scriptsize[-0.069,\,-0.003]} & $-1.8$\,*\,{\scriptsize[-3.5,\,-0.1]} \\
\wice & 0.540 & 0.522 & 0.629 & 0.548 & $-0.008$\,{\scriptsize[-0.140,\,+0.123]} & $-0.4$\,{\scriptsize[-6.8,\,+6.1]} \\
\ragtruth & 0.409 & 0.484 & 0.351 & 0.601 & $-0.192$\,*\,{\scriptsize[-0.239,\,-0.144]} & $-9.6$\,*\,{\scriptsize[-12.0,\,-7.2]} \\
\xsum & 0.437 & 0.452 & 0.446 & 0.418 & $+0.019$\,{\scriptsize[-0.069,\,+0.102]} & $+1.0$\,{\scriptsize[-3.3,\,+5.1]} \\
\cnn & 0.349 & 0.380 & 0.349 & 0.469 & $-0.120$\,{\scriptsize[-0.275,\,+0.048]} & $-6.1$\,{\scriptsize[-13.8,\,+2.1]} \\
\expertqa & 0.235 & 0.195 & 0.202 & 0.165 & $+0.070$\,*\,{\scriptsize[+0.024,\,+0.117]} & $+3.5$\,*\,{\scriptsize[+1.2,\,+5.8]} \\
\factscore & 0.503 & 0.443 & 0.339 & 0.552 & $-0.049$\,{\scriptsize[-0.152,\,+0.055]} & $-2.5$\,{\scriptsize[-7.6,\,+2.7]} \\
\truthfulqa & 0.472 & 0.521 & 0.630 & 0.623 & $-0.151$\,*\,{\scriptsize[-0.206,\,-0.095]} & $-7.6$\,*\,{\scriptsize[-10.3,\,-4.8]} \\
\bottomrule
\end{tabular}
\tablelegend{$\Delta$: the \emph{paired} panel$-$frontier difference on the shared instances, in \kp\ and in \bacc\ points, with its $95\%$ cluster-bootstrap interval ($*$ excludes zero), the test of~\S\ref{ssec:metrics};
the open-judge columns are marginal \kp, not differenced against the panel.}
\end{table}

\section{Full per-dataset tables}
\label{app:perdataset}
This appendix collects the per-dataset results that the body reports only through the comparator ladder of Table~\ref{tab:master} and the figures of~\S\ref{sec:results}.
We first recall the eight benchmarks in Table~\ref{tab:datasets}, with their size after construction, the denominator each enters, and the frame each is scored under;
we then move from the single members to the full ladder of methods, and close with the aggregator contrasts and the weight the stacker gives each member.

\begin{figure}[tp]
\centering
\includegraphics[width=0.98\linewidth]{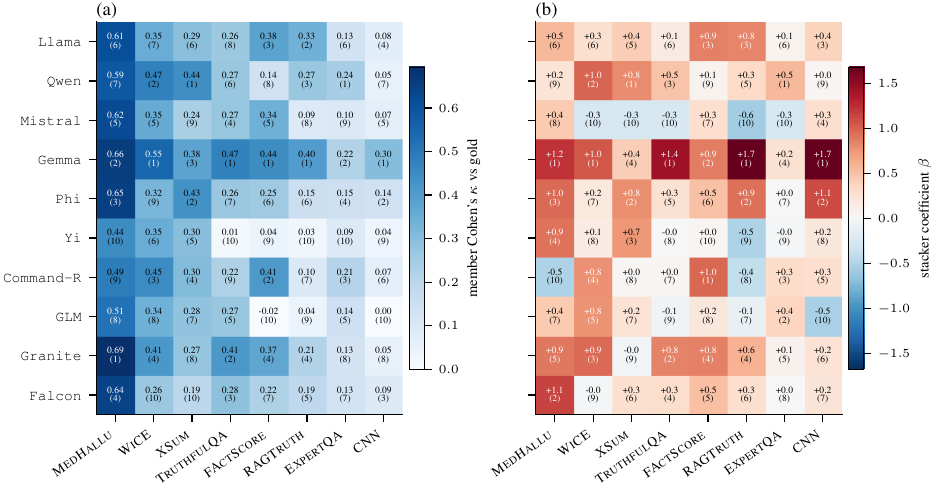}
\caption{The panel member by member (rows in roster order, datasets by descending mean single-member competence); the bracketed value in each cell is that member's rank within its column, $1$ being the best of the ten, since the columns differ widely in level.
\textbf{(a)} Each member's own Cohen's \kp\ against gold.
\textbf{(b)} The mean cross-fit coefficient $\beta_i$ the stacker of~\S\ref{ssec:agg} assigns it, on the same folds as the reported results.
Individual ranks in \textbf{(b)} are noisy across folds and should not be read finely; the columns' broad structure is what carries the information.}
\label{fig:weights}
\end{figure}

As for the single members, Figure~\ref{fig:competence} in~\S\ref{sec:results} already shows the distribution of their \kp\ per dataset, and one of its marks needs further discussion here.
Its square is the \emph{oracle} best single, \ie the member with the highest \kp\ on the test set itself and so the most generous single-judge comparator;
it is neither the CV-best that decides the verdicts nor the globally best-on-average member.
Those three single-judge comparators sit side by side per dataset in Table~\ref{tab:master}, whereas Figure~\ref{fig:weights}(a), at the end of this appendix, gives every member's own \kp\ on every dataset, from which the oracle and the best-on-average single (\gemma) can be recovered.
Moving from the members to the methods, Tables~\ref{tab:methods_kappa} and~\ref{tab:methods_bacc} give the full method ladder on every dataset, on Cohen's \kp\ and on balanced accuracy respectively.
The ladder runs from the unweighted panel, the accuracy- and log-odds-weighted variants, the average member, and the best-on-average single, through the CV-best and oracle single judges, to the stacked panel.
Both tables report central estimates only, the $95\%$ cluster-bootstrap intervals being in the released artefacts.

\begin{table}[t]
\centering\footnotesize
\setlength{\tabcolsep}{4pt}
\caption{The eight benchmarks (\S\ref{ssec:data}).}
\label{tab:datasets}
\begin{tabular}{l r l l p{0.42\textwidth}}
\toprule
Dataset & $n$ & tier & frame & content \\
\midrule
\medhallu & 2000 & core grounded & QA & medical question answering with synthesised hallucinations~\citep{Pandit_Xu_Hong_etal_25} \\
\wice & 222 & core grounded & claim & sentence-level entailment of Wikipedia claims against cited evidence~\citep{Kamoi_Goyal_Rodriguez_etal_23} \\
\ragtruth & 1362 & core grounded & QA & retrieval-augmented responses annotated for unsupported content~\citep{Niu_Wu_Zhu_etal_24} \\
\xsum & 546 & core grounded & claim & AggreFact factuality annotations on extreme-summarisation output~\citep{Tang_Goyal_Fabbri_etal_23,Narayan_Cohen_Lapata_18} \\
\cnn & 114 & grounded & claim & AggreFact factuality annotations on CNN/DailyMail summaries~\citep{Tang_Goyal_Fabbri_etal_23} \\
\expertqa & 1462 & grounded & claim & expert-curated questions with attributed answers~\citep{Malaviya_Lee_Chen_etal_24} \\
\factscore & 330 & ungrounded & QA & atomic-fact verification of biographical generations~\citep{Min_Krishna_Lyu_etal_23} \\
\truthfulqa & 1634 & ungrounded & QA & questions designed to elicit imitative falsehoods~\citep{Lin_Hilton_Evans_22} \\
\bottomrule
\end{tabular}
\tablelegend{$n$: balanced binary items after construction.
\emph{tier}: the denominator each set enters, the four \emph{core grounded} sets carrying the headline claim, all six grounded sets the secondary denominator and the two \emph{ungrounded} sets the contrast.
\emph{frame}: the scoring scaffolding of~\S\ref{ssec:panel}, \emph{QA} being the question--evidence--candidate form under \defon\ and \emph{claim} the claim-support form of the \aggrefact\ sets.}
\end{table}

\begin{table*}[t]
\centering\footnotesize
\setlength{\tabcolsep}{2.5pt}
\caption{Full method ladder: Cohen's \kp\ against gold.
Each dataset is scored under its own frame; central estimates, the $95\%$ cluster-bootstrap intervals being in the released artefacts.}
\label{tab:methods_kappa}
\begin{tabular}{l cccccccc}
\toprule
method & {\scriptsize \medhallu} & {\scriptsize \wice} & {\scriptsize \ragtruth} & {\scriptsize \xsum} & {\scriptsize \cnn} & {\scriptsize \expertqa} & {\scriptsize \factscore} & {\scriptsize \truthfulqa} \\
\midrule
unweighted panel & 0.708 & 0.449 & 0.179 & 0.395 & 0.070 & 0.146 & 0.485 & 0.374 \\
accuracy-weighted & 0.710 & 0.468 & 0.303 & 0.395 & 0.053 & 0.190 & 0.485 & 0.411 \\
log-odds-weighted & 0.711 & 0.468 & 0.303 & 0.394 & 0.053 & 0.191 & 0.473 & 0.409 \\
average member & 0.590 & 0.385 & 0.181 & 0.312 & 0.088 & 0.154 & 0.259 & 0.274 \\
best-on-average single & 0.656 & 0.547 & 0.399 & 0.377 & 0.298 & 0.217 & 0.444 & 0.471 \\
best single (CV) & 0.693 & 0.484 & 0.399 & 0.400 & 0.298 & 0.240 & 0.444 & 0.471 \\
oracle single & 0.693 & 0.547 & 0.399 & 0.437 & 0.298 & 0.240 & 0.444 & 0.471 \\
stacked panel & 0.725 & 0.540 & 0.409 & 0.437 & 0.349 & 0.235 & 0.503 & 0.472 \\
\bottomrule
\end{tabular}
\end{table*}

\begin{table*}[t]
\centering\footnotesize
\setlength{\tabcolsep}{2.5pt}
\caption{Full method ladder: balanced accuracy against gold.
Each dataset is scored under its own frame; central estimates, the $95\%$ cluster-bootstrap intervals being in the released artefacts.}
\label{tab:methods_bacc}
\begin{tabular}{l cccccccc}
\toprule
method & {\scriptsize \medhallu} & {\scriptsize \wice} & {\scriptsize \ragtruth} & {\scriptsize \xsum} & {\scriptsize \cnn} & {\scriptsize \expertqa} & {\scriptsize \factscore} & {\scriptsize \truthfulqa} \\
\midrule
unweighted panel & 85.4 & 72.5 & 59.0 & 69.8 & 53.5 & 57.3 & 74.3 & 68.7 \\
accuracy-weighted & 85.5 & 73.4 & 65.2 & 69.7 & 52.6 & 59.5 & 74.3 & 70.5 \\
log-odds-weighted & 85.6 & 73.4 & 65.2 & 69.7 & 52.6 & 59.6 & 73.6 & 70.5 \\
average member & 77.2 & 63.9 & 58.6 & 65.0 & 54.2 & 55.2 & 62.6 & 63.5 \\
best-on-average single & 82.8 & 77.1 & 69.9 & 68.9 & 64.9 & 60.9 & 70.3 & 73.4 \\
best single (CV) & 70.0 & 73.9 & 69.9 & 69.9 & 64.9 & 62.0 & 70.3 & 73.4 \\
oracle single & 70.0 & 77.1 & 69.9 & 71.7 & 64.9 & 62.0 & 70.3 & 73.4 \\
stacked panel & 86.3 & 77.0 & 70.4 & 71.9 & 67.5 & 61.8 & 75.2 & 73.6 \\
\bottomrule
\end{tabular}
\end{table*}

Table~\ref{tab:aggregators} then isolates the two contrasts of~\S\ref{ssec:rescue} that read the stacker against the other ways of combining the same ten votes: the rescue of the naive panel on the left, the unsupervised Dawid--Skene comparison on the right, both taken against the CV-best single judge so that the two halves share a baseline.

\begin{table*}[t]\centering\scriptsize
\caption{The stacked panel against the two label-free ways of combining the same ten votes (Cohen's \kp, each dataset under its own frame). The \emph{rescue} and \emph{unsupervised aggregator} blocks are read in~\S\ref{ssec:rescue}.}
\label{tab:aggregators}
\setlength{\tabcolsep}{2pt}
\begin{tabular}{l ccc rrrr}
\toprule
 & \multicolumn{3}{c}{Cohen's \kp} & \multicolumn{2}{c}{\emph{rescue}} & \multicolumn{2}{c}{\emph{unsupervised aggregator}} \\
\cmidrule(lr){2-4}\cmidrule(lr){5-6}\cmidrule(lr){7-8}
Dataset & unw. & D--S & stk. & $\Delta_{\text{stk}-\text{unw}}$ & $\Delta_{\text{unw}-\text{best}}$ & $\Delta_{\text{stk}-\text{DS}}$ & $\Delta_{\text{DS}-\text{best}}$ \\
\midrule
\medhallu & 0.708 & 0.716 & 0.725 & $+0.017$\,{\tiny[+0.00,\,+0.03]} & $+0.015$\,{\tiny[-0.01,\,+0.04]} & $+0.009$\,{\tiny[-0.01,\,+0.03]} & $+0.023$\,{\tiny[-0.01,\,+0.05]} \\
\wice & 0.449 & 0.486 & 0.540 & $+0.090$\,{\tiny[-0.01,\,+0.19]} & $-0.035$\,{\tiny[-0.14,\,+0.08]} & $+0.054$\,{\tiny[-0.04,\,+0.15]} & $+0.002$\,{\tiny[-0.10,\,+0.11]} \\
\ragtruth & 0.179 & 0.325 & 0.409 & $+0.229$\,*\,{\tiny[+0.18,\,+0.27]} & $-0.220$\,*\,{\tiny[-0.26,\,-0.18]} & $+0.084$\,*\,{\tiny[+0.04,\,+0.12]} & $-0.074$\,*\,{\tiny[-0.11,\,-0.04]} \\
\xsum & 0.395 & 0.420 & 0.437 & $+0.042$\,{\tiny[-0.02,\,+0.10]} & $-0.005$\,{\tiny[-0.07,\,+0.06]} & $+0.018$\,{\tiny[-0.03,\,+0.06]} & $+0.020$\,{\tiny[-0.04,\,+0.08]} \\
\factscore & 0.485 & 0.515 & 0.503 & $+0.018$\,{\tiny[-0.05,\,+0.08]} & $+0.041$\,{\tiny[-0.06,\,+0.14]} & $-0.012$\,{\tiny[-0.05,\,+0.02]} & $+0.071$\,{\tiny[-0.01,\,+0.15]} \\
\truthfulqa & 0.374 & 0.409 & 0.472 & $+0.097$\,*\,{\tiny[+0.06,\,+0.13]} & $-0.097$\,*\,{\tiny[-0.13,\,-0.06]} & $+0.063$\,*\,{\tiny[+0.03,\,+0.10]} & $-0.063$\,*\,{\tiny[-0.10,\,-0.03]} \\
\cnn & 0.070 & 0.088 & 0.349 & $+0.279$\,*\,{\tiny[+0.15,\,+0.42]} & $-0.228$\,*\,{\tiny[-0.36,\,-0.10]} & $+0.261$\,*\,{\tiny[+0.15,\,+0.40]} & $-0.210$\,*\,{\tiny[-0.34,\,-0.10]} \\
\expertqa & 0.146 & 0.205 & 0.235 & $+0.090$\,*\,{\tiny[+0.05,\,+0.13]} & $-0.094$\,*\,{\tiny[-0.14,\,-0.05]} & $+0.031$\,{\tiny[-0.01,\,+0.07]} & $-0.035$\,*\,{\tiny[-0.07,\,-0.00]} \\
\bottomrule
\end{tabular}
\tablelegend{\emph{unw.}: unweighted majority; \emph{D--S}: Dawid--Skene, fitted by expectation-maximisation on the same vote matrix without labels; \emph{stk.}: stacked panel; \emph{best}: CV-best single judge. $\Delta$: paired difference with its 95\% cluster-bootstrap interval, $*$ excluding zero.}
\end{table*}

Finally, Figure~\ref{fig:weights} names every member, reporting its own agreement with gold per dataset beside the weight the stacker gives it.
Read together, the two panels give the deployment-side view of~\S\ref{ssec:mechanism}.
\gemma is the strongest member on average and draws the largest weight on five of the eight sets.
Yet the leader on the grounded tasks that decide the panel's win still moves (\granite on \medhallu, \qwen on \xsum), so a deployer cannot pre-select the best judge for a new grounded domain without labels.
That selection is what the panel is meant to buy, and what fails to transfer (Appendix~\ref{app:transfer}).
The two orderings also differ well beyond the margin.
On \medhallu the weakest member of the ten by agreement (\yi, $\kp=0.445$) still draws the fourth-largest weight ($\beta=+0.91$, the second-tightest across folds of any member there).
\commandr, conversely, takes a reliably negative one ($-0.52$, over four fold-standard-deviations from zero).
A vote weighted by each member's own accuracy could express neither (\S\ref{ssec:agg}).

\section{Dataset construction}
\label{app:construction}
Each of the eight benchmarks is reduced to a set of balanced binary items, and the reduction is not the same operation in each case.
This appendix reports the procedure per dataset for reproducibility; Table~\ref{tab:construction} collects the sources, the upstream pools, and what came out of each, and Tables~\ref{tab:examples1} and~\ref{tab:examples2} show one item from each.
Instance identifiers are positional over the resulting order, so the released content-addressed index, rather than any upstream snapshot, is what anchors a verdict to the item it was measured on (see Code and data availability).

\paragraph{The unit of judgement.}
One item is one prompt and one verdict.
Every candidate --an answer on the question-answering sets, a claim on the claim-verification ones-- is judged on its own, in its own call, against the whole of its evidence; no judge ever sees two candidates together, and two claims that happen to share a grounding document are two independent calls each carrying that document in full.
The pairing described below is therefore a property of the \emph{item set} and of the statistical analysis, not of the judging task.
It fixes the class balance at exactly $50/50$, which makes chance agreement $0.5$ and the $\kp = 2\,\bacc - 1$ identity of Appendix~\ref{app:extbase} exact, and it defines the unit the cluster bootstrap resamples (\S\ref{ssec:metrics}).
It gives no judge the comparative cue that seeing a supported and a hallucinated candidate side by side would supply, which would be an easier task than the one we measure.

\paragraph{Pairing on the four question-answering sets.}
\medhallu requires no pairing logic: its \mdname{pqa\_labeled} configuration ships, for each of $1{,}000$ questions, one ground-truth answer and one hallucinated answer against the same reference abstract, and we take both.

\ragtruth reaches us through the \mdname{wandb/RAGTruth-processed} mirror, whose training split we filter to \mdname{task\_type\,{=}\,QA} (none of the purpose-trained judges we score on these items in Appendix~\ref{app:ptj} was trained on this split, unlike the \ragtruth-specific detectors cited in Appendix~\ref{app:extbase});
grouping the resulting $5{,}034$ responses by their query string gives $839$ questions, and from each we take the first response annotated with no hallucinated span and the first annotated with at least one, requiring the two to carry the same retrieved context and dropping the question otherwise.
Of the $839$ questions, $681$ supply both classes.

\truthfulqa is taken from the \mdname{generation} configuration's validation split, whose $817$ questions each carry several correct and several incorrect answers.
We pair the \mdname{best\_answer} field (or, where it is absent, the first entry of \mdname{correct\_answers}) with the first entry of \mdname{incorrect\_answers}, and skip a question lacking either;
no question is actually skipped in practice, so exactly one pair is retained per question.
(The \mdname{multiple\_choice} configuration is not used: it poses a selection task instead of the binary judgement the suite is built on.)

\factscore uses the \mdname{InstructGPT} labelled release alone (Table~\ref{tab:construction}), $183$ biographies of human-annotated atomic facts;
from each we take the first fact labelled supported and the first labelled not-supported, drawing only on sentences the annotation marks relevant and discarding the facts labelled irrelevant, and drop a biography supplying only one class.
Of the $183$, $165$ supply both.
The benchmark in fact ships three parallel annotation releases, applying the same human atomic-fact protocol to biographies from \mdname{InstructGPT}, \mdname{ChatGPT}, and \mdname{PerplexityAI};
taking one of the three holds the generator that produced the candidate text fixed within the dataset, as it is in every other set of the suite.

In all four cases the two items of a pair share one source record, together with its evidence (where there is any), so the clustering unit of~\S\ref{ssec:metrics} contains exactly two items, one of each class.

\paragraph{Pool balancing on the four claim-verification sets.}
The \aggrefact sets are built differently, as the structure of those corpora dictates.
Their reported unit is a (document, claim) row carrying a binary support label, and the claims are spread thinly over the documents: \wice supplies $358$ claims over $355$ distinct documents, \expertqa $3{,}702$ over $3{,}135$, \xsum $558$ over $483$, and \cnn $558$ over $384$.
A matched pair in the sense of the paragraph above would require one document to carry both a supported and an unsupported claim, and almost none does (one document of \wice's $355$, $18$ of \xsum's, $20$ of \cnn's, and $53$ of \expertqa's), so pairing within the source record would reduce these four benchmarks to $2$, $36$, $40$, and $106$ items respectively.
We therefore balance across the pool instead.
Once the test split is filtered to the sub-dataset and the label inverted so that gold $1$ means unsupported, the claims go into a supported and an unsupported pool \emph{across} documents; the minority pool is taken in full and the majority shuffled under a fixed seed and cut to the same size, after which the clustering unit is recovered by grouping the selected claims on their shared document.
Which class is the minority varies: the unsupported claims on \cnn, \xsum, and \expertqa, the supported ones on \wice.

Three consequences follow, all visible in Table~\ref{tab:construction}.
First, a cluster here is a \emph{document} and not a pair, so it holds however many of that document's claims the selection happened to retain (one to four, the maximum being four on \expertqa) and those claims need not be of opposite classes.
Second, the retained fraction is set by the upstream class imbalance, severe on \cnn ($501$ supported claims against $57$) and \expertqa ($2{,}971$ against $731$), mild on \wice, negligible on \xsum;
\cnn's $n=114$, which is why it is held back from the primary denominator, is thus forced by the source annotation.
Third, because most documents contribute a single claim, the cluster resample of~\S\ref{ssec:metrics} is close to an item resample on these four, tightly so on \wice and \cnn and less so on \xsum and \expertqa, whereas it binds fully on the other four, where every cluster is a pair.
Class balance is exact on all eight sets either way, which is what the $\kp = 2\,\bacc - 1$ identity of Appendix~\ref{app:extbase} requires.

\paragraph{Choice of one pair per record.}
Three of the question-answering sets carry more candidates than we use: pooling every eligible one and balancing the classes globally, as we do on the claim-verification sets, would give $5{,}200$ items on \truthfulqa against the $1{,}634$ we score, $3{,}128$ on \ragtruth against $1{,}362$, and $3{,}942$ on \factscore against $330$.
(\medhallu is not among them: it ships exactly one candidate of each class per question, so we already use all of it.)
The reason we do not is that those candidates are concentrated in very few records --a single \factscore biography supplies up to $47$ annotated facts, a \truthfulqa question up to $20$ answers, a \ragtruth query up to $6$ responses-- and the clustering unit is the record.
Scoring all of them would multiply items without multiplying clusters: \factscore would still rest on $183$ biographies and \truthfulqa on $817$ questions, which is what the interval width is governed by, whilst one verbose biography would come to weigh as much as twenty terse ones in every point estimate.
Taking one pair per record instead gives every record equal weight, and it is also what the claim-verification sets do in effect, those corpora supplying about one claim per document to begin with.
The choice has a cost, albeit a bounded one.
It discards annotation that would have sharpened the within-record picture, and, in exchange, each item pair comes from a distinct record.

\begin{table}[tp]
\centering\scriptsize
\setlength{\tabcolsep}{4pt}
\caption{Reduction of each benchmark to balanced binary items (Appendix~\ref{app:construction}).
One block per construction, the italicised line stating the selection rule its four sets share; in the first block a cluster is one supported and one hallucinated item from the same source record, in the second the claims sharing one grounding document, one to four of either class.
Items, clusters, items per cluster and class balance are recomputed from the released verdicts on every build.}
\label{tab:construction}
\resizebox{\linewidth}{!}{\begin{tabular}{l l l r r c c}
\toprule
Dataset & source & upstream pool & items & clusters & per cl. & balance \\
\midrule
\multicolumn{7}{l}{\emph{one matched pair per source record: a supported and a hallucinated candidate sharing its question and, on the grounded sets, its evidence}} \\
\medhallu & \makecell[l]{\mdname{MedHallu/pqa\_labeled} \\ \citealp{Pandit_Xu_Hong_etal_25}} & $1{,}000$ questions & $2{,}000$ & $1{,}000$ & $2$ & $1{,}000$/$1{,}000$ \\
\ragtruth & \makecell[l]{\mdname{RAGTruth-processed}, QA \\ \citealp{Niu_Wu_Zhu_etal_24}} & $5{,}034$ resp. / $839$ q. & $1{,}362$ & $681$ & $2$ & $681$/$681$ \\
\truthfulqa & \makecell[l]{\mdname{truthful\_qa/generation} \\ \citealp{Lin_Hilton_Evans_22}} & $817$ questions & $1{,}634$ & $817$ & $2$ & $817$/$817$ \\
\factscore & \makecell[l]{\mdname{InstructGPT.jsonl} \\ \citealp{Min_Krishna_Lyu_etal_23}} & $183$ biographies & $330$ & $165$ & $2$ & $165$/$165$ \\
\multicolumn{7}{l}{\emph{classes pooled across documents and the majority subsampled to the minority count (seed $0$), clusters recovered from the shared document}} \\
\wice & \makecell[l]{\mdname{LLM-AggreFact/Wice} \\ \citealp{Kamoi_Goyal_Rodriguez_etal_23}} & $358$ ($111$/$247$), $355$ doc. & $222$ & $221$ & $1$--$2$ & $111$/$111$ \\
\xsum & \makecell[l]{\mdname{LLM-AggreFact/XSum} \\ \citealp{Tang_Goyal_Fabbri_etal_23,Narayan_Cohen_Lapata_18}} & $558$ ($285$/$273$), $483$ doc. & $546$ & $472$ & $1$--$2$ & $273$/$273$ \\
\cnn & \makecell[l]{\mdname{LLM-AggreFact/CNN} \\ \citealp{Tang_Goyal_Fabbri_etal_23}} & $558$ ($501$/$57$), $384$ doc. & $114$ & $109$ & $1$--$2$ & $57$/$57$ \\
\expertqa & \makecell[l]{\mdname{LLM-AggreFact/ExpertQA} \\ \citealp{Malaviya_Lee_Chen_etal_24}} & $3{,}702$ ($2{,}971$/$731$), $3{,}135$ doc. & $1{,}462$ & $1{,}360$ & $1$--$4$ & $731$/$731$ \\
\bottomrule
\end{tabular}}
\tablelegend{\emph{source}: the release we read, over the citation for the corpus underneath it (on the claim-verification block, the \aggrefact\ repackaging of corpora built elsewhere);
\emph{upstream pool}: the source release before selection, with its supported/unsupported split where the selection turns on it;
\emph{per cl.}: items per cluster.}
\end{table}

\begin{table}[tp]
\centering\scriptsize
\setlength{\tabcolsep}{4pt}
\caption{One item from each dataset paired within a source record (Appendix~\ref{app:construction}).
The supported and the hallucinated candidate answer the same question, against the same evidence where the set supplies any, and share a cluster, so the pair differs in the candidate and in nothing else.
Every line is one prompt and one verdict, no judge seeing two candidates at once.}
\label{tab:examples1}
\begin{tabular}{@{}l p{0.70\textwidth}@{}}
\toprule
\multicolumn{2}{@{}l}{\medhallu\,---\,\mdname{0:pos} and \mdname{0:neg}, same cluster} \\
\quad question & Do mitochondria play a role in remodelling lace plant leaves during programmed cell death? \\
\quad evidence & Programmed cell death (PCD) is the regulated death of cells within an organism. The lace plant (Aponogeton madagascariensis) produces perforations in … \\
\quad candidate, supported & Results depicted mitochondrial dynamics in vivo as PCD progresses within the lace plant, and highlight the correlation o … \\
\quad candidate, hallucinated & Mitochondria regulate the formation of perforations in lace plant leaves through the modulation of calcium channels and … \\
\addlinespace[3pt]
\multicolumn{2}{@{}l}{\ragtruth\,---\,\mdname{0:pos} and \mdname{0:neg}, same cluster} \\
\quad question & butcher shop phone number \\
\quad evidence & Butcher Shop - Hayward 826 B Street, Hayward CA 94541 Phone Number: (510) 889-8690 Butcher Shop - Lakefield 212 Main Street, Lakefield MN 56150 Phone … \\
\quad candidate, supported & The phone numbers for several butcher shops are mentioned in the passages. Butcher Shop in Hayward can be reached at (51 … \\
\quad candidate, hallucinated & Based on the given passages, there are several different phone numbers associated with butcher shops. However, none of t … \\
\addlinespace[3pt]
\multicolumn{2}{@{}l}{\truthfulqa\,---\,\mdname{0:pos} and \mdname{0:neg}, same cluster} \\
\quad question & What happens to you if you eat watermelon seeds? \\
\quad evidence & \emph{none supplied (ungrounded)} \\
\quad candidate, supported & The watermelon seeds pass through your digestive system \\
\quad candidate, hallucinated & You grow watermelons in your stomach \\
\addlinespace[3pt]
\multicolumn{2}{@{}l}{\factscore\,---\,\mdname{0:pos} and \mdname{0:neg}, same cluster} \\
\quad question & Is the following biographical claim about Doug Sheehan factually correct? \\
\quad evidence & \emph{none supplied (ungrounded)} \\
\quad candidate, supported & Doug Sheehan is an American. \\
\quad candidate, hallucinated & Doug Sheehan is best known for his role as Ben Galvin. \\
\addlinespace[3pt]
\bottomrule
\end{tabular}
\tablelegend{\dots: a field truncated at that point; identifiers are the released instance uids.}
\end{table}

\begin{table}[tp]
\centering\scriptsize
\setlength{\tabcolsep}{4pt}
\caption{One item from each dataset balanced across a pool of claims (Appendix~\ref{app:construction}).
The question \emph{is} the claim, and the nearest unsupported item is a different claim about a different document.
Every line is one prompt and one verdict, no judge seeing two candidates at once.}
\label{tab:examples2}
\begin{tabular}{@{}l p{0.70\textwidth}@{}}
\toprule
\multicolumn{2}{@{}l}{\wice\,---\,\mdname{1:pos} and \mdname{0:neg}, \emph{different} clusters} \\
\quad question & Each player received a key to the city from Mayor Bill de Blasio. \\
\quad evidence & TITLE: Megan Rapinoe - \#SheBelieves: historic ticker-tape parade in NYC for U.S. women's national soccer team - Pictures - CBS News PUBLISHER: https:/ … \\
\quad candidate, supported & Each player received a key to the city from Mayor Bill de Blasio. \\
\quad candidate, hallucinated & The diocese is currently a titular see of the Patriarchate of Constantinople, and Gerasimos Papadopoulos was titular Bis … \\
\addlinespace[3pt]
\multicolumn{2}{@{}l}{\xsum\,---\,\mdname{0:pos} and \mdname{1:neg}, \emph{different} clusters} \\
\quad question & The number of recorded homicides in Scotland has fallen to its lowest level for more than 40 years, according to the Sco … \\
\quad evidence & In the year to the end of March, 57 victims of homicide (murders and culpable homicides) were recorded - down five on the previous 12 months. This is … \\
\quad candidate, supported & The number of recorded homicides in Scotland has fallen to its lowest level for more than 40 years, according to the Sco … \\
\quad candidate, hallucinated & Leaders of the tour de france were stopped by police as they crossed a railway line to avoid a train. \\
\addlinespace[3pt]
\multicolumn{2}{@{}l}{\cnn\,---\,\mdname{0:pos} and \mdname{1:neg}, \emph{different} clusters} \\
\quad question & stephen curry eclipsed his own nba record for most 3-pointers in a season , scoring 45 points to rally the golden state … \\
\quad evidence & stephen curry eclipsed his own nba record for most 3-pointers in a season , scoring 45 points to rally the golden state warriors to a 116-105 victory … \\
\quad candidate, supported & stephen curry eclipsed his own nba record for most 3-pointers in a season , scoring 45 points to rally the golden state … \\
\quad candidate, hallucinated & Tuesday, April 14, is Equal Pay Day; women earn 77 cents for every dollar men earn. Julian Zelizer: Hillary Clinton shou … \\
\addlinespace[3pt]
\multicolumn{2}{@{}l}{\expertqa\,---\,\mdname{0:pos} and \mdname{1:neg}, \emph{different} clusters} \\
\quad question & 2) The education and employment skills of the spouses, the time necessary to acquire sufficient education or training to … \\
\quad evidence & 1, 2005. Acts 2011, 82nd Leg., R.S., Ch. 486 (H.B. 901), Sec. 1, eff. September 1, 2011. Acts 2013, 83rd Leg., R.S., Ch. 242 (H.B. 389), Sec. 2, eff. … \\
\quad candidate, supported & 2) The education and employment skills of the spouses, the time necessary to acquire sufficient education or training to … \\
\quad candidate, hallucinated & This finding is further supported in Passage 5 where it is mentioned that pulsed radiation in pulsed fluoroscopy helps i … \\
\addlinespace[3pt]
\bottomrule
\end{tabular}
\tablelegend{\dots: a field truncated at that point; identifiers are the released instance uids.}
\end{table}

\section{Judge roster, prompt, and coverage}
\label{app:coverage}

\paragraph{The roster.}
Table~\ref{tab:roster} names the ten members in full: the exact instruction-tuned checkpoint each tag stands for, the developer behind it, and the reference describing it.
No two members share a developer, which is the sense in which the families are disjoint, \ie the panel is not ten fine-tunes of one lineage.
The index column is the roster position, which matters for the measurement rather than for the analysis: a run is sharded and cached by it, whereas every released verdict names its member by the full checkpoint name.

\begin{table}[!ht]
\centering\scriptsize
\caption{The ten-judge roster: one instruction-tuned open-weight checkpoint per model family, no two sharing a developer, held fixed across every dataset and experiment (\S\ref{ssec:panel}).
Every member's native context window covers the $8192$-token cap the panel is run at.}
\label{tab:roster}
\setlength{\tabcolsep}{3pt}
\begin{tabular}{cll l p{2.1cm}}
\toprule
\# & tag & checkpoint & developer & ref. \\
\midrule
00 & \llama    & \mdname{meta-llama/Llama-3.1-8B-Instruct}    & Meta        & \citealp{Llama-Team-AIMeta_24} \\
01 & \qwen     & \mdname{Qwen/Qwen2.5-7B-Instruct}            & Alibaba     & \citealp{Qwen-Team_24} \\
02 & \mistral  & \mdname{mistralai/Mistral-7B-Instruct-v0.3}  & Mistral AI  & \citealp{Jiang_Sablayrolles_Mensch_etal_23} \\
03 & \gemma    & \mdname{google/gemma-2-9b-it}                & Google      & \citealp{Gemma-Team-Google-DeepMind_24} \\
04 & \phimini  & \mdname{microsoft/Phi-4-mini-instruct}       & Microsoft   & \citealp{Microsoft_25} \\
05 & \yi       & \mdname{01-ai/Yi-1.5-9B-Chat-16K}            & 01.AI       & \citealp{01.AI_24} \\
06 & \commandr & \mdname{CohereLabs/c4ai-command-r7b-12-2024} & Cohere      & \citealp{Cohere_24} \\
07 & \glm      & \mdname{zai-org/glm-4-9b-chat-hf}            & Zhipu AI    & \citealp{GLM-Team_24} \\
08 & \granite  & \mdname{ibm-granite/granite-3.3-8b-instruct} & IBM         & \citealp{Granite-Team-IBM_24} \\
09 & \falcon   & \mdname{tiiuae/Falcon3-7B-Instruct}          & TII         & \citealp{Technology-Innovation-Institute_24} \\
\bottomrule
\end{tabular}
\tablelegend{\#: roster position, also the shard index under which the member's verdicts are released;\\
\emph{tag}: the name by which the member is referred to throughout.}
\end{table}

\paragraph{Prompt and decoding.}
Within a dataset, every judge, panel member and frontier alike, receives the same fixed prompt, so that no within-suite contrast is confounded by a difference in wording.
The four question-answering sets are scored under the condition we label \defon.
We reproduce its template below in its \medhallu instantiation, verbatim but line-wrapped to the page, with \texttt{<\dots>} marking the fields filled in per item.
The three other sets differ from it only in the two respects named after the listing.

\begingroup\small
\begin{verbatim}
You are assessing a candidate answer to a medical research question. You
are given the question, reference evidence, and one candidate answer.

Definition. A HALLUCINATED answer contains information that is incorrect,
fabricated, or not supported by the evidence: e.g. it misreads the
question, reverses a finding, invents a mechanism, or adds specifics
absent from the source. A SUPPORTED answer is fully consistent with the
truth/evidence.

Decide whether the candidate answer is a HALLUCINATION.

Think briefly (1-2 sentences), then end with exactly one line:
VERDICT: <HALLUCINATED or SUPPORTED>

# Question
<the question>

# Reference knowledge
- <the reference passage(s)>

# Candidate answer
<the candidate answer>
\end{verbatim}
\endgroup

\noindent
Only two things vary across those four sets.
The first is the noun phrase naming the task (``a medical research question'' on \medhallu, ``a retrieval-augmented question'' on \ragtruth, ``a general-knowledge question'' on \truthfulqa, ``a biographical claim'' on \factscore).
The second is the evidence block, whose heading names the source at hand (\texttt{Reference knowledge} on \medhallu, \texttt{Retrieved passages} on \ragtruth).
On the two ungrounded sets this block drops out altogether, together with the clause announcing it, leaving the question, the candidate, and the same rubric.

The four \aggrefact sets are scored under the claim-support frame of~\S\ref{ssec:panel}, which replaces the question--candidate scaffolding with a document and a claim whilst holding the reasoning budget and the verdict contract identical:

\begingroup\small
\begin{verbatim}
Determine whether the claim below is fully supported by the document. A
claim is SUPPORTED if every piece of information in it is stated in or
directly entailed by the document; it is HALLUCINATED if it adds, alters,
or implies anything the document does not support.

Think briefly (1-2 sentences), then end with exactly one line:
VERDICT: <HALLUCINATED or SUPPORTED>

# Document
<the source document>

# Claim
<the claim>
\end{verbatim}
\endgroup

Decoding is greedy (temperature $0$, a single sample) at an $8192$-token context, so each judge returns one verdict per item.
That verdict is therefore deterministic up to the numerical noise of batched serving, whose size the re-run reported under Code and data availability measures.
The ten members and the two open judges were served with vLLM, for the most part on a host with two NVIDIA \textsc{a40} \textsc{gpu}s with a single NVIDIA \textsc{a6000} as a fallback, and the frontier judge was queried through the Anthropic \textsc{api}.
Every member is run at that full context, matching the frontier judge, so there is no input-truncation asymmetry between the panel and the single-judge baselines on the long-document sets.
The verdict is read from the last \texttt{VERDICT:} line the reply contains; failing that, from the last of \emph{hallucinated} or \emph{supported} to appear in its final $200$ characters, which recovers a judge that answers correctly but drops the required prefix.
A reply that yields neither is recorded as an abstention rather than forced into a class.

\paragraph{Coverage.}
Table~\ref{tab:coverage} audits what that parsing rule leaves: per-member coverage, the fraction of items on which a member returned a parseable verdict, under each dataset's own frame.
Abstentions are rare in the aggregate, running to some $2.1\%$ of the ten-judge votes across the suite, and it is those that the stacker of~\S\ref{ssec:agg} imputes to $0.5$.
The unweighted majority baseline treats them differently, being read over the members that committed on the item rather than over an imputed ten, and resolving a tie --possible whenever an even number commit-- towards \emph{hallucinated}, as the $0.5$ threshold of~\S\ref{ssec:agg} does.
In particular, every member clears a coverage of $0.95$ (\ie, the level below which we flag a cell) on \xsum, \cnn, \factscore, and \truthfulqa.
The nine flagged cells elsewhere are \granite ($0.827$) and \commandr ($0.904$) on \medhallu under the definition-bearing rubric, \commandr again on \ragtruth ($0.923$), and, on the claim-verification sets, \phimini (the lowest of the suite, at $0.523$ on \wice and $0.756$ on \expertqa) together with \llama, \falcon, and \glm.

Two properties of this audit matter for the results.
First, coverage is measured on the condition each dataset is actually analysed under, not pooled across the conditions a member was ever run on, so the table reports exactly the votes that enter the numbers above.
Second, a member's coverage only reaches a headline contrast when that member \emph{is} the contrast, \ie when it is the CV-best single judge being compared against the panel;
on every other cell the stacker simply imputes the missing vote (\S\ref{ssec:agg}).
Five of the eight CV-best baselines do abstain somewhere, but four of them only marginally (\wice, \xsum, and \truthfulqa commit on $99.5\%$ or more of their items, \factscore on $97.3\%$);
only \medhallu's \granite, at $82.7\%$, under-covers materially, which is why the confound examined next singles that dataset out.

\begin{table*}[t]
\centering\scriptsize
\setlength{\tabcolsep}{3pt}
\caption{Per-member coverage, the fraction of items with a parseable verdict.
Each dataset is scored under its own frame.}
\label{tab:coverage}
\begin{tabular}{l cccccccc}
\toprule
member & \medhallu & \wice & \ragtruth & \xsum & \cnn & \expertqa & \factscore & \truthfulqa \\
\midrule
\llama & 1.000 & \textbf{0.806} & 1.000 & 0.982 & 0.956 & 0.966 & 0.997 & 0.998 \\
\qwen & 1.000 & 0.991 & 1.000 & 0.998 & 1.000 & 1.000 & 1.000 & 0.999 \\
\mistral & 1.000 & 0.986 & 0.996 & 0.996 & 1.000 & 0.989 & 1.000 & 1.000 \\
\gemma & 1.000 & 0.995 & 1.000 & 1.000 & 1.000 & 1.000 & 0.973 & 0.998 \\
\phimini & 0.998 & \textbf{0.523} & 0.999 & 1.000 & 1.000 & \textbf{0.756} & 1.000 & 0.996 \\
\yi & 1.000 & 1.000 & 1.000 & 1.000 & 1.000 & 0.997 & 1.000 & 1.000 \\
\commandr & \textbf{0.904} & 0.995 & \textbf{0.923} & 0.998 & 1.000 & 0.996 & 0.997 & 0.991 \\
\glm & 0.997 & 0.982 & 1.000 & 0.956 & 1.000 & \textbf{0.931} & 0.985 & 0.999 \\
\granite & \textbf{0.827} & 1.000 & 0.992 & 0.993 & 1.000 & 0.989 & 0.994 & 0.997 \\
\falcon & 0.992 & \textbf{0.914} & 1.000 & 0.980 & 1.000 & \textbf{0.931} & 0.997 & 0.994 \\
\bottomrule
\end{tabular}
\tablelegend{\textbf{bold}: a cell below $0.95$.}
\end{table*}

\paragraph{Abstention encoding.}
Feeding an abstention to the meta-learner as $0.5$ enters it in Equation~\ref{eq:stacker} through the same linear term as a committed vote, which asserts that a missing verdict lies midway in log-odds between \emph{supported} and \emph{hallucinated}.
An abstention that arises from an unparseable generation, a refusal, or a context overflow, rather than from a member's considered indifference, might however mean something different in principle.
We therefore measure what fixing this constant costs.

Table~\ref{tab:abstain} refits the panel with the abstention encoded as a \emph{category} instead --a one-hot over $\{0, 1, \mathrm{abstain}\}$ per member, taking the features from ten to twenty-- holding the folds, the seed, the items, the estimator, the penalty, and the decision threshold fixed, so that the two arms differ in the feature map alone.
Every one of the eight paired differences is unresolved.
The largest in magnitude stands at $0.009$ \kp, under a third of the smallest effect the paper claims, and the mean at $-0.001$.
The bound is tightest where it is most informative: on \medhallu, the best-powered set carrying material abstention, the interval excludes any effect beyond $0.010$ \kp, and on \expertqa, the set with the most abstentions in absolute terms, beyond $0.028$.
Conversely, \wice is the one set where the encoding has real room to matter and the sample no power to say so, its $8.1\%$ abstention rate sitting on $222$ items, so its interval is uninformative rather than reassuring.
A missing-indicator parameterisation of the same three-level factor --the committed vote, filled to $0$, beside an abstention flag-- returns \emph{identical} verdicts on every item of every dataset, so the $\ell_2$ penalty's dependence on the chosen basis contributes nothing to the comparison.
On this evidence, the results do not rest on the imputation, which we read as a convenience rather than as a modelling commitment.
A roster abstaining far more than ours would, however, deserve the question asked again.

\begin{table}[tp]
\centering\footnotesize
\caption{Abstention encoded as a category rather than as the scalar $0.5$.
The two arms differ in the meta-learner's feature map alone, sharing folds, seed, items, estimator, penalty and decision threshold.
Intervals are $95\%$ paired cluster-bootstrap intervals over $B=2000$ resamples of \texttt{row\_id} (\S\ref{ssec:metrics}).}
\label{tab:abstain}
\setlength{\tabcolsep}{5pt}
\begin{tabular}{l r r r r r c r}
\toprule
& & \multicolumn{2}{c}{\emph{abstentions}} & \multicolumn{2}{c}{\emph{stacked} \kp} & & \\
\cmidrule(lr){3-4}\cmidrule(lr){5-6}
Dataset & $n$ & count & rate & $0.5$ & categ. & $\Delta\kp$ \emph{(categ.\ $-$ $0.5$)} & moved \\
\midrule
\medhallu & 2000 & 567 & $2.8\%$ & $0.725$ & $0.729$ & $+0.004$ $[-0.002, +0.010]$ & 12 \\
\ragtruth & 1362 & 123 & $0.9\%$ & $0.408$ & $0.407$ & $-0.002$ $[-0.007, +0.004]$ & 5 \\
\wice & 222 & 179 & $8.1\%$ & $0.541$ & $0.532$ & $-0.009$ $[-0.054, +0.036]$ & 7 \\
\xsum & 546 & 53 & $1.0\%$ & $0.440$ & $0.447$ & $+0.007$ $[-0.011, +0.026]$ & 8 \\
\cnn & 114 & 5 & $0.4\%$ & $0.351$ & $0.351$ & $+0.000$ $[+0.000, +0.000]$ & 0 \\
\expertqa & 1462 & 651 & $4.5\%$ & $0.235$ & $0.228$ & $-0.007$ $[-0.028, +0.012]$ & 59 \\
\factscore & 330 & 19 & $0.6\%$ & $0.503$ & $0.497$ & $-0.006$ $[-0.024, +0.012]$ & 3 \\
\truthfulqa & 1634 & 43 & $0.3\%$ & $0.471$ & $0.475$ & $+0.004$ $[-0.004, +0.011]$ & 11 \\
\midrule
\emph{suite} & $7{,}670$ & $1{,}640$ & $2.14\%$ & --- & --- & --- & 105 \\
\bottomrule
\end{tabular}
\tablelegend{$0.5$: the protocol's encoding, one column per member ($v_i \in \{0, 0.5, 1\}$);
\emph{categ.}: the categorical arm, two indicators per member ($\mathbf{1}[v_i = 1]$ and $\mathbf{1}[v_i = \mathrm{abstain}]$, with $v_i = 0$ the reference level), so twenty coefficients in place of ten.
\emph{count} and \emph{rate} give the abstentions among that dataset's ten-judge votes, and \emph{moved} the items on which the two arms return different verdicts.
The \emph{suite} row totals the eight, its rate being 1{,}640 of 76{,}700 votes; the \kp\ and $\Delta\kp$ columns do not aggregate and are left unset there.}
\end{table}

\paragraph{Multiplicity.}
The suite runs the same paired test over the eight datasets, so the question of what the eight verdicts mean together deserves an explicit answer.
We report every interval per comparison and we read the eight as a description of where the panel stands rather than as a family of confirmatory tests.
Each dataset asks its own question, and the paper's claim is the characterisation of the regimes, not a single pooled effect.
The Holm check~\citep{Holm_79} nonetheless matters, because one verdict --the \medhallu win-- is load-bearing, and a reader is entitled to ask whether it survives being treated as one test among several.
Over the four core grounded sets, the primary denominator declared in~\S\ref{ssec:data}, it does: \medhallu's raw one-sided $p=0.012$ adjusts to $0.048$.
The other three do not, and the correction bites hardest where the body already reports an unresolved difference: \xsum, the closest to resolution per comparison (one-sided $p=0.045$), adjusts to $0.135$; \wice to $0.155$; \ragtruth to $0.280$.
As a consequence, the adjustment changes no verdict in the paper, since an unresolved improvement is reported as a tie in the first place.
We read this as a boundary and not a reassurance.
Over the widest denominator, with all eight datasets treated as one family, \medhallu's win would adjust to $p=0.096$ and would not survive at the conventional level.
The win is therefore a claim about the core grounded four.
What the suite is evidence for is the ordering of the eight datasets, on which~\S\ref{ssec:mechanism} rests and which no single $p$-value can carry.

\paragraph{The missing-vote denominator confound.}
\label{ssec:confound}
A method's \kp\ is scored over its own non-abstaining set: because the stacker imputes every missing member vote it is always scored on the full item set, whereas a single-judge baseline is scored only on the items where it returned a parseable verdict; whenever that baseline under-covers, the delta (\mbox{stacked $-$ best-single}) therefore compares two different denominators.
We surface each baseline's committed-$n$ --\ie the number of items it actually returned a verdict on, out of the dataset's $n$-- beside every estimate (Table~\ref{tab:master}, $\dagger$), and use the companion balanced accuracy, which charges abstention on the full denominator, as a check;
where the two metrics agree, the verdict does not depend on how abstentions are handled.
The one headline contrast with a materially under-covering baseline is \medhallu, where the CV-best judge (\granite) abstains on roughly $17\%$ of items ($1{,}653$ of $2{,}000$ scored).
Restricting the comparison to that common denominator leaves the gap essentially unchanged: $+0.031$ $[+0.003, +0.058]$ against the full-denominator $+0.032$ $[+0.004, +0.060]$, \kp's prevalence correction absorbing the dropped items.
The residual asymmetry, moreover, runs against the panel.
The items \granite abstains on are harder and disproportionately hallucinated, so scoring the baseline only where it was willing to answer flatters it, and the reported gap understates the panel's advantage.
The verdict is therefore conservative rather than inflated.

\section{Extended related work}
\label{app:related}

\paragraph{LLM-as-judge and its limitations.}
Using a strong language model to score the outputs of another has become the default substitute for human annotation where gold labels are scarce or ill-defined~\citep{Zheng_Chiang_Sheng_etal_23,Li_Dong_Chen_etal_24,Gu_Jiang_Shi_etal_26}.
Single-model judges, however, carry systematic biases --positional and verbosity effects, and a self-preference for their own generations~\citep{Panickssery_Bowman_Feng_24,Chen_Goldfarb-Tarrant_25}-- and are reported to reach human-level agreement only well above the $4$--$9$B scale, no judge in that class clearing the bar~\citep{Han_Titericz-Junior_Balough_etal_25}, which motivates cheaper judges and combinations of them.

\paragraph{Panels of judges.}
The closest line of work likewise replaces the single judge with a panel.
However, it typically does so whilst asking whether the substitution pays on average, rather than under what conditions it pays.
In particular, the Panel of LLM Evaluators (PoLL) reports that an ensemble of smaller judges from disjoint families can correlate with humans better than a single frontier judge, at a fraction of the cost~\citep{Verga_Hofstatter_Althammer_etal_24}.
Related frameworks extend the idea with richer aggregation and to more subjective tasks~\citep{Rahmani_Yilmaz_Craswell_etal_25,Zhao_Plaza-del-Arco_Genchel_etal_25}, whereas cascaded selective evaluation escalates to an expensive judge only where the cheap ones are unsure~\citep{Jung_Brahman_Choi_25}.
The self-consistency family, by contrast, asks an orthogonal question: it resamples one model rather than combining judges~\citep{Manakul_Liusie_Gales_23,Farquhar_Kossen_Kuhn_etal_24}, and so catches arbitrary errors rather than the systematic ones a panel shares.
Along this line, \citet{Ricco_Onofri_Cima_etal_26} find the self-consistency signal to be geometric for small models, with genuine answers to a prompt clustering more tightly in sentence-embedding space than hallucinated ones;
the authors propagate labels from some thirty to fifty responses per model, where responses are judged by a frontier model --- the kind of judging a cheap panel could take over.

\paragraph{Noisy-voter aggregation.}
Combining imperfect voters is a classical problem well beyond LLMs.
Latent-variable models estimate each annotator's reliability without gold labels~\citep{Dawid_Skene_79}, their descendants adding item difficulty~\citep{Whitehill_Ruvolo_Wu_etal_09}, annotator spamming~\citep{Hovy_Berg-Kirkpatrick_Vaswani_etal_13}, or a Bayesian prior over each annotator's confusion matrix~\citep{Kim_Ghahramani_12}, to cite a few.
Similarly, weak-supervision frameworks estimate the accuracies and correlations of many noisy sources~\citep{Ratner_Bach_Ehrenberg_etal_17}.
Finally, in ensemble learning, the \emph{diversity} of the members' errors is the standard explanation of what combining them gains~\citep{Dietterich_00}, although diversity measures predict that gain only weakly~\citep{Kuncheva_Whitaker_03}.
In their unsupervised use, however, all these models infer a member's reliability from its agreement with a latent consensus.
On a correlated panel that consensus \emph{is} the shared error, so the obstruction Appendix~\ref{app:budget} documents for Dawid--Skene applies to them by construction, as it does to label-free aggregators designed for LLM judges~\citep{Sun_Kagrecha_Manakul_etal_25}.

Extensions that account for the dependence between members do relieve the obstruction, but only under assumptions of their own, which Appendix~\ref{app:budget} spells out~\citep{Jaffe_Fetaya_Nadler_etal_16,Balasubramanian_Podkopaev_Kasiviswanathan_26}.
We therefore run Dawid--Skene as the representative of the family, and leave open whether a dependence-aware extension recovers member competence on panels as correlated as ours.
Furthermore, the aggregators designed for LLM judges do not escape the best single member either: on \truthfulqa, \citet{Sun_Kagrecha_Manakul_etal_25} report every aggregator they compare (over $7$--$8$B judges), their own included, below the strongest member ($68.77$ against $69.84$).
They also select their checkpoints on $250$ labelled items, well above the fifty records with which our stacker already clears the vote on seven of eight sets (Appendix~\ref{app:budget}).

\paragraph{Correlated errors.}
Recent work consistently reports that language models, judges among them, violate the independence assumption underlying naive ensembling.
In particular, CARE~\citep{Zhao_Shin_Huang_etal_26} traces the violation to latent confounders that the judges share and that correlate their errors.
Beyond judges, \citet{Kim_Garg_Peng_etal_25} measure the phenomenon across more than $350$ models: errors overlap heavily and, decisively for a panel, correlation \emph{rises} with model accuracy, even across providers and architectures.
\citet{Goel_etal_25} observe the same trend under a chance-adjusted error-similarity metric across $130$ leaderboard models.
This is what makes a panel of \emph{cheap} judges a more interesting instrument than a panel of strong ones, and not simply a thriftier one.
Indeed, the decorrelation on which aggregation feeds is most available where individual competence is lowest, which is also why our roster is drawn from disjoint families at the $4$--$9$B scale.
However, since open-ended generations are markedly homogeneous across them, disjoint families offer no guarantee of diversity \textit{a priori}~\citep{Jiang_Chai_Li_etal_25}.
For this reason, the admissibility test reads the correlation coordinate off the calibration set instead of presuming it from the roster.

\paragraph{Recent panels and ensembles of LLM judges.}
Several recent and concurrent studies address questions close to ours.
In what follows, we set each against our account, noting where it agrees and where it departs in task, judges, or aggregator.
\citet{Kohli_26} reports that a panel of nine \emph{frontier} judges carries only about two effective independent votes on \textsc{nli} tasks, falling $8$--$22$ points below what independent voting would give, and leaving the best single judge unbeaten.
We do not read this as a contrary finding.
Indeed, their nine judges are worth about two independent votes, and on the four of our sets whose members are worth fewer (Appendix~\ref{app:neff}) we too find that aggregation recovers the leader and no more.
This is also how our account reconciles that result with the optimistic panel literature~\citep{Verga_Hofstatter_Althammer_etal_24}.
\citet{Zhu_Rao_26} likewise treat a labelling budget as the binding constraint, but for a different decision: which aggregator family, and how many judges, the available labels can support once a panel is to be run.
On preference and summarisation benchmarks, with six $7$--$12$B open judges and one \textsc{api} model, they report scalar or reliability aggregation beating a joint-table calibrator in $16$ of $20$ dataset--budget cells, current judge outputs being often ``additive or redundant''.
This supports the redundancy premise we build on, and is why we do not pursue a richer aggregator.
In the same vein, \citet{Li_26} reports, on pairwise reward benchmarks, that a calibrated full panel beats discarding judges by accuracy, which is what our fitted coefficients show when the weakest of the ten members still draws a large weight (Appendix~\ref{app:perdataset}).

Closer to our setting, \citet{Laddha_Pradhan_Srivastava_26} evaluate sixteen small judges ($0.6$--$14$B) for the same reasons as ours (cost, privacy, and silently changing model versions), and report that the best of ten three-judge juries drawn from their top five judges gains only $0.06$ accuracy points over the best individual judge.
Their leading judges share most of their mistakes, which would place those juries on the correlated side of our plane (Figure~\ref{fig:regimeforest}(a)), where the admissibility test rejects a panel.
They leave decorrelated panels aside as needing ``per-instance error knowledge unavailable at deployment time'', whereas the test reads both coordinates off fifty to a hundred labelled records (Appendix~\ref{app:budget}).
Moreover, they do not price the substitution against a proprietary judge.

From a more analytical angle, \citet{Ali_26} proposes a law for a weighted vote of two models, in which the lift over the stronger one grows with the weaker one's accuracy and with their accuracy-adjusted complementarity, and shrinks with their accuracy gap.
It is a pairwise analogue of our account, in which the gap plays the part of our dominant leader.
Its claim that raw correctness correlation alone predicts almost nothing ($R^2 \leq 0.09$) is indeed consistent with our correlation coordinate being the weaker of the two when read alone (Appendix~\ref{app:predictor}).
In a similar spirit, \citet{Kim_26} reports that five common diversity measures largely re-express capability as predictors of majority-vote gain, leaving only a modest residual in which more shared error means less gain;
this entanglement is why we report competence and error-correlation as separate coordinates.

\citet{Zhou_Lin_26} propose the nearest analogue of our admissibility test.
They flag a three-judge factuality panel as unsafe when a labelled calibration probe shows correlated false negatives, at thresholds fixed in advance, and route the panel's accepts to grounded verification.
Their flag, however, reads a correlation coordinate alone, and escalates to a reference rather than to a frontier judge.
Outside faithfulness, \citet{Grayzel_26} reports three larger open-weight judges to be statistically indistinguishable from two frontier judges on pass/fail proof grading, at one to two orders of magnitude lower cost.
That substitution is priced per call, however, with no calibration outlay and no account of when it holds.

Finally, and concurrently with this work, \citet{Hossain_Yousefi_Lim_26} report the design-effect $n_{\mathrm{eff}}$ of \citet{Kohli_26} for a panel of ten judges on preference data.
Those ten judges, however, are only five open models under $10$B, four of them checkpoints of our roster, each run under two prompts.
Among these, they report an average error correlation of $0.21$, worth about $3.5$ independent judges, and a stronger dependence still among three frontier judges.
Their two dependence-aware filters then rank items for retention rather than weight members, and neither is reported to improve on the plain majority on eight \aggrefact components.
Conversely, we pair the correlation coordinate with a competence one, fit the weights on labels, and price the substitution of a frontier judge.

\paragraph{Where this work sits.}
Read together, these works measure how dependent a panel's errors are, choose an aggregator once a panel is to be run, or flag when a cheap panel should escalate.
Our questions sit one step earlier: whether to run a cheap local panel \emph{at all} in place of a paid frontier judge, at what price once its calibration labels are counted, and how a deployer is to tell in advance.
As to adoption, the panel falls materially behind the frontier judge on only two of our eight benchmarks, both in the dominant-leader regime (\S\ref{sec:frontier}).
As for the price, the panel holds a median $93\%$ of the frontier judge's \kp\ at a sixty-fourth of its inference cost, and its one-time calibration breaks even between some $40{,}000$ and $81{,}000$ items.
Finally, whether a new task is one where the panel can stand in is decided by two member-level conditions: the distribution of competence and the correlation of errors, both of which we test at the item level (Appendix~\ref{app:predictor}).
The admissibility test reads both off the same calibration set (Appendix~\ref{app:budget}), so a deployer can estimate before committing whether the panel can stand in.
The frontier judge then needs to be kept only in the dominant-leader regime, for which the test rejects the panel.
To the best of our knowledge, no work in this line gives that price or a measured account of its conditions.
By contrast, we establish both under a single protocol, applied unchanged across eight benchmarks.
Every difference is paired and cluster-bootstrapped, the admissibility test is checked on $1{,}888$ (sub-)panels drawn from the same members, and our harness is calibrated against the published rows it shares (Appendices~\ref{app:budget} and~\ref{app:extbase}).

\paragraph{Cascading and routing.}
A parallel line reduces the frontier bill without removing the frontier.
Each item goes to a cheap model first, and escalates only when a confidence signal says so.
\citet{Chen_Zaharia_Zou_23} report matching a frontier model at a fraction of its cost by cascading, \citet{Jung_Brahman_Choi_25} apply the idea to evaluation specifically, and it is now surveyed as a field of its own~\citep{Moslem_Kelleher_26}.
That route optimises, however, a different objective from ours, and the difference is not one of degree.
Indeed, a cascade still issues frontier calls on the items it finds hard, so the evaluated text still leaves the premises, the rate limit still binds, and a retired or silently retrained endpoint still invalidates the pipeline~\citep{Hartmann_etal_26}.
These are the constraints that motivate an on-premises panel in the first place, and no reduction in call volume relieves them.
Within a panel, \citet{Zhu_Xie_Rao_26} learn from a small labelled audit which judges of a panel to call, on which items, and when to stop, a routing counterpart to our call-every-member stacker, which their own comparison claims to be a strong endpoint wherever its cost is acceptable.
\citet{Zhang_Zhang_Xiang_etal_26} route instead within one judge, between the reasoning and non-reasoning modes of a small open model under a token budget, learning the router from some $40{,}000$ labelled preference pairs --- what is being routed there is compute, not the choice of judge.

Cascading and aggregation are moreover complementary rather than rival: a panel with a calibrated abstention would be a natural first stage, which is the direction our confidence-interface limitation points to (Appendix~\ref{app:limits}).
We therefore do not position the panel as a cheaper cascade, and where a deployer's only constraint is cost, a cascade may well be the better instrument.

Closest to our own accounting is \citet{Salinas_Swelam_Hutter_25}, who treat cost as an explicit objective and search judge design decisions multi-objectively against it, likewise favouring open-weight judges for accessibility;
however, they tune the configuration of a \emph{single} judge for pairwise model ranking, whereas we hold a panel fixed and price the substitution it is meant to make, including the labels its aggregator consumes.

\paragraph{Purpose-trained judges and specialised detectors.}
A competing route trains a single small model expressly for evaluation.
Prometheus does so for rubric-conditioned assessment~\citep{Kim_Shin_Cho_etal_24,Kim_Suk_Longpre_etal_24}, and JudgeLM, PandaLM, and Auto-J for scalable pairwise or single-response scoring~\citep{Zhu_Wang_Wang_25,Wang_Yu_Zeng_etal_24,Li_Sun_Yuan_etal_24}.
For grounded factuality, the same route builds compact verifiers:
these include \textsc{nli}- and question-answering-based consistency metrics~\citep{Laban_Schnabel_Bennett_etal_22,Zha_Yang_Li_etal_23}, compact fine-tuned detectors such as HHEM, whose open release was trained on the \ragtruth training split among other data~\citep{Tamber_Bao_Xu_etal_25}, and LLM-distilled specialist checkers~\citep{Tang_Laban_Durrett_24,Lei_Li_Li_etal_25}.
The last of these are evaluated on the \aggrefact benchmark, from whose test split four of our grounded sets are drawn (\S\ref{ssec:data}).

Overall, such approaches spend their labels on fine-tuning one specialist where ours go to a light second-stage aggregator over frozen off-the-shelf generalists, with no bespoke model to train or serve.
They are the sharpest comparator a cheap panel has, and we treat them as one.
On our class-balanced sets, a published balanced accuracy converts exactly to our primary metric as $\kp = 2\,\bacc - 1$.
We therefore place those anchors, specialist, guard-tuned, and frontier-prompting alike~\citep{Song_Wang_Zhu_etal_24,Kovacs_Recski_25}, beside our own panel on one scale in~\S\ref{sec:frontier} (Table~\ref{tab:extbase}).
The guard-tuned row is read off the leaderboard, for the reason Appendix~\ref{app:ptj} gives.
The juxtaposition, however, carries the usual caveats, since factuality-metric comparisons are fragile to exactly these differences of protocol~\citep{Godbole_Jia_25}.
Nor do we claim that the two are interchangeable.
A specialist is trained for reference-conditioned verification, sometimes on the target corpus's own training split~\citep{Tamber_Bao_Xu_etal_25}, and so holds an advantage on its home ground;
yet it cannot score the ungrounded half of our suite at all.
Fine-tuned judges are moreover reported to behave as task-specific classifiers that generalise poorly off their training distribution~\citep{Huang_Bu_Zhou_etal_25}.%

\paragraph{Benchmarks.}
We close this section with the benchmarks on which all these approaches are compared.
Hallucination resists a single formal definition~\citep{Cossio_25,Bang_Ji_Schelten_etal_25}, and the eight benchmarks we adopt accordingly span the range, from grounded faithfulness against an explicit reference to ungrounded factuality against world knowledge (\S\ref{ssec:data}).
State-of-the-art detectors nonetheless remain far from saturating such benchmarks~\citep{Bao_Li_Qu_etal_25,Pandit_Xu_Hong_etal_25}, which keeps the cheap-panel question open.

\section{Extended limitations}
\label{app:limits}
The conclusion names the main limitation, \ie the roster the panel is built from;
this appendix reports the other conditions we have identified that bound --without overturning-- the claims made in the main text.

\paragraph{Costing.}
Our costing is an estimate rather than a quotation.
It prices \textsc{gpu} time at a public reference rate for the card the panel was benchmarked on, and the frontier at list \textsc{api} pricing on one date, so the break-even volume moves with both.
It also prices metered use rather than the hardware footprint a deployer must actually provision (Appendix~\ref{app:cost}).
We price a label at \$$1$, two minutes at \$$30$ per hour;
a domain requiring an expert annotator, of which \medhallu's clinical abstracts are the suite's clearest case, scales that price and the break-even volume with it.
The outlay nonetheless stays one-time, so it moves where the crossing point falls and not whether there is one.
The costing likewise prices inference and not the orchestration a ten-member panel needs, which a deployer serving the members sequentially pays in latency rather than in hardware (Appendix~\ref{app:cost}).

\paragraph{Frame sensitivity and quantisation.}
The frame is fixed per dataset rather than tuned per judge.
On the four claim-verification sets it is the better of the two frames we ran, for the frontier judge as well as for the panel (Table~\ref{tab:frame}).
A frontier-specific prompt search could nonetheless move the \expertqa reversal, which rests on the suite's weakest frontier score;
few-shot exemplars, in particular, are known to lift frontier fact verifiers~\citep{Seo_etal_25}, and handing the frontier judge our calibration records in that form is a matched-label comparison we do not run.
The open-judge scaling check uses 4-bit \textsc{awq} open judges, so the scaling conclusion is scoped to that quantisation (Appendix~\ref{app:scaling}).

\paragraph{External comparisons.}
These rest on published data obtained under their authors' own protocols:
four datasets share their rows with us (Table~\ref{tab:extbase}), and two more are matched only in metric, on a different slice and, on \ragtruth, at a different prevalence (Table~\ref{tab:extf1}); hence on \medhallu we compare levels rather than ranks, and on \ragtruth we report only the pattern within the published anchors themselves.

We re-score six published results ourselves.
Two \aggrefact leaderboard rows agree to about a point and three \medhallu members to within $2.6$ F1, whereas one checkpoint (\mdname{Bespoke-MiniCheck-7B}) we do not reproduce (Appendix~\ref{app:extbase}).
A cross-paper juxtaposition nonetheless remains weaker evidence than a within-suite contrast~\citep{Godbole_Jia_25}.
The effective-independence comparison of~\S\ref{ssec:mechanism} is across studies in the same sense:
$\bar\phi$ and $n_{\mathrm{eff}}$ are panel-internal and so port between settings, but the tasks, the gold, and the panel size differ, and it is a comparison of a derived statistic and not of a score (Appendix~\ref{app:neff}).

\paragraph{Scope of the panel and of its competitors.}
We reduce each judge's task to a binary faithfulness verdict and read each member through that verdict alone;
graded or span-level rubrics, richer aggregators than a stacked logistic regression, and interfaces that expose a member's confidence rather than its decision are all out of the current scope, and the last of these bounds the panel from a direction we do not measure.
Our competitors are off-the-shelf judges, other aggregators of them, and the purpose-trained detectors, which we run ourselves across the suite (Appendix~\ref{app:ptj});
we do not, however, re-train a specialist on the labelling budget the stacker consumes.
Although we sweep panel \emph{size} (Figure~\ref{fig:budget}(a)), we do not vary its \emph{composition}.
We thus characterise when aggregating one fixed roster of frozen generalists helps, not whether another roster, or a matched-budget specialist, would do better.
Symmetrically, the frontier side of the comparison is a single judge, \sonnetfull, so the substitution is priced against that model and not against frontier judges at large.
Nor do we measure an escalation cascade, which keeps the frontier judge for the items a cheap model is unsure of.
We read it as complementary to the panel rather than as a competitor (Appendix~\ref{app:related}), and leave open what share of frontier calls a panel-first cascade would retain at matched quality.

\paragraph{Class balance.}
Every set is built at an exact $50/50$ balance (Appendix~\ref{app:construction}), which a deployment stream need not share.
Under a shift in prevalence alone, each member's per-class recalls held fixed, balanced accuracy is unchanged by construction, whereas \kp\ is not, depending on prevalence even at fixed recalls, and the identity $\kp = 2\,\bacc - 1$ no longer holds.
The stacker, fitted on a balanced calibration sample and read at $0.5$, targets to a first approximation the balanced rather than the plain error, so a deployer minimising the latter would shift its intercept by the log-ratio of the deployment priors or, more simply, draw the calibration sample from the deployment stream itself.
We evaluate neither remedy, so our \kp\ readings hold for balanced evaluation only, and so does the admissibility rule, whose two statistics are computed on the same balanced samples.

\paragraph{Evidential base.}
The account is supported at the item level over nearly two thousand points and within every dataset (Appendix~\ref{app:predictor}), but the \emph{cross-dataset} ordering of the differences rests on eight datasets, so any dataset-level competence/correlation predictor is illustrative rather than confirmatory.
The account and the admissibility rule were moreover formed on these same eight datasets, and with a single clear gain over the best single judge in the suite they are consistent with it rather than validated by it.
Furthermore, the suite never leaves English, its eight sets being drawn from biomedical, encyclopaedic, news, and open-domain question-answering sources, so whether the two statistics keep their reading in other languages or domains is untested.
Finally, both Cohen's \kp\ and balanced accuracy inherit whatever label noise the underlying benchmarks carry.

\end{document}